\documentclass[lettersize,journal]{IEEEtran}
\usepackage{amsmath,amsfonts}
\usepackage{algorithmic}
\usepackage{amsthm}
\usepackage{algorithm}
\usepackage{array}
\usepackage{textcomp}
\usepackage{multicol}
\usepackage{adjustbox}
\usepackage{makecell}
\usepackage{ctable}
\usepackage{arydshln} 
\usepackage{booktabs} 
\usepackage{multirow}
\usepackage{stfloats}
\usepackage{url}
\usepackage{verbatim}
\usepackage{graphicx}
\usepackage{bbding}
\usepackage{pifont}
\usepackage{cite}
\usepackage{subfigure}
\usepackage{textcomp}
\usepackage{multicol}
\usepackage{adjustbox}
\usepackage{makecell}
\usepackage{ctable}
\usepackage{booktabs} 
\usepackage{multirow}
\usepackage{stfloats}
\usepackage{subfigure}
\usepackage{url}
\usepackage{bbding}
\usepackage{pifont}
\usepackage{amssymb}

\newcommand{\cmark}{\checkmark}
\newcommand{\cmarkHaar}{\textcolor{blue}{\ding{51}}} 

\theoremstyle{definition}
\newtheorem{definition}{Definition}[section]

\begin{document}

\title{Image Denoising Using Global and Local \\ Circulant Representation}

\author{Zhaoming Kong, Xiaowei Yang, and Jiahuan Zhang
\thanks{Z. Kong and X. Yang are with the School of Software Engineering, South China University of Technology, Guangzhou, 510006, China (e-mail: kong.zm@mail.scut.edu.cn; xwyang@scut.edu.cn).}
\thanks{J. Zhang is with the Department of Clinical Laboratory Medicine, Guangdong Provincial People's Hospital, Southern Medical University, Guangzhou, 510000, China (email: zhangjiahuan@gdph.org.cn).}}

\maketitle

\begin{abstract}
The proliferation of imaging devices and countless image data generated every day impose an increasingly high demand on efficient and effective image denoising. In this paper, we establish a theoretical connection between principal component analysis (PCA) and the Haar transform under circulant representation, and present a computationally simple denoising algorithm. The proposed method, termed Haar-tSVD, exploits a unified tensor singular value decomposition (t-SVD) projection combined with Haar transform to efficiently capture global and local patch correlations. Haar-tSVD operates as a one-step, parallelizable plug-and-play denoiser that eliminates the need for learning local bases, thereby striking a balance between denoising speed and performance. Besides, an adaptive noise estimation scheme is introduced to improve robustness according to eigenvalue analysis of the circulant structure. To further enhance the performance under severe noise conditions, we integrate deep neural networks with Haar-tSVD based on the established Haar-PCA relationship. Experimental results on various denoising datasets demonstrate the efficiency and effectiveness of proposed method for noise removal. Our code is publicly available at https://github.com/ZhaomingKong/Haar-tSVD.
\end{abstract}

\section{Introduction}
\IEEEPARstart{T}he rapid development of modern imaging systems and technologies has significantly enriched the information captured and conveyed by images, delivering a more faithful representation for real scenes. However, images are inevitably corrupted by noise during acquisition and transmission, which can severely degrade the visual quality of acquired data. Therefore, effective image denoising remains a fundamental task in many applications, including feature extraction, object tracking, and medical diagnosis \cite{elad2023image, Collection_denoising_methods}. \\
\indent Many state-of-the-art image denoising techniques weigh highly on deep neural network (DNN) architectures, which are usually trained on datasets containing clean and noisy image pairs. For example, Zhang et al. \cite{zhang2017beyond} incorporated batch normalization (BN) \cite{ioffe2015batch}, rectified linear unit (ReLU) \cite{nair2010rectified} and residual learning \cite{he2016deep} into the convolutional neural network (CNN) model. Chen et al. \cite{chen2022simple}
considered an efficient activation free network. Zamir et al. \cite{zamir2022restormer} introduced transformer \cite{vaswani2017attention} to capture long-range image pixel interactions. While the data-driven models have shown impressive performance for image restoration, their performance may degrade sharply when the test images do not match the distribution of the current scene. Besides, collecting high-quality data is time-consuming and expensive, and the platform required for training and inference may be inaccessible to ordinary users and researchers. \\
\indent Therefore, it is interesting to devise denoising methods with less dependence on ground-truth data and low computational complexity. As an alternative to DNN models, the classic patch-based denoising framework still shows competitive performance in various tasks \cite{kong2023comparison}. In general, related works filter out noise based solely on the input noisy observation with different regularization terms and image priors \cite{katkovnik2010local, zoran2011learning}. The representative BM3D method \cite{dabov2007image} integrated the NLSS characteristic of natural images \cite{buades2005review}, sparse representation \cite{elad2006image} and transform-domain techniques \cite{yaroslavsky2001transform} into a subtle paradigm. Gu et al. \cite{gu2014weighted} replaced the sparsity constraint with the low-rank assumption. Xu et al. \cite{xu2018trilateral} employed the Maximum A-Posterior (MAP) estimation technique \cite{murphy2012machine} and proposed a trilateral weighted sparse coding scheme.\\
\indent Despite steady progress, existing methods still face several inherent limitations \cite{lucas2018using}, including the need for solving complex optimization problems in the test phase, reliance on manually tuned parameters, and insufficient exploitation of auxiliary information. To circumvent these issues and achieve a balance between denoising performance and computational resources, we take advantage of the circulant representation \cite{tee2007eigenvectors, zhang2016exact} and propose Haar-tSVD to capture the local and nonlocal similarity. Furthermore, we investigate an effective combination of DNNs and the proposed denoiser to utilize both external information and internal image structures. Following the patch-based denoising paradigm, the proposed method is distinguished by several characteristics. \\
\indent First, we exploit the global circulant representation to model patch-level correlation, which can be efficiently captured by a pair of shared t-SVD bases. We then extend this global patch circulant structure to a local group-level circulant formulation, thereby revealing an intrinsic connection between the Haar transform and PCA. The resulting global-local circulant correlation leads to a unified integration of t-SVD and Haar bases, eliminating the need for local transform learning, yielding a parallelizable and plug-and-play denoiser. Besides, to improve the adaptability and avoid manual parameter tuning, we develop an adaptive variant termed A-Haar-tSVD, which adjusts the noise estimation value based on the eigenvalue characteristics of circulant structures. Moreover, to improve robustness under severe noise, we propose fusing DNNs with the established Haar–PCA link to leverage both external information and intrinsic image features. \\
\indent Our main contributions can be summarized as follows: 
\begin{itemize}
  \item \textbf{Unified bases:} We propose to leverage the circulant representation to capture both the intra- and inter-correlations among image patches. Furthermore, by building the connection between the Haar transform and PCA via the eigenvalue decomposition (EVD) of the circulant structure, we demonstrate that the global-local circulant similarity can be efficiently exploited by a unified t-SVD projection and Haar transform. 
  \item \textbf{Adaptive scheme:} We develop an adaptive scheme to enhance the flexibility and robustness of the proposed method by exploring a CNN-based noise estimator alongside eigenvalue characteristics of circulant structures. To reduce computational overhead, a fast implementation is achieved through parallel programming techniques.
  \item \textbf{Enhancement strategy:} We leverage the established Haar–PCA relationship and introduce a learning-based enhancement strategy that integrates a data-driven module with the proposed denoiser, improving robustness and denoising performance in challenging scenarios.
  \item \textbf{Diverse datasets:} We evaluate the applicability of the proposed method across different real-world denoising tasks such as images, videos and hyperspectral imaging (HSI). Experiments demonstrate the competitive performance of the proposed method in terms of both effectiveness and efficiency.
\end{itemize}

The rest of the paper is structured as follows. Section II summarizes related works. Section III describes the proposed Haar-tSVD denoising method and its variants in detail. Section IV presents datasets, experimental settings and results. Besides, discussions of the ablation studies are included in this section. Section V concludes this work.

\section{Related Works} \label{section_background}
\vspace{-1.9pt}
\subsection{Symbols and Notations}
\indent We follow the tensor notations in \cite{kolda2009tensor} for image representation. Vectors and matrices are first- and second-order tensors, denoted by bold lowercase letters $\mathbf{a}$ and bold capital letters $\mathbf{A}$, respectively. A higher order tensor is denoted by calligraphic letters, e.g., $\mathcal{A}$. An $N$th-order tensor is denoted as $\mathcal{A} \in \mathbb{R}^{I_1\times I_2\times\cdots\times I_N}$. The $n$-mode product of a tensor $\mathcal{A}$ by a matrix $\mathbf{U}\in \mathrm{R}^{P_n\times I_n}$, denoted by $\mathcal{A}\times _n\mathbf{U}$ is also a tensor. The mode-$n$ unfolding of $\mathcal{A}$ is denoted by $\mathbf{A}_{(n)}$. 
\vspace{-2pt}
\vspace{-3.8pt}
\subsection{Patch-based Framework}
\begin{figure}[htbp]
 \vspace{-0.19pt}
  \centering
  \graphicspath{{Figs/Frameworks/}}
  \includegraphics[width=3.46in]{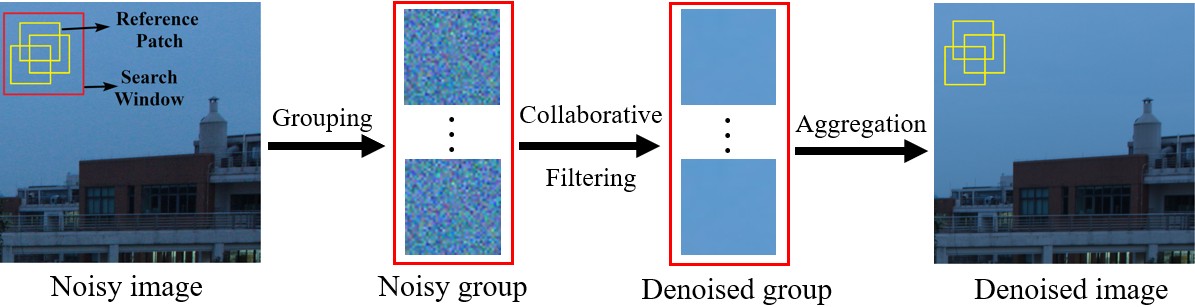}\\
  \vspace{-3.19pt}
  \caption{Illustration of the patch-based framework for traditional denoisers.}
  \label{Fig_traditional_framework}
  \vspace{-8.6pt}
\end{figure}

Patch-based denoisers typically exploit the NLSS property through three stages: grouping, collaborative filtering, and aggregation, as illustrated in Fig. \ref{Fig_traditional_framework}. For a reference patch $\mathcal{P}_{n} \in \mathbb{R}^{ps\times ps\times c}$, the grouping step stacks $K$ similar patches within a local window $W$ into a 4D group $\mathcal{G}_n \in \mathbb{R}^{ps \times ps \times c \times K}$ based on certain patch-matching criteria \cite{foi2007pointwise, buades2016patch, Foi2020}. Collaborative filtering then estimates the underlying clean group via
\begin{equation}\label{tensor_collaborative_filtering}
\hat{\mathcal{G}}_c = \mathop{\arg\min_{\mathcal{G}_c}} \| \mathcal{G}_n - \mathcal{G}_c \|_{F}^2 + \rho\cdot\Psi(\mathcal{G}_c),
\end{equation}
where $\Psi(\cdot)$ encodes prior knowledge. Finally, the clean patches in $\hat{\mathcal{G}}_c$ are aggregated back to their original locations to suppress residual noise.

\vspace{-2.8pt}
\subsection{Circulant Representation and t-SVD for Image Denoising}
\indent Recently, circulant structure and representation \cite{chen2015image, zhang2016exact, kong2019color} has been utilized for image restoration due to its effectiveness of exploring redundancy and encoding the inner structure of an image patch by cyclic shift. The block circulant representation (BCR) \cite{zhang2016exact} of an image patch $\mathcal{P} \in \mathbb{R}^{ps \times ps \times c}$ models patch-level redundancy as
\begin{equation}\label{Equ_block_circulant_matrix}
  bcirc(\mathcal{P}) = \begin{pmatrix}
                    \mathcal{P}^{(1)} & \mathcal{P}^{(c)} & \cdots & \mathcal{P}^{(2)}\\
                    \mathcal{P}^{(2)} & \mathcal{P}^{(1)} & \cdots & \mathcal{P}^{(3)} \\
                    \vdots & \vdots & \ddots & \vdots \\
                    \mathcal{P}^{(c)} & \mathcal{P}^{(c-1)} & \cdots & \mathcal{P}^{(1)} \\
                  \end{pmatrix},
\end{equation}
where $\mathcal{P}^{(i)} = \mathcal{P}(:,:,i)$ denotes the $i$-th frontal slice, and $bcirc(\mathcal{P}) \in \mathbb{R}^{psc \times psc}$ is a block circulant matrix. This representation preserves the patch structure while capturing inter-slice correlations, making it well suited for multi-dimensional image data. Moreover, explicit construction of BCR is unnecessary, since block circulant operations can be efficiently implemented via the tensor t-product \cite{kilmer2011factorization, kilmer2013third}.
\begin{definition}[T-product]
Suppose $\mathcal{A}\in \mathbb{R}^{n_1 \times m \times n_3}$ and $\mathcal{B}\in \mathbb{R}^{m \times n_2 \times n_3}$, then the t-product $\mathcal{C} = \mathcal{A}*\mathcal{B} \in \mathbb{R}^{n_1 \times n_2 \times n_3}$ is defined as 
\begin{equation}\label{Equ_t_product}
  bcirc(\mathcal{C}) = bcirc(\mathcal{A}) bcirc(\mathcal{B}).
\end{equation}
\end{definition}
\noindent Equ. (\ref{Equ_t_product}) can be efficiently computed in the Fourier domain
\begin{equation}\label{Equ_fft_product}
  \mathcal{C}_{F}^{(i)} = \widehat{\mathcal{A}}^{(i)} \widehat{\mathcal{B}}^{(i)}, \quad i = 1, 2, \ldots, m,
\end{equation}
where $\widehat{\mathcal{A}}$ is obtained by applying the fast Fourier transform (FFT) along the third mode of $\mathcal{A}$ via
\begin{equation}\label{Equ_fft_third_mode}
  \widehat{\mathcal{A}} = \mathcal{A} \times _3\mathbf{W}_{FFT}, 
\end{equation} 
where $\mathbf{W}_{FFT}$ refers to the FFT matrix. Then the popular t-SVD \cite{kilmer2013third} can be defined based on the t-product.
\begin{definition}[T-SVD]
For $\mathcal{A} \in \mathbb{R}^{n_1 \times n_2 \times n_3}$, its t-SVD decomposition is given by 
\begin{equation}\label{Equ_t_SVD}
  \mathcal{A} = \mathcal{U} * \mathcal{S} * \mathcal{V}^T,
\end{equation}
\end{definition}
\noindent where $\mathcal{U} \in \mathbb{R}^{n_1 \times n_1 \times n_3}$ and $\mathcal{V} \in \mathbb{R}^{n_2 \times n_2 \times n_3}$ are orthogonal tensors, and the entries in $\mathcal{S}\in \mathbb{R}^{n_1 \times n_2 \times n_3}$ can be viewed as singular values or coefficients of $\mathcal{A}$. \\
\indent t-SVD has proven effective for denoising \cite{zhang2016exact, pd2019entropy, xue2024tensor}, by exploring the block circulant structure of image patches, alleviating tensor imbalance issues \cite{bengua2017efficient}, and sharing properties with matrix-based formulations. However, existing t-SVD–based methods typically require iterative optimization and local basis learning during the test phase. This motivates the design of a simple and effective denoising scheme with t-SVD and circulant representation.
\subsection{Haar Transform}
As a popular transformation in image filtering \cite{blu2007sure, hou2020nlh}, the Haar transform belongs to the wavelet family, which is derived from the Haar matrix \cite{roeser1982fast}. Specifically, starting from a simple $2 \times 2$ Haar transformation matrix
\begin{equation}\label{Equ_Haar_mtx_example}
    \mathbf{H}_2 = \frac{1}{\sqrt{2}}\begin{bmatrix}
        1 &  1\\
        1 & -1
    \end{bmatrix}.
\end{equation}
We can define the general $2N \times 2N$ Haar matrix by
\begin{equation}\label{Equ_general_Haar_mtx}
  \mathbf{H}_{2N} = \frac{1}{\sqrt{2}} \begin{bmatrix}
        \mathbf{H}_N \otimes [1, 1] \\
        \mathbf{I}_N \otimes [1, -1]
    \end{bmatrix},
\end{equation}
where $\mathbf{I}_N \in \mathbb{R}^{N\times N}$ denotes the identity matrix and $\otimes$ represents the Kronecker product \cite{kolda2009tensor}. The Haar matrix plays an important role in the analysis of the localized features due to its simplicity and orthogonality. In this paper, we study the Haar transform from the perspective of circulant structure, and discuss how it can be efficiently and effectively integrated into the patch-based denoising framework.

\section{Method}
\begin{figure*}[htbp]
\graphicspath{{Figs/Frameworks/}}
  \centering
  \includegraphics[width=6.198in]{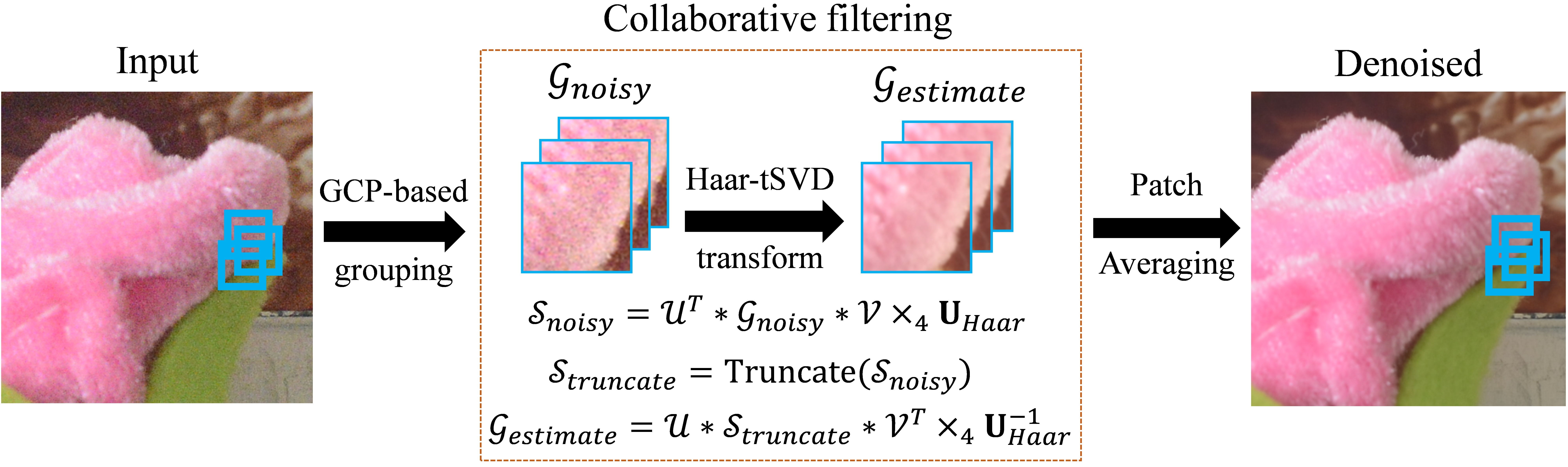}
  \vspace{-2.18pt}
  \caption{Flowchart of the proposed Haar-tSVD method. It is a one-step filtering algorithm built upon global and local circulant representation.}
  \label{Fig_framework_Haar_tSVD}
  \vspace{-12.18pt}
\end{figure*}

In this section, we first provide a detailed description of the proposed Haar-tSVD method, whose flowchart is depicted in Fig. \ref{Fig_framework_Haar_tSVD}. We then introduce its adaptive variant and the corresponding robust enhancement strategy.

\vspace{-5.8pt}

\subsection{Searching Similar Patches}
Directly computing the Euclidean distance between two patches $\mathcal{P}_i, \mathcal{P}_j \in \mathbb{R}^{ps\times ps\times 3}$ is both time-consuming and sensitive to noise. Based on the observation that the green channel generally has higher SNR \cite{guo2021joint}, we adopt a GCP-based patch-matching method \cite{kong2025}, which computes the distance $d_{ij}$ between two patches via
\begin{equation}\label{Equ_GCP_distance_calculation}
  d_{ij}=\left\{
    \begin{aligned}
    &\| \mathcal{P}_{i}^G -  \mathcal{P}_j^G \|, \, \, \, \|\mathcal{P}_{i}^G\| \geq \text{max}(\frac{1}{\gamma}\|\mathcal{P}_{i}^R\|, \frac{1}{\gamma} \|\mathcal{P}_{i}^B\|) \\
    & \| \mathcal{P}_{i}^{avg} -  \mathcal{P}_j^{avg} \|, \, \, \, \text{otherwise},
    \end{aligned}
    \right.
 \end{equation}
where $\mathcal{P}^R$, $\mathcal{P}^G$ and $\mathcal{P}^B$ represents the R, G and B channels of an image patch $\mathcal{P}$, respectively. $\mathcal{P}^{avg}$ is the average value of RGB channels. The weight parameter $\gamma$ is used to measure the importance of the green channel. Similar to \cite{kong2025}, we empirically set $\gamma = 1.2$. This scheme reduces patch-matching time by $\frac{2}{3}$ and improves robustness to noise by leveraging the cleaner green channel, facilitating group-level sparsity for collaborative filtering.

\subsection{Global Circulant Reperesentation}
In the collaborative filtering step, the choice of algebraic representation plays a crucial role. In this section, we present an extension of the nonlocal circulant representation to a global formulation for patch modeling, enabling efficient capture of latent structures across different images.
\subsubsection{Nonlocal Circulant Formulation}
Following \cite{kong2019color}, by integrating the BCR into the nonlocal SVD problem and leveraging the notations of t-SVD and t-product, we can explore the patch-level correlation for each group via 
\vspace{-3.399pt}
\begin{equation}\label{Equ_BCR_nonlocal_SVD_simplified}
\begin{aligned}
  & \text{min} \sum_{i = 1}^K \| \mathcal{P}_i -  \mathcal{U} * \mathcal{S}_i * \mathcal{V}^T\|^2, \\
  & \text{s.t} \quad   \mathcal{U}^T  \mathcal{U}  = \mathcal{I}, \,  \mathcal{V}^T  \mathcal{V}  = \mathcal{I},
\end{aligned}
\vspace{-0.9pt}
\end{equation}
where $\mathcal{S}_i \in \mathbb{R}^{ps\times ps \times 3}$ is the coefficient tensor, and $\mathcal{U},\mathcal{V} \in \mathbb{R}^{ps\times ps \times 3}$ are orthogonal bases. This nonlocal t-SVD can be reduced to independent SVDs in the Fourier domain:
\begin{equation}\label{non_local_SVD_in_fourier_domain}
\vspace{-2.99pt}
\begin{aligned}
  & \, \, \, \, \, \, \, \, \, \, \, \, \text{min} \sum_{i = 1}^K \| \widehat{\mathcal{P}}_{i}^{(j)} -  \widehat{\mathcal{U}}^{(j)} \widehat{\mathcal{S}}_{i}^{(j)} \widehat{\mathcal{V}}^{(j)T}\|^2,\\
  & \text{s.t} \quad   \widehat{\mathcal{U}}^{(j)T} \widehat{\mathcal{U}}^{(j)}  = \mathbf{I}, \,  \widehat{\mathcal{V}}^{(j)T} \widehat{\mathcal{V}}^{(j)}  = \mathbf{I}, \,\, \, \, \forall j \in {1,2,3}, 
\end{aligned}
\vspace{-0.9pt}
\end{equation}
where $\widehat{\mathcal{P}}_{i}^{(j)}$ represents the $j$-th frontal slice of patch $\mathcal{P}_i$ in the Fourier domain. The projection pairs $\widehat{\mathcal{U}}^{(j)}$ and $\widehat{\mathcal{V}}^{(j)}$ are given by the eigenvectors of the ensemble row-wise and column-wise correlation matrices $\mathbf{C}_{row} = \sum_{i=1}^{K} \widehat{\mathcal{P}}_{i}^{(j)}\widehat{\mathcal{P}}_{i}^{(j)T}$ and $\mathbf{C}_{col} = \sum_{i=1}^{K} \widehat{\mathcal{P}}_{i}^{(j)T}\widehat{\mathcal{P}}_{i}^{(j)}$, respectively.
\subsubsection{Global Patch-Level Projection}
\begin{figure}[htbp]
\vspace{-6.18pt}
\graphicspath{{Figs/Fig_UV_projection/}}
  \centering
  \includegraphics[width=2.098in]{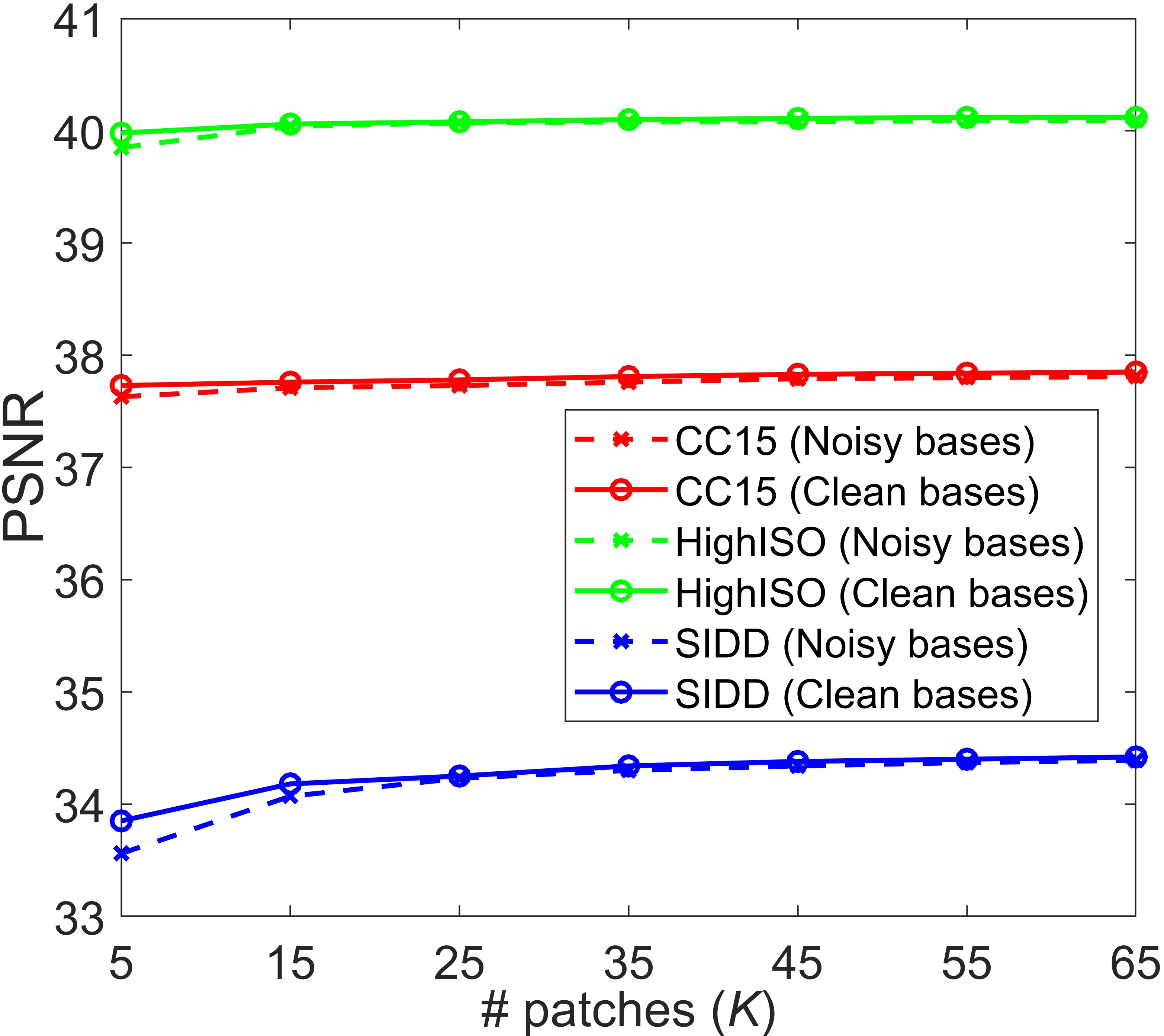}
  \vspace{-3.98pt}
  \caption{Bases $\mathcal{U}$ and $\mathcal{V}$ learned from noisy and clean patches for varying $K$.}
  \label{Fig_UV_projection}
  \vspace{-12.18pt}
\end{figure}

\begin{figure}[htbp]
\vspace{-6.18pt}
\graphicspath{{Figs/Fig_UV_projection/}}
  \centering
  \includegraphics[width=3.5068in]{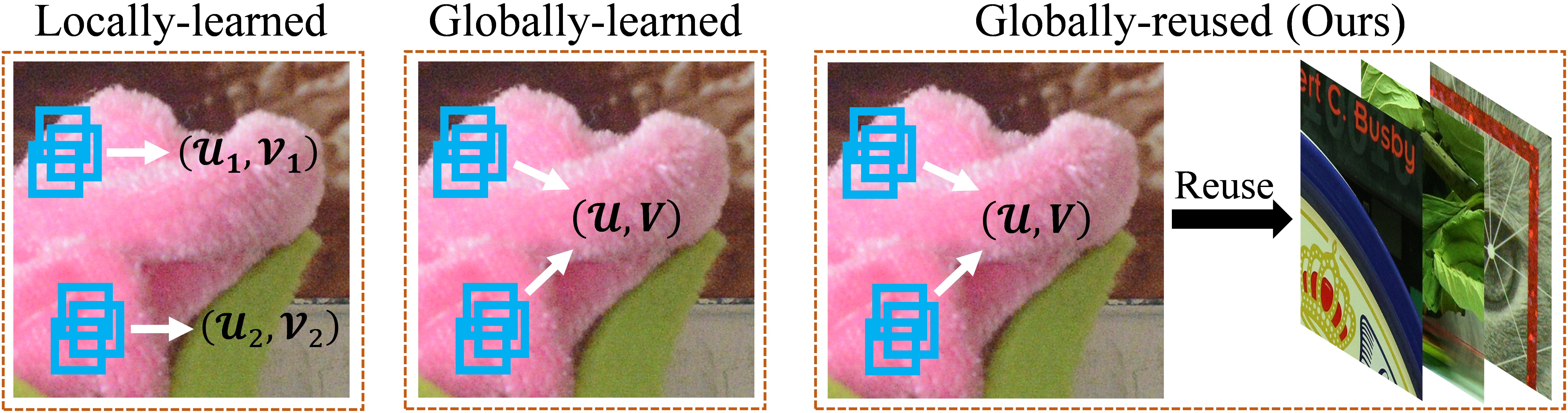}
  \vspace{-13.98pt}
  \caption{Different strategies for obtaining t-SVD bases $\mathcal{U}$ and $\mathcal{V}$.}
  \label{Fig_global_local_learned_bases}
  \vspace{-3.18pt}
\end{figure}

It is noticed that if the observed patches are corrupted by additive white Gaussian noise (AWGN) $\mathbf{n}$ $\sim$ $N(0, \sigma^2)$, then for a large $K$
\begin{equation}\label{Equation_infinite_patches}
\vspace{-2.39pt}
  \sum_{i = 1}^K \widehat{\mathcal{P}}_{noisy_i}^{(j)}\widehat{\mathcal{P}}_{noisy_i}^{(j)T} = \sum_{i = 1}^K \widehat{\mathcal{P}}_{clean_i}^{(j)}\widehat{\mathcal{P}}_{clean_i}^{(j)T} + \sigma^2{\mathbf{I}}.
  \vspace{-0.39pt}
\end{equation}
This implies that the noise-free t-SVD bases of $\mathcal{P}_{clean}$ can be approximated using numerous patches. To assess the effect of $K$, we appy a t-SVD denoiser \cite{kong2019color} on real-world datasets (CC15 \cite{nam2016holistic}, HighISO \cite{yue2019high}, SIDD \cite{abdelhamed2018high}). As shown in Fig. \ref{Fig_UV_projection}, performance gains become marginal for $K>45$, indicating that excessively large groups are unnecessary.
\\
\indent However, learning separate bases for each group \cite{kong2019color} introduces additional complexity. Since an image patch may share features with patches across the entire image, we consider to estimate a global projection pair $\mathcal{U}$ and $\mathcal{V}$ from all reference patches according to Equ. (\ref{non_local_SVD_in_fourier_domain}). We refer to this implementation as \textit{globally-learned} t-SVD. Besides, it is interesting to investigate if the t-SVD bases acquired from a single image can be transferred to other images. Specifically, we utilize the first image from the CC15 dataset to learn a global pair of $\mathcal{U}$ and $\mathcal{V}$, and then reuse them for all subsequent images. This variant is termed the \textit{globally-reused} t-SVD. Fig. \ref{Fig_global_local_learned_bases} illustrates different strategies of obtaining $\mathcal{U}$ and $\mathcal{V}$. As shown in Table \ref{Table_compare_UV_implementation}, the globally-reused approach achieves reliable estimation with reduced computational cost. This suggests that the learned bases can be directly applied to new observations while maintaining performance.

\begin{table}[htbp]
\vspace{-8.98pt}
\scriptsize
  \centering
  \caption{PSNR/SSIM \cite{wang2004image} comparison of t-SVD bases $\mathcal{U}$ and $\mathcal{V}$ obtained with different strategies.}
  \vspace{-2.99pt}
  \scalebox{0.968}{
  \begin{tabular}{ccccc}
    \toprule
    Dataset & Locally-learned & Globally-learned  & Reused (Ours) & Clean \\
    \midrule
    CC15  & 37.86/0.956 & 37.89/0.957 & \textcolor[rgb]{ 0,  .439,  .753}{\textbf{37.90/0.957}} & \textbf{37.92/0.957} \\
    \midrule
    HighISO  & 40.05/0.968 & 40.13/0.969 & \textcolor[rgb]{ 0,  .439,  .753}{\textbf{40.15/0.970}} & \textbf{40.19/0.970} \\
    \midrule
    SIDD-val & 34.35/0.876 & \textcolor[rgb]{ 0,  .439,  .753}{\textbf{34.39/0.876}} & 34.38/0.876 & \textbf{34.45/0.876} \\
    \midrule
    Time (s) & 6.95  & 6.13  & \textbf{6.02} & - \\
    \bottomrule
    \end{tabular}}%
  \label{Table_compare_UV_implementation}%
  \vspace{-6.98pt}
\end{table}%

\vspace{-6.18pt}

\subsection{Local Circulant Structure}
To further promote sparsity and suppress noise in the transform domain, exploiting group-level correlations among similar patches is essential. A common approach is to learn an invertible transform $\mathbf{U}_{group}\in \mathbb{R}^{K\times K}$ along the grouping dimension via PCA \cite{zhang2010two, guo2015efficient, kong2019color}. However, performing local PCA for each group incurs considerable computational cost. In this section, we present an alternative approach to bypass the explicit PCA learning.
\subsubsection{Group circulant formulation} Due to limited search windows and noise contamination, only a small number of similar patches can usually be identified. To better exploit their correlations, we propose to extend the circulant structure to the stacked similar patches, enabling recursive modeling of their interdependencies. Specifically, the circulant structure of a group $\mathcal{G}$ containing $K$ image patches can be defined as
\begin{equation}\label{Equ_group_circulant_structure}
  circ(\mathcal{G}) = \begin{pmatrix}
                    \mathbf{p}_1^T & \mathbf{p}_K^T & \cdots & \mathbf{p}_2^T\\
                    \mathbf{p}_2^T & \mathbf{p}_1^T & \cdots & \mathbf{p}_{K-1}^T \\
                    \vdots & \vdots & \ddots & \vdots \\
                    \mathbf{p}_K^T & \mathbf{p}_{K-1}^T & \cdots & \mathbf{p}_1^T \\
                  \end{pmatrix} \in \mathbb{R}^{K \times 3Kps^2},
\end{equation}
where $\mathbf{p}_i \in \mathbb{R}^{3ps^2\times 1}$ is the vector form of the $i$-th patch $\mathcal{P}_i$ of $\mathcal{G}$. This construction replicates each patch across all rows, thereby explicitly encoding group-level redundancy. To efficiently exploit such correlations without explicit learning on the large matrix $circ(\mathcal{G})$, we examine its structural property. Let $\mathbf{r}^T_i$ denotes the $i$-th row of $circ(\mathcal{G})$, then 
\begin{equation}\label{Equ_row_of_circ_G}
  \mathbf{r}^T_i = \big[ \mathbf{p}_{(j+i-2 \bmod K)+1}^T \big]_{j=1}^{K}, \, \, \forall i = 1,2,\ldots, K.
\end{equation}
From Equ. (\ref{Equ_group_circulant_structure}) and Equ. (\ref{Equ_row_of_circ_G}), we notice that each row is a cyclic permutation of $K$ patches, thus the Gram matrix $circ(\mathcal{G})circ(\mathcal{G})^T \in \mathbb{R}^{K\times K}$ preserves the circulant structure:
\begin{equation}\label{Equ_circ_circ_T}
  circ(\mathcal{G})circ(\mathcal{G})^T = \begin{pmatrix}
                    \mathbf{r}_1^T \mathbf{r}_1& \mathbf{r}_1^T \mathbf{r}_2 & \cdots & \mathbf{r}_1^T \mathbf{r}_K\\
                    \mathbf{r}_2^T \mathbf{r}_1 & \mathbf{r}_2^T \mathbf{r}_2 & \cdots & \mathbf{r}_2^T \mathbf{r}_K \\
                    \vdots & \vdots & \ddots & \vdots \\
                    \mathbf{r}_K^T \mathbf{r}_1 & \mathbf{r}_K^T \mathbf{r}_2 & \cdots & \mathbf{r}_K^T \mathbf{r}_K \\
                  \end{pmatrix}.
\end{equation}
\subsubsection{Relationship to PCA and the Haar Transform} The EVD of the matrix in Equ. (\ref{Equ_circ_circ_T}) is given by
\begin{equation}\label{Equ_eigen_group_circ}
\vspace{-1.98pt}
  circ(\mathcal{G})circ(\mathcal{G})^T\mathbf{u} = \lambda\mathbf{u},
\vspace{-1.98pt}
\end{equation}
where $\lambda$ and $\mathbf{u}\in\mathbb{R}^{K}$ denote an eigenvalue and its corresponding eigenvector, respectively. An interesting observation of the circulant pattern $circ(\mathcal{G})$ from Equ. (\ref{Equ_group_circulant_structure}) and Equ. (\ref{Equ_row_of_circ_G}) is that the sum of each row and column of $circ(\mathcal{G})$ is identical. Denoting this common row sum by $\mathbf{r}_{sum}^T$, we have
\vspace{-3.98pt}
\begin{equation}\label{Equ_sum_row}
  \mathbf{r}_{sum}^T = \sum_{i = 1}^{K} \mathbf{p}_i^T.
  \vspace{-1.28pt}
\end{equation}
Leveraging the circulant property of both $circ(\mathcal{G})$ and its Gram matrix in Equ. (\ref{Equ_circ_circ_T}), the dominant eigenpair $(\lambda_{max}, \mathbf{u}_{max})$ of $circ(\mathcal{G})circ(\mathcal{G})^T$ can be obtained by
\vspace{-2.98pt}
\begin{equation}\label{Equ_first_eigen}
\begin{aligned} 
    \lambda_{max} & = \sum_{i = 1}^{K} \mathbf{p}_i^T \sum_{i = 1}^{K} \mathbf{p}_i= \mathbf{r}_{sum}^T \mathbf{r}_{sum}, \\ 
    \mathbf{u}_{max} & = \frac{1}{\sqrt{K}}(1,1,\ldots,1)^T.
\end{aligned}
\vspace{-0.38pt}
\end{equation}
Interestingly, when $K$ is a power of 2, the dominant eigenvector $\mathbf{u}_{max}^T$ coincides with the first row of the Haar transform matrix defined in Equ. (\ref{Equ_general_Haar_mtx}). This reveals that the Haar transform inherently encodes the first principal component of $circ(\mathcal{G})circ(\mathcal{G})^T$. Since noise can be suppressed by discarding small coefficients in the transform domain, the orthogonal Haar bases provide an efficient data-agnostic alternative to PCA under circulant formulation. Accordingly, the group-level transform can be modeled by the Haar transform. \\
\indent Employing the Haar transform enjoys two computational benefits. First, it enables modeling the group-level redundancy without explicitly constructing the circulant matrix. Second, the predefined Haar matrix $\mathbf{U}_{Haar}$ eliminates the need to obtain a distinct $\mathbf{U}_{group}$ for each group $\mathcal{G}_{noisy}$. 
\vspace{-5.8pt}
\subsection{The Haar-tSVD Transform} 
\subsubsection{One-step Filtering} The above analysis of the circulant representation allows us to get rid of local transform learning. By combining the patch-level projection pair $(\mathcal{U}, \mathcal{V})$ with the group-level Haar matrix $\mathbf{U}_{group}$, we obtain a plug-and-play transform. As a result, the proposed Haar-tSVD method admits a simple one-step collaborative filtering scheme, as summarized in Algorithm \ref{Algorithm_Haar-tSVD}. Specifically, the coefficients $\mathcal{S}_{noisy}$ can be derived by performing the forward transform $\mathcal{T}_{forward}$ via
\begin{equation}\label{Equ_forward_transform_all}
  \mathcal{S}_{noisy} = \mathcal{U}^T*\mathcal{G}_{noisy}*\mathcal{V} \times _4\mathbf{U}_{Haar}.
\end{equation}
Hard-thresholding $\mathcal{T}_{threshold}$ \cite{donoho1994ideal} is then applied to shrink the coefficients of $\mathcal{S}_{noisy}$ under threshold $\tau$ via
 \begin{equation}\label{Equ_hard_thresholding}
 \vspace{-1.8pt}
  \mathcal{S}_{truncate}=\left\{
    \begin{aligned}
    {\mathcal{S}}_{noisy}, \quad |{\mathcal{S}}_{noisy}| \geq \tau, \\
    0, \quad |{\mathcal{S}}_{noisy}| < \tau.
    \end{aligned}
    \right.
 \end{equation}
The estimated clean group $\mathcal{G}_{estimate}$ is recovered by taking the inverse $\mathcal{T}_{inverse}$ of Equ. (\ref{Equ_forward_transform_all}) with $\mathcal{S}_{truncate}$:
\begin{equation}\label{Equ_backward_transform_all}
  \mathcal{G}_{estimate} = \mathcal{U}*\mathcal{S}_{truncate}*\mathcal{V}^T \times _4\mathbf{U}_{Haar}^{-1}.
\end{equation}
Finally, the denoised patches in $\mathcal{G}{estimate}$ are aggregated back to their original image locations.
\setlength{\textfloatsep}{1pt}
\renewcommand{\algorithmicrequire}{\textbf{Input:}}
\renewcommand{\algorithmicensure}{\textbf{Output:}}
\vspace{-3.98pt}
\begin{algorithm}[!htbp]
\caption{Haar-tSVD}
\label{Algorithm_Haar-tSVD}
\begin{algorithmic}[1]
\REQUIRE Noisy image $\mathcal{Y}$, patch size $ps$, number of similar patches $K$, search window size $W$, and noise level $\sigma$.
\ENSURE Estimated clean image $\mathcal{X}$.
\FOR{each reference patch $\mathcal{P}_{ref}$}
    \STATE \textbf{Grouping:} Search $K$ most similar patches within window $W$ using Equ.~(\ref{Equ_GCP_distance_calculation}) to form group $\mathcal{G}_{noisy}$.
    \STATE \textbf{Filtering:} Apply the forward-threshold-inverse transform in Equs.~(\ref{Equ_forward_transform_all})–(\ref{Equ_backward_transform_all}) for the estimated group via $\mathcal{G}_{estimate} = \mathcal{T}_{inverse} \circ \mathcal{T}_{threshold}\circ\mathcal{T}_{forward}(\mathcal{G}_{noisy})$ .    
    \STATE \textbf{Aggregation:} Averagely write all patches of $\mathcal{G}_{estimate}$ back to their original locations.
\ENDFOR
\end{algorithmic}
\end{algorithm}
\vspace{-6.8pt}

\subsubsection{Complexity analysis} The computational cost of the proposed Haar-tSVD for each local group consists of three parts: (i) the search of $K$ similar patches within a window $O(KW^2ps)$, (ii) the global t-SVD projection $O(Kps^3)$, and (iii) the Haar transform projection $O(K^3)$. By adopting the fast Haar transform \cite{roeser1982fast}, the group-level projection can be reduced to $O(K\log K)$. Therefore, an overall computational complexity of Haar-tSVD is $O([KW^2ps + Kps^3 + KlogK])$. Compared to many effective t-SVD based approaches \cite{zhang2016exact, kong2019color, gong2020low, Shi2022tSVD, kong2025}, the proposed method is more efficient, because it does not involve any local learning procedure. 
\subsubsection{Fast implementation} The proposed method consists of two major components: patch-matching and global-local transform, both of which are inherently parallelizable. Specifically, patch-matching involves only Euclidean distance computations, while the transform bases $\mathcal{U}$, $\mathcal{V}$ and $\mathbf{U}_{Haar}$ are shared by all patch groups. Therefore, the computation is dominated by matrix–vector multiplications, making the method suitable for parallel implementation. We achieve over $10\times$ speedup compared to the baseline MATLAB serial implementation. In practice, further acceleration can be obtained by caching intermediate results such as patch indices and transform coefficients $\mathcal{S}{noisy}$ during parameter selection.
\vspace{-1.8pt}
\subsection{Adaptive Scheme and Enhancement Strategy}
\subsubsection{Adaptive noise estimation} 
\begin{figure}[htbp]
\vspace{-6.8pt}
\graphicspath{{Figs/Method/Noise_estimation/}}
  \centering
  \includegraphics[width=3.18in]{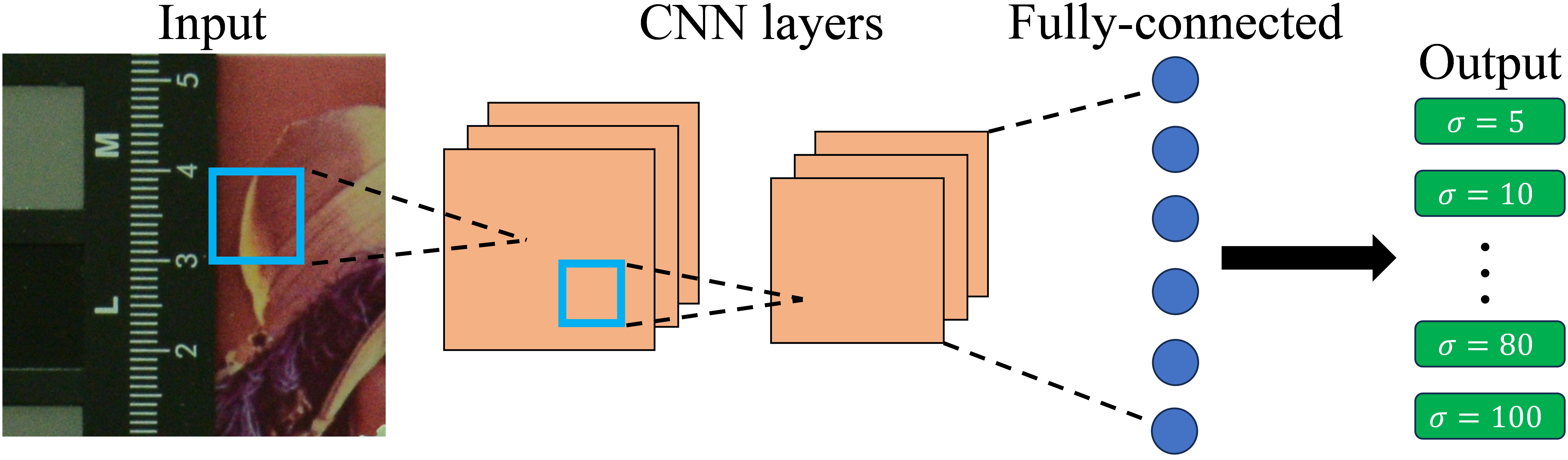}
  \caption{The CNN-based noise estimation approach.}
  \label{Fig_CNN_noise_est}
  \vspace{-6.98pt}
\end{figure}

\begin{figure}[htbp]
\vspace{-3.98pt}
\graphicspath{{Figs/Method/}}
\centering
\subfigure[Comparison of Haar and PCA]{
\label{Fig4}
\includegraphics[width=1.528in]{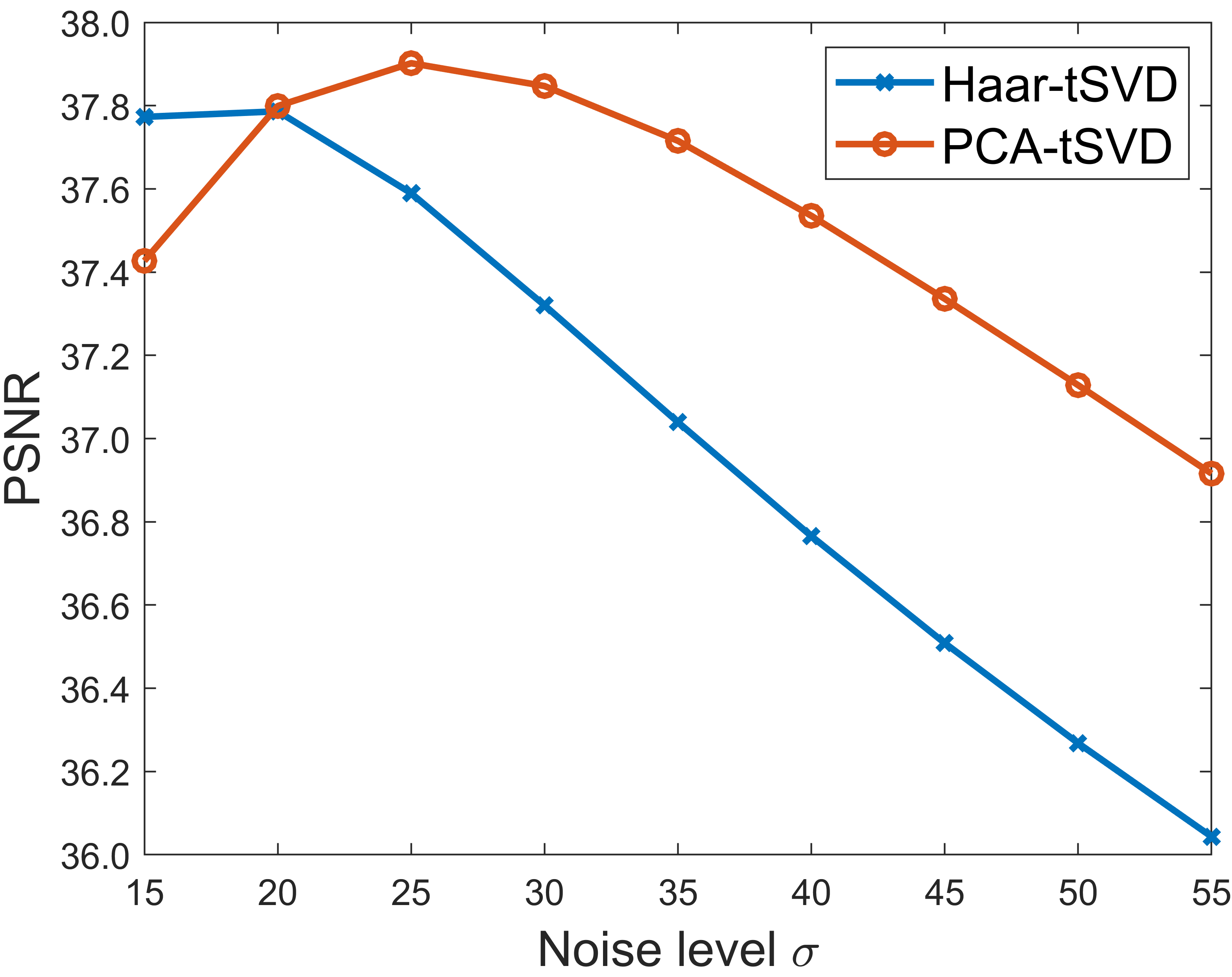}}
\subfigure[Visual effects ($\sigma = 40$)]{
\label{Fig4}
\includegraphics[width=1.566in]{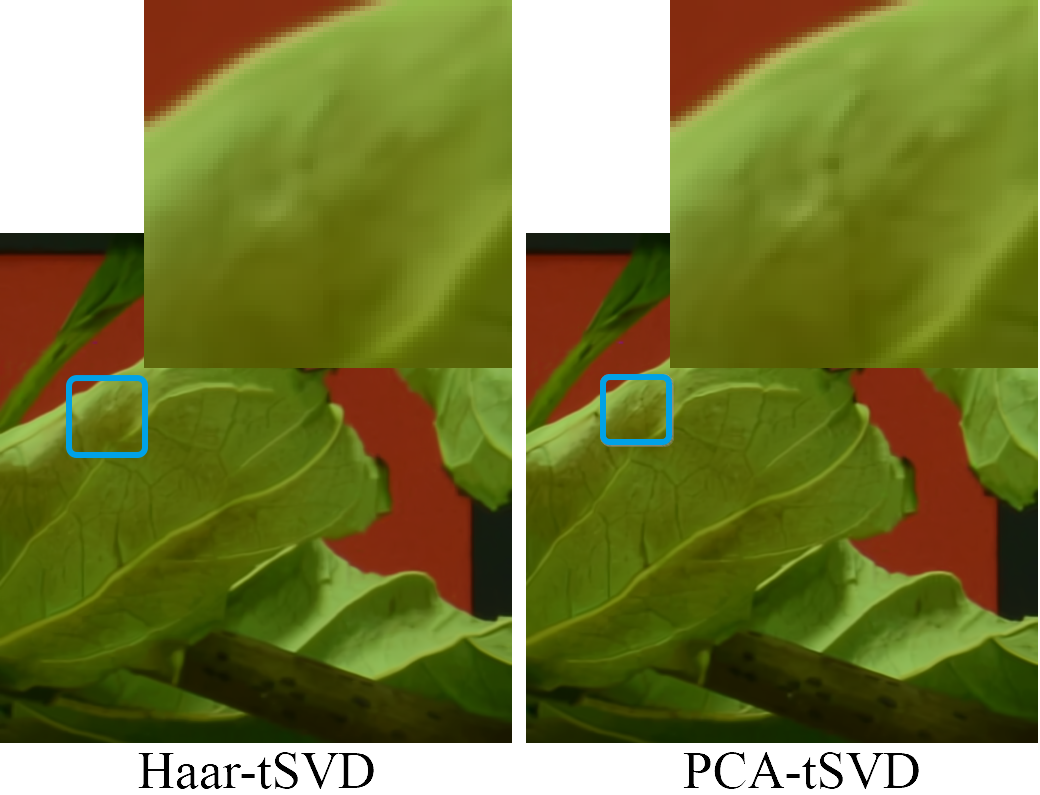}}
\vspace{-2.8pt}
\caption{Denoising effects of the Haar and PCA transforms for $\mathbf{U}_{group}$.}
\label{Fig_compare_Haar_PCA_noise_variation}
\vspace{-3.98pt}
\end{figure}

Apart from an effective collaborative filtering scheme, noise estimation is crucial for an efficient denoiser. For Haar-tSVD, the noise level $\sigma$ characterizes the sparsity of coefficients in the transform domain. To model the sparsity level, we can reformulate the noise estimation problem as a classification task and train a noise estimator using CNN \cite{kong2025}, as illustrated in Fig. \ref{Fig_CNN_noise_est}.
\\
\indent However, we notice that the rationale for incorporating a DNN-based noise estimator also hinges on the robustness of local adaptive transform to small perturbations in $\sigma$. The prediction accuracy of the CNN estimator on real-world data reported in Table \ref{Table_CNN_estimator_accuracy} indicates limited generalization results. Misclassification of the noise level $\sigma$ can degrade the performance of Haar-tSVD. As shown in Fig. \ref{Fig_compare_Haar_PCA_noise_variation}, although Haar-tSVD achieves denoising quality comparable to PCA-tSVD \cite{kong2025}, its fixed bases are more sensitive to noise-level variations, potentially leading to oversmoothing effects.
\begin{table}[htbp]
\vspace{-3.8pt}
  \centering
  \caption{Average prediction accuracy of the CNN-based estimator on real-world datasets.}
    \begin{tabular}{cccc}
    \toprule
    Dataset & CC15  & HighISO & SIDD-validation \\
    \midrule
    Accuracy (\%) & 69.3  & 71.2 & 75.1 \\
    \bottomrule
    \end{tabular}%
  \label{Table_CNN_estimator_accuracy}%
  \vspace{-8.8pt}
\end{table}%

\begin{figure}[htbp]
\vspace{-6.98pt}
\graphicspath{{Figs/Method/Noise_estimation/}}
\centering
\subfigure[Noise-free image]{
\label{Fig4}
\includegraphics[width=0.988in]{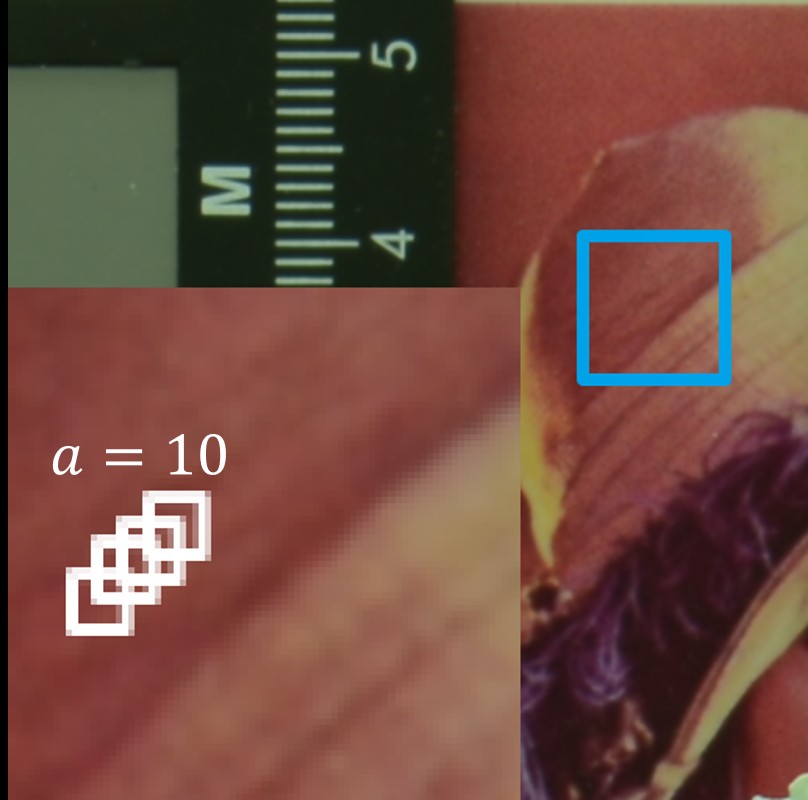}}
\subfigure[Noisy image]{
\label{Fig4}
\includegraphics[width=0.988in]{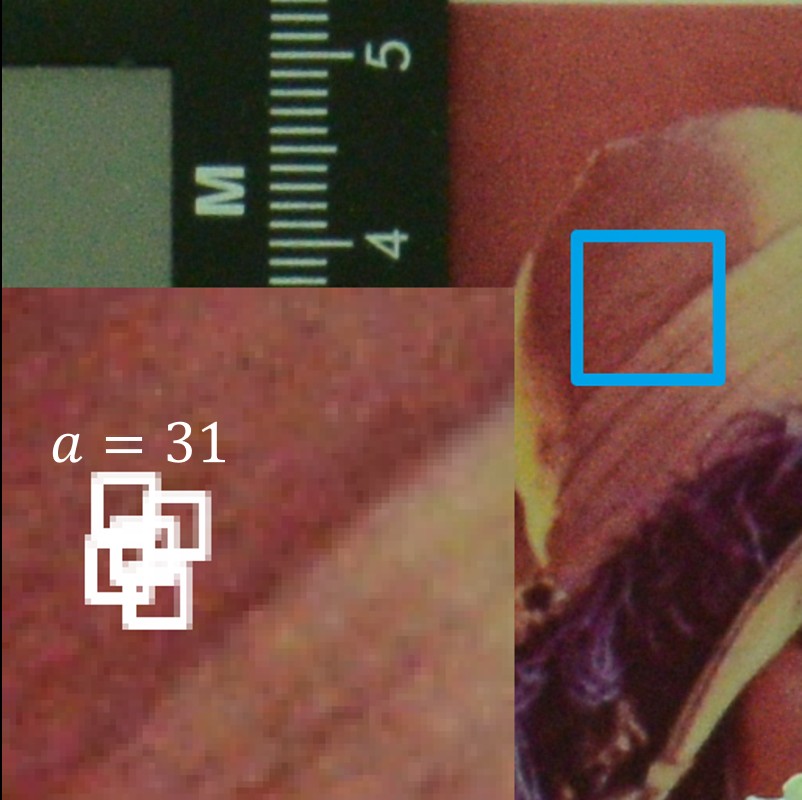}}
\vspace{-3.8pt}
\caption{Influence of real-world noise on patch-search and rank position $a$.}
\vspace{-5.8pt}
\label{Fig_change_lambda}
\end{figure}

\begin{figure}[htbp]
\vspace{-3.98pt}
\graphicspath{{Figs/Frameworks/}}
  \centering
  \includegraphics[width=3.50018in]{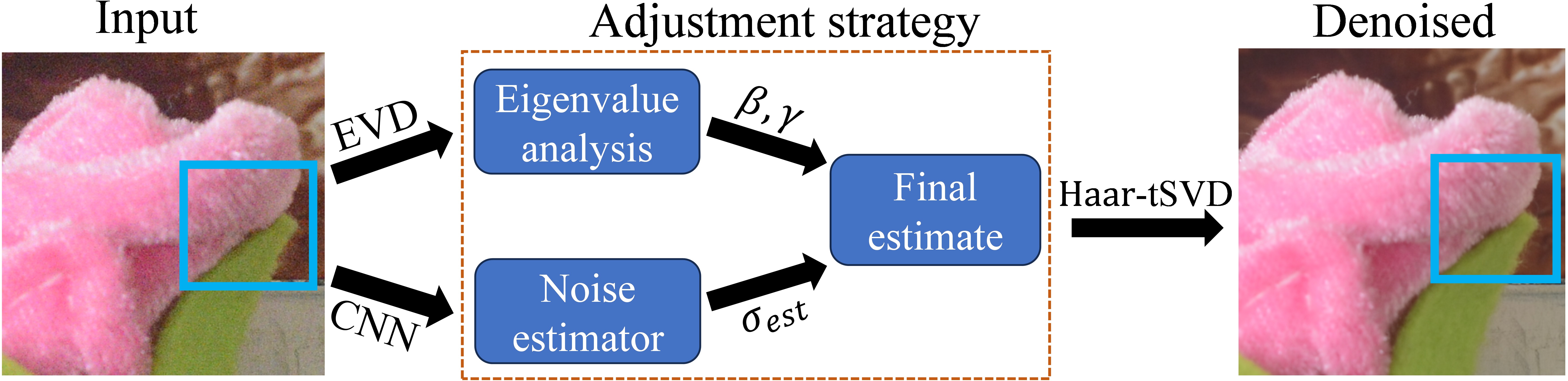}
  \vspace{-9.98pt}
  \caption{Flowchart of the adaptive variant A-Haar-tSVD.}
  \label{Fig_flowchart_A_Haar_tSVD}
  \vspace{-2.698pt}
\end{figure}

  \\
\indent To enhance adaptiveness, we further analyze the eigenvalue characteristics of the circulant matrix $circ(\mathcal{G})circ(\mathcal{G})^T$. Due to its circulant structure, an eigenpair $(\hat{\lambda},\hat{\mathbf{u}})$ is given by
\vspace{-3.68pt}
\begin{equation}\label{Equ_other_eigen_value}
  \begin{aligned} 
    \hat{\lambda} & = \sum_{i = 1}^{K} (-1)^i \mathbf{p}_i^T \sum_{i = 1}^{K} (-1)^i\mathbf{p}_i, \\ 
    \hat{\mathbf{u}} & = \frac{1}{\sqrt{K}}(-1,1,\ldots,-1,1)^T.
  \end{aligned}
\vspace{-0.68pt}  
\end{equation}
Equ. (\ref{Equ_other_eigen_value}) indicates that the eigenpair captures the alternating contrast between adjacent patches within a local group $\mathcal{G}$. Therefore, we can use $\hat{\lambda}$ to measure the inner-group similarity. Specifically, in low-noise conditions, high inner-group similarity yields small $\hat{\lambda}$, whereas severe noise disrupts patch-matching and shifts $\hat{\lambda}$ toward larger eigenvalues. To demonstrate the impact of noise, we denote by $a$ the rank position of $\hat{\lambda}$ among the $K$ eigenvalues sorted in ascending order. As illustrated in Fig. \ref{Fig_change_lambda}, for $K=32$, $a$ increases significantly with noise severity, suggesting that $a$ can serve as an indicator of noise strength.
 \\
\indent Accordingly, the CNN-estimated noise level $\sigma_{est}$ is adaptively adjusted as
\begin{equation}\label{Equ_adjust_noise_level}
  \hat{\sigma}=
  \left\{
    \begin{aligned}
    \frac{1}{\beta} \sigma_{est}, \quad a \leq \gamma, \\
    \sigma_{est}, \quad a > \gamma,
    \end{aligned}
  \right.
\end{equation}
where $\hat{\sigma}$ is the final estimated noise level, $\beta$ and $\gamma$ are weighting parameters empirically set to 1.2 and 13, respectively. From Equ. (\ref{Equ_adjust_noise_level}), we notice that it is unnecessary to perform full EVD to determine $a$. To further reduce computational complexity, we avoid adjusting $\sigma_{est}$ for every local group $\mathcal{G}$. Instead, we randomly sample a few groups within a subimage and obtain the corresponding $\hat{\sigma}$ through majority voting. \\
\indent We term the proposed method with the adaptive noise estimation strategy as \textit{A-Haar-tSVD}. As illustrated in Fig. \ref{Fig_flowchart_A_Haar_tSVD}, the adaptive approach incorporates the effectiveness of CNNs, preserves the simplicity of patch-based framework and leverages the adaptability of eigenvalue analysis.

\subsubsection{Enhancement under severe noise} 
\begin{figure}[htbp]
\graphicspath{{Figs/Method/Robust_Enhancement/SIDD/sample2/combined/}}
\centering
\subfigure[Noisy]{
\label{Fig4}
\includegraphics[width=0.83018in]{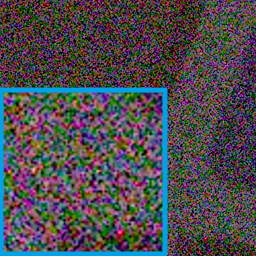}}
\hspace{-0.398em}%
\subfigure[CBM3D]{
\label{Fig4}
\includegraphics[width=0.83018in]{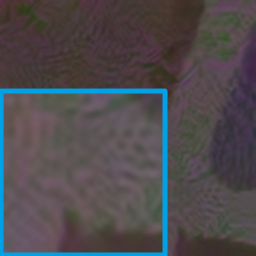}}
\hspace{-0.398em}%
\subfigure[A-Haar-tSVD]{
\label{Fig4}
\includegraphics[width=0.83018in]{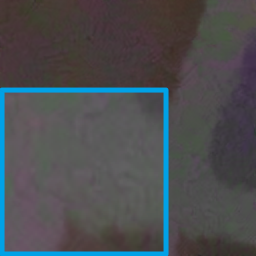}}
\hspace{-0.398em}%
\subfigure[RA-Haar-tSVD]{
\label{Fig4}
\includegraphics[width=0.83018in]{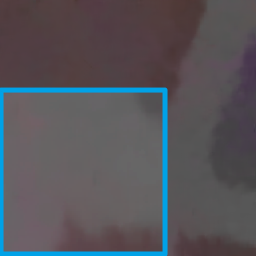}}
\vspace{-5.88pt}
\caption{Effects of the robust enhancement strategy under severe noise.}
\label{Fig_robust_enhance_comparison_SIDD}
\vspace{-1.8pt}
\end{figure}

\begin{figure}[htbp]
\vspace{-6.8pt}
\graphicspath{{Figs/Frameworks/}}
  \centering
  \includegraphics[width=3.50018in]{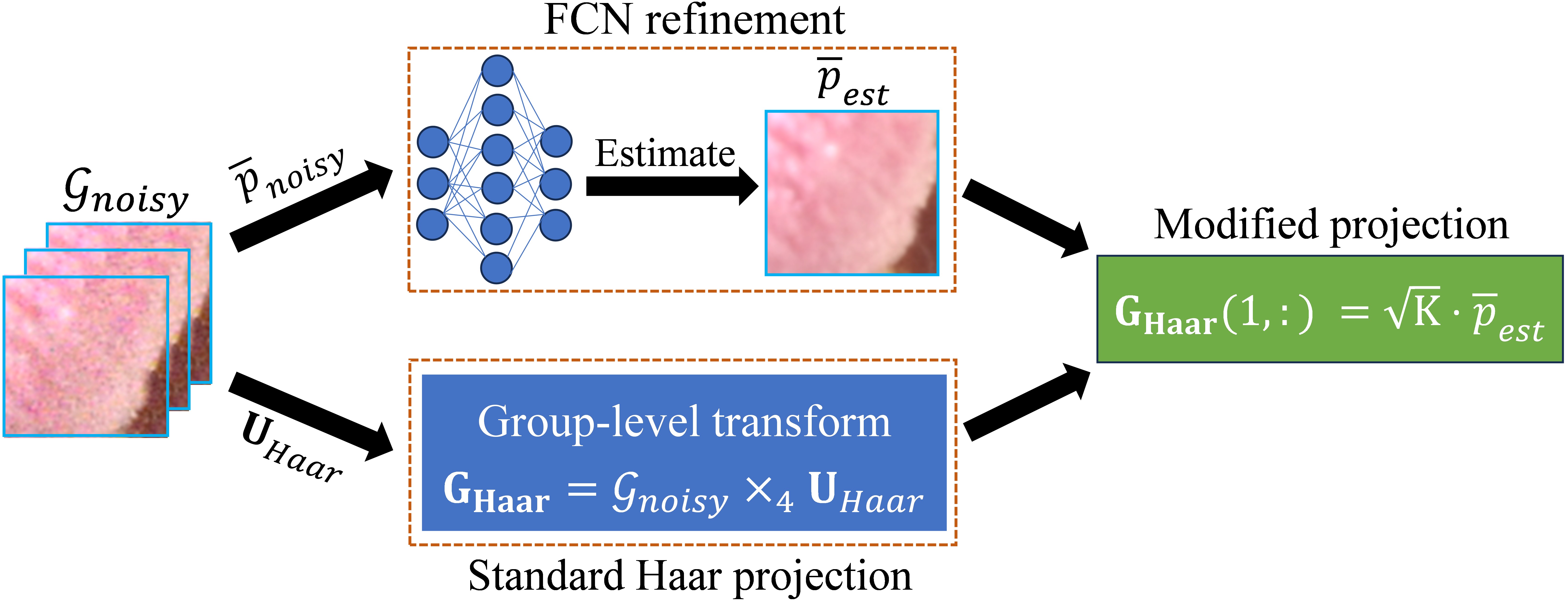}
  \vspace{-9.98pt}
  \caption{Flowchart of the RA-Haar-tSVD enhancement strategy for a group.}
  \label{Fig_flowchart_enhancement_strategy}
  \vspace{-10.698pt}
\end{figure}

Classic patch-based denoisers are particularly subject to severe noise \cite{elad2023image, kong2023comparison}, which degrades patch-matching accuracy and contaminates transform-domain coefficients. As illustrated in Fig. \ref{Fig_robust_enhance_comparison_SIDD}, both the proposed Haar-tSVD and the state-of-the-art BM3D exhibit noticeable color artifacts, even when the noise level is properly estimated. To further improve robustness of the proposed method under severe noise, we present an enhancement strategy, termed \textit{RA-Haar-tSVD}, which integrates a lightweight, trainable refinement module into the Haar-tSVD framework. Fig. \ref{Fig_robust_enhance_comparison_SIDD} shows its effectiveness to remove artifacts and restore natural color in the challenging high-noise scenario.
 \\
\indent According to the Haar-PCA link in Equ. (\ref{Equ_first_eigen}), the first row of the Haar matrix corresponds to the dominant eigenvector
$\mathbf{u}_{max}$ and captures the weighted mean of all patches $\mathbf{p}_{noisy} \in \mathbb{R}^{K \times 1}$ of a group $\mathcal{G}_{noisy}$ in the transform domain:
\begin{equation}\label{Equ_mean_patch}
\vspace{-2.96pt}
  \mathcal{G}_{noisy} \times _4 \mathbf{u}_{max}^T = \frac{1}{\sqrt{K}}\sum_{i = 1}^{K} \mathbf{p}_{noisy_i} = \sqrt{K}\cdot\overline{\mathbf{p}}_{noisy},
  \vspace{-0.36pt}
\end{equation}
where the weighted mean patch $\sqrt{K} \cdot \overline{\mathbf{p}}_{noisy}$ is the first row of the group-level projection $\mathcal{G}_{noisy}\times_4 \mathbf{U}_{Haar}$ and preserves most of its signal energy. However, $\overline{\mathbf{p}}_{noisy}$ is severely contaminated due to the rare-patch effect, which propagates visible artifacts to the reconstructed image. To address this issue, we employ a plain fully connected network (FCN) \cite{burger2012image} to obtain a cleaner version of the dominant coefficients $\overline{\mathbf{p}}_{est}$ via
\begin{equation}\label{Equ_est_p_mean_with_NN}
  \overline{\mathbf{p}}_{est} = FCN_\theta(\overline{\mathbf{p}}_{noisy}).
\end{equation}
\indent As illustrated in Fig. \ref{Fig_flowchart_enhancement_strategy}, the estimated mean patch $\overline{\mathbf{p}}_{est}$ of the FCN is used to refine the first row of the group-level projection $\mathbf{G}_{Haar}$. This strategy offers three advantages: (i) it utilizes the feature extraction ability of the network to obtain the dominant group component under severe noise; (ii) abundant noisy-clean pairs of mean patches are available from overlapping groups for training; and (iii) only the dominant Haar coefficients are modified, the remaining components are preserved and local structural information is retained.

\subsection{Related Patch-based Denoisers}
Table \ref{Table_traditional_method_comparison} compares the proposed Haar-tSVD and its variants with representative patch-based denoisers in terms of learning strategy and implementation. Built upon the t-SVD and Haar transforms, the proposed method absorbs the ideas from the simple and effective design of classic denoisers. Instead of recursively solving complex optimization problems, the proposed Haar-tSVD aims at one-step filtering that eliminates the need to train local bases for each image. Under circulant representation, we employ a unified global t-SVD and Haar transform to efficiently capture patch- and group-level correlation. Besides, EVD is leveraged for adaptive noise estimation, and the integration of NNs with the Haar-tSVD provides an extra enhancement mechanism under severe noise.

\begin{table*}[htbp]
  \centering
  \caption{Representative traditional denoisers with different algebraic representations and enhancement strategies.}
  \scalebox{0.83998}{
    \begin{tabular}{|c|c|c|c|c|c|c|c|c|}
    \hline
    Category & Algebra & Methods & Parallel impl. & Global transform & One-step filter & Adaptive scheme & Enhancement & Key words \\
    \hline
    \multirow{18}{*}{\shortstack[c]{Patch-based \\ denoisers}} 
    & \multirow{6}[3]{*}{Matrix} 
        & DCT \cite{yaroslavsky1996local} & \cmark & \cmark & \cmark & - & - & Discrete cosine transform filter \\
\cline{3-9} & & NLPCA \cite{zhang2010two} & \cmark & - & - & - & \cmark & Nonlocal PCA \\
\cline{3-9} & & MCWNNM \cite{xu2017multi} & - & - & - & - & \cmark & Weighted nuclear norm \\
\cline{3-9} & & TWSC \cite{xu2018trilateral} & - & - & - & \cmark & \cmark & Trilateral sparse coding \\
\cline{3-9} & & GID \cite{xu2018external} & - & - & - & \cmark & \cmark & External + internal prior \\
\cline{3-9} & & NLH \cite{hou2020nlh} & \cmark & \cmark & - & - & \cmark & Haar-based NLM \\
\cline{3-9} & & Bitonic \cite{treece2022real} & \cmark & - & - & \cmark & \cmark & Bitonic filtering \\
\cline{2-9} 
    & \multirow{9}[6]{*}{Tensor} 
       & BM4D \cite{maggioni2012nonlocal} & \cmark & \cmark & - & - & \cmark & 4D extension of BM3D \\
\cline{3-9} & & HOSVD \cite{rajwade2012image} & - & \cmark & \cmark & - & \cmark & 4D HOSVD transform \\
\cline{3-9} & & TDL \cite{peng2014decomposable} & - & - & - & - & \cmark & Tucker dictionary learning \\
\cline{3-9} & & LLRT \cite{chang2017hyper} & - & - & - & \cmark & \cmark & LR tensor + Laplacian \\
\cline{3-9} & & MSt-SVD \cite{kong2019color} & \cmark & - & \cmark & - & - & One-step t-SVD \\
\cline{3-9} & & LTDL \cite{gong2020low} & - & - & - & - & \cmark & LR tensor dictionary learning \\
\cline{3-9} & & GCP-ID \cite{kong2025} & \cmark & - & \cmark & \cmark & - & Green channel prior + t-SVD \\
\cline{3-9} & & Haar-tSVD (Ours) & \cmarkHaar & \cmarkHaar & \cmarkHaar & - & - & Global-local circulant structure \\
\cline{3-9} & & A-Haar-tSVD (Ours) & \cmarkHaar & \cmarkHaar & \cmarkHaar & \cmarkHaar & - & Adaptive noise estimation \\
\cline{3-9} & & RA-Haar-tSVD (Ours) & \cmarkHaar & \cmarkHaar & \cmarkHaar & \cmarkHaar & \cmarkHaar & Adaptive noise estimation \\
    \hline
    \end{tabular}}
  \label{Table_traditional_method_comparison}
  \vspace{-12.98pt}
\end{table*}

\section{Experiments}
In this section, we evaluate about 50 denoising methods across images, video, and HSI datasets. For each method, original implementations or published results are used, with parameters/models carefully chosen for optimal performance. GPU-accelerated methods are executed using the computational resources of Google Colab Pro, while all remaining experiments are conducted on a workstation equipped with an Intel Core i7-10700F CPU @ 2.9 GHz and 16 GB RAM. More details are provided in the supplementary material. 
\vspace{-3.6pt}
\subsection{Implementation Details}
\textbf{Haar-tSVD}. It involves four primary parameters: the patch size $ps$, the number of similar patches $K$ within a group, the window size $W$ for patch search, and the hard-thresholding parameter $\tau$. For image, video and HSI data, we set $ps = 8$, $K = 32$, $W < 20$ and $\tau = \sigma \sqrt{2log(cKps^2)}$, where $c$ denotes the number of channels or spectral bands. \\
\indent \textbf{A-Haar-tSVD} and \textbf{RA-Haar-tSVD}. A multi-layer CNN \cite{pytorch_cifar10_tutorial} and a FCN are adopted for noise level prediction and mean-patch estimation, respectively. The SIDD small and CC60 \cite{xu2018external} data are used for training. For A-Haar-tSVD, each input is divided into subimages of size $128 \times 128$, and the noise level $\sigma$ is selected from $\{1.25, 5, 10, 20, 30, 40, 50, 60, 80, 100, 120\}$. For RA-Haar-tSVD, each input $\overline{\mathbf{p}}_{\text{noisy}}$ is the mean patch of a local group, and the network is trained using residual-learning. 
\subsection{Color Image Denoising}
\indent We perform synthetic and real-world experiments. For synthetic denoising, we use the Kodak dataset \cite{Kodak} and add independent noise to each RGB to simulate channel-specific corruption. In \textit{case 1}, the noise variances are [15, 10, 20], and in \textit{case 2}, they are [40, 30, 50]. For Gaussian denoisers, the noise levels $\sigma$ are set to 20 and 50, respectively.
\begin{table*}[htbp]
\vspace{-1.68pt}
\scriptsize
  \centering
  \vspace{-3.68pt}
  \caption{Denoising comparison on real-world sRGB color image datasets. `*': The results are from the authors' papers.}
  \renewcommand{\arraystretch}{0.50998}
  \scalebox{0.90198}{
    \begin{tabular}{cccccccccc}
    \toprule
    \multicolumn{2}{c}{\multirow{1}[1]{*}{Methods/Models}} & \multirow{2}[1]{*}{Kodak (case 1)} & \multirow{2}[1]{*}{Kodak (case 2)} & \multirow{2}[1]{*}{DND \cite{plotz2017benchmarking}} & \multirow{2}[1]{*}{SIDD \cite{abdelhamed2018high}} & \multirow{2}[1]{*}{CC15 \cite{nam2016holistic}} & \multirow{2}[1]{*}{PolyU \cite{xu2018real}} & \multirow{2}[1]{*}{HighISO \cite{yue2019high}} & \multirow{2}[1]{*}{IOCI \cite{kong2023comparison}} \\
    \multicolumn{2}{c}{} &       &       &       &       &       &       &       &  \\
    \midrule
    \multirow{7}[8]{*}{\shortstack[c]{Traditional \\ denoisers}} & Bitonic  \cite{treece2022real} & -     & -     & 37.85/0.936 & 36.67/0.933 & 35.22/0.924 & 36.64/0.939 & 37.37/0.943 & 39.10/0.954 \\
\cmidrule{2-10}          & MCWNNM \cite{xu2017multi} & 28.40/0.792 & 26.54/0.686 & 37.38/0.929 & 29.54/0.888 & 37.02/0.950 & 38.26/0.965 & 39.89/0.970 & 41.04/0.972 \\
\cmidrule{2-10}          & NLHCC \cite{hou2020nlh}  & -     & -     & 38.85/0.953 & 35.31/0.930 & \textbf{38.49/0.965} & 38.36/0.965 & 40.29/0.971 & 41.22/0.974 \\
\cmidrule{2-10}          & MSt-SVD \cite{kong2019color}  & 33.99/0.906 & 29.12/0.770 & 38.01/0.938 & 34.38/0.901 & 37.95/0.959 & \textcolor[rgb]{ 0,  .439,  .753}{\textbf{38.85/0.971}} & 40.49/0.974 & 41.48/0.977 \\
\cmidrule{2-10}          & CBM3D \cite{dabov2007color} & 33.51/0.893 & 28.90/0.780 & 37.73/0.934 & 34.74/0.922 & 37.70/0.957 & 38.69/0.970 & 40.35/0.974 & 41.46/0.976 \\
\cmidrule{2-10}          & Haar-tSVD (ours) & \textbf{34.13/0.910} & 29.16/0.780 & 38.11/0.939 & 35.05/0.914 & 38.10/0.961 & 38.78/0.969 & 40.51/0.973 & 41.45/0.977 \\
    \midrule
    \midrule
    \multirow{2}[6]{*}{\shortstack[c]{Traditional \\ + DNN}} & A-Haar-tSVD (ours) & 34.04/\textcolor[rgb]{ 0,  .439,  .753}{\textbf{0.907}} & \textcolor[rgb]{ 0,  .439,  .753}{\textbf{29.18/0.783}} & 38.25/0.944 & 35.58/0.925 & 38.24/\textcolor[rgb]{ 0,  .439,  .753}{\textbf{0.963}} & \textbf{38.89/0.971} & \textcolor[rgb]{ 0,  .439,  .753}{\textbf{40.62/0.974}} & \textbf{41.52/0.978} \\
    \cmidrule{2-10}          & RA-Haar-tSVD (ours)  & 33.89/0.901 & \textbf{29.19/0.787} & 38.64/0.945     & 37.48/- & \textcolor[rgb]{ 0,  .439,  .753}{\textbf{38.30}}/0.962 & 38.85/0.970 & \textbf{40.63/0.975} & \textcolor[rgb]{ 0,  .439,  .753}{\textbf{41.50/0.977}} \\
\cmidrule{2-10}          & Pixel2Pixel \cite{ma2025pixel2pixel} & 30.26/0.841 & 26.88/0.718 & -     & 34.34*/- & 35.51/0.923 & 37.19/0.951 & 37.77/0.945 & - \\
    \midrule
    \midrule
    \multirow{6}[12]{*}{\shortstack[c]{DNN models \\ (Self-supervised)}}  & APR-RD \cite{kim2025apr}  & 26.12/0.713 & 23.21/0.634 & 38.57/0.942 & 38.23*/- & 35.83/0.946 & 37.01/0.953 & 38.75/0.968 & 39.36/0.964 \\
\cmidrule{2-10}          & B2UB \cite{wang2022blind2unblind}  & 29.91/0.816 & 26.80/0.698 & -     & -     & 36.51/0.935 & 38.25/0.968 & 39.03/0.962 & 40.29/0.972 \\
\cmidrule{2-10}          & Noise2VST  \cite{herbreteau2024noise2vst} & \textcolor[rgb]{ 0,  .439,  .753}{\textbf{34.08/0.907}} & 28.34/0.771 & -     & -     & 33.96/0.868 & 36.60/0.929 & 35.99/0.893 & 39.11/0.946 \\
\cmidrule{2-10}          & SASL \cite{li2023spatially}  & 26.52/0.728 & 23.69/0.638 & 38.01/0.936 & -     & 34.93/0.936 & 37.13/0.954 & 38.24/0.964 & 39.44/0.964 \\
\cmidrule{2-10}          & TBSN \cite{li2024rethinking}  & 27.47/0.745 & 24.63/0.662 & 37.79/0.940 & 39.01/0.945 & 35.73/0.931 & 36.51/0.954 & 38.84/0.966 & 40.03/0.966 \\
\cmidrule{2-10}          & Zero-shot  \cite{quan2025zero} & -     & -     & -     & 35.05/0.922 & 37.20/0.948 & 37.88/0.959 & -     & - \\
    \midrule
    \midrule
   \multirow{12}[19]{*}{\shortstack[c]{DNN models \\ (Supervised)}} & ClipDe  \cite{cheng2024transfer} & 30.94/0.871 & 20.97/0.618 & 39.57/0.954 & 39.42/0.956 & 35.40/0.915 & 36.87/0.939 & 37.61/0.954 & 40.03/0.969 \\
\cmidrule{2-10}          & Condformer \cite{huang2024beyond} & -     & -     & \textbf{40.10*/0.956*} & \textbf{40.23*}/- & 36.34/0.922 & 37.27/0.947 & 37.79/0.927 & 39.86/0.956 \\
\cmidrule{2-10}          & DeepSN  \cite{deng2025deepsn}  & -     & -     & 39.92/0.956 & 39.79/0.958 & 35.89/0.936 & 37.25/0.956 & 38.12/0.950 & 39.04/0.966 \\
\cmidrule{2-10}          & DIDN \cite{yu2019deep}  & 27.68/0.802 & 21.85/0.654 & 39.64/0.953 & 39.78/0.958 & 36.06/0.946 & 37.36/0.953 & 38.24/0.950 & 39.86/0.964 \\
\cmidrule{2-10}          & DMID \cite{li2024stimulating}  & -     & -     & -     & -     & 37.09*/- & 37.59/0.946 & 37.90/0.931 & 39.16/0.951 \\
\cmidrule{2-10}          & FFDNet  \cite{zhang2018ffdnet} & 33.57/0.891 & 28.94/0.765 & 37.61/0.942 & 38.27/0.948 & 37.67/0.956 & 38.76/0.970 & 40.28/0.973 & 41.49/0.977 \\
\cmidrule{2-10}          & IDF \cite{kim2025idf}  & 31.88/0.851 & 27.96/0.764 & 33.72/0.811 & -     & 36.43/0.942 & 37.77/0.962 & 38.63/0.960 & 40.23/0.970 \\
\cmidrule{2-10}          & MaIR  \cite{MaIR}  & 32.86/0.873 & 28.61/0.781 & -     & 39.92/0.959 & 36.05/0.938 & 37.64/0.958 & 38.38/0.953 & 40.11/0.969 \\
\cmidrule{2-10}          & NAFNet  \cite{chen2022simple} & -     & -     & 38.36/0.943 & \textcolor[rgb]{ 0,  .439,  .753}{\textbf{40.15/0.960}} & 34.39/0.923 & 36.38/0.947 & 37.88/0.954 & 38.27/0.939 \\
\cmidrule{2-10}          & Restormer \cite{zamir2022restormer} & 31.51/0.845 & 27.31/0.745 & \textcolor[rgb]{ 0,  .439,  .753}{\textbf{40.03/0.956}} & 40.02/\textcolor[rgb]{ 0,  .439,  .753}{\textbf{0.960}} & 36.33/0.941 & 37.66/0.956 & 38.29/0.948 & 40.10/0.966 \\
    \bottomrule
    \end{tabular}}%
 \label{Table_Color_Image_results}%
 \vspace{-11.68pt}
\end{table*}%

\subsubsection{Objective results}
Table \ref{Table_Color_Image_results} evaluates the results of traditional patch-based denoisers and supervised/self-supervised DNN models. Overall, Haar-tSVD achieves competitive performance, its adaptive variant A-Haar-tSVD improves results across datasets, and the enhancement strategy RA-Haar-tSVD further boosts performance under severe noise. On the Kodak dataset, we notice that RA-Haar-tSVD is particularly beneficial when the noise level is high, whereas at lower noise levels, NN-based estimation may suffer from oversmoothness. Therefore, on real-world datasets, we apply the enhancement scheme when $\sigma>30$. Beyond Kodak, A-Haar-tSVD achieves PSNR gains of at least 0.24dB and 0.84dB over CBM3D and MSt-SVD on DND and SIDD, respectively, while RA-Haar-tSVD shows additional improvements in challenging scenarios. On CC, PolyU, HighISO, and IOCI, the proposed methods demonstrate their adaptiveness and generalization compared with many DNN models that degrade when training or validation data are not available.\\
\indent Ideally, the enhancement strategy should be applied only when necessary. To better understand its potential, Table \ref{Table_ideal_enhancement} assesses its effects in cases where measurable gains are observed. Under this ideal setting, RA-Haar-tSVD consistently outperforms the adaptive baseline, with obvious advantages under severe noise, demonstrating its potential in handling different degradations.
\begin{table}[H]
\vspace{-5.18pt}
\scriptsize
  \centering
  \caption{Evaluation of RA-Haar-tSVD under ideal settings.}
  \scalebox{0.918}{
    \begin{tabular}{ccccccc}
    \toprule
    Method & DND   & SIDD  & CC15  & PolyU & HighISO & IOCI \\
    \midrule
    A-Haar-tSVD & 38.25 & 35.58 & 38.24 & 38.89 & 40.62 & 41.52 \\
    \midrule
     RA-Haar-tSVD (ideal) & \textbf{38.64} & \textbf{37.48} & \textbf{38.56} & \textbf{38.95} & \textbf{40.68} & \textbf{41.60} \\
    \midrule
    Improvement & 0.39  & 1.90  & 0.32  & 0.06  & 0.06  & 0.08 \\
    \bottomrule
    \end{tabular}}%
  \label{Table_ideal_enhancement}%
\vspace{-5.18pt}
\end{table}%

\subsubsection{Visual evaluations} 
\begin{figure}[htbp]
\vspace{-1.68pt}
\graphicspath{{Figs/Selected_color_images/Kodak_selected/sigma_40_30_50/Sample2/combined/}}
\centering
\subfigure[Reference]{
\label{Fig4}
\includegraphics[width=0.83018in]{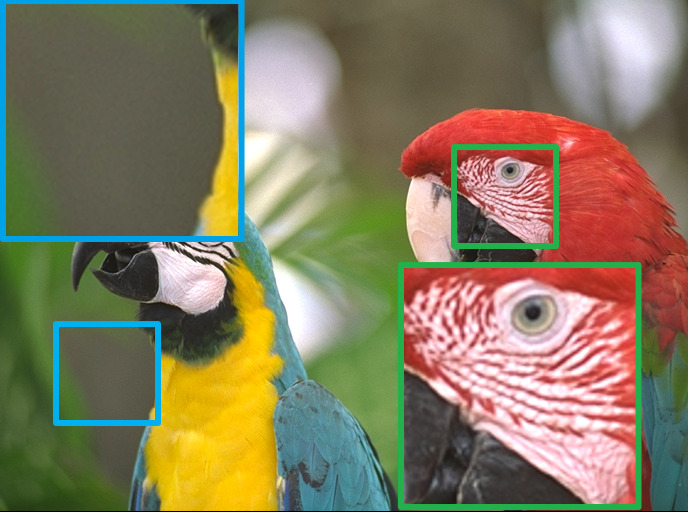}}
\hspace{-0.398em}%
\subfigure[Noisy]{
\label{Fig4}
\includegraphics[width=0.83018in]{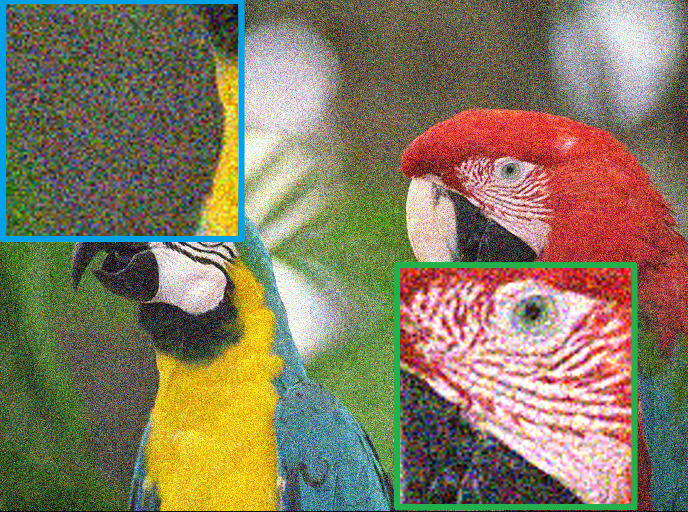}}
\hspace{-0.398em}%
\subfigure[FFDNet]{
\label{Fig4}
\includegraphics[width=0.83018in]{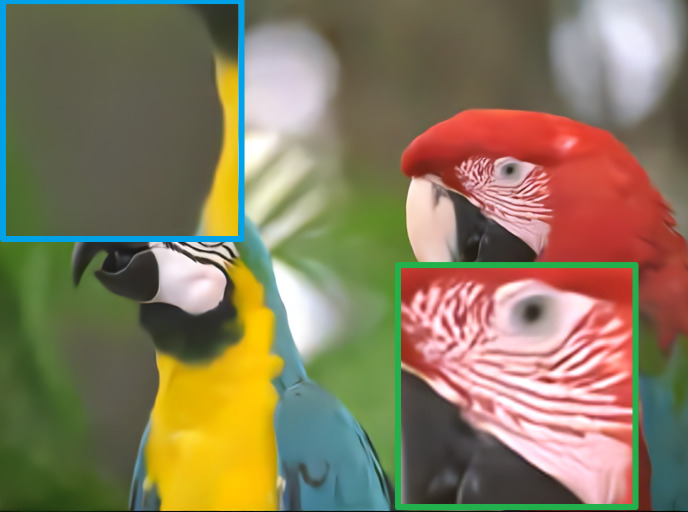}}
\hspace{-0.398em}%
\subfigure[MaIR]{
\label{Fig4}
\includegraphics[width=0.83018in]{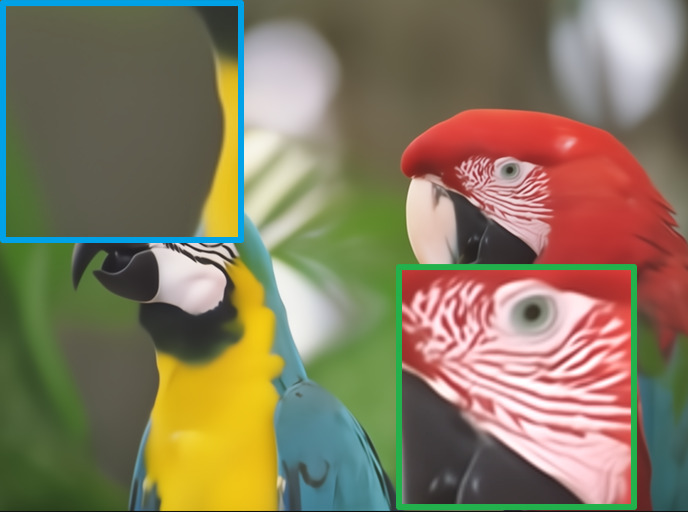}}\\
\vspace{-6.68pt}
\subfigure[Restormer]{
\label{Fig4}
\includegraphics[width=0.83018in]{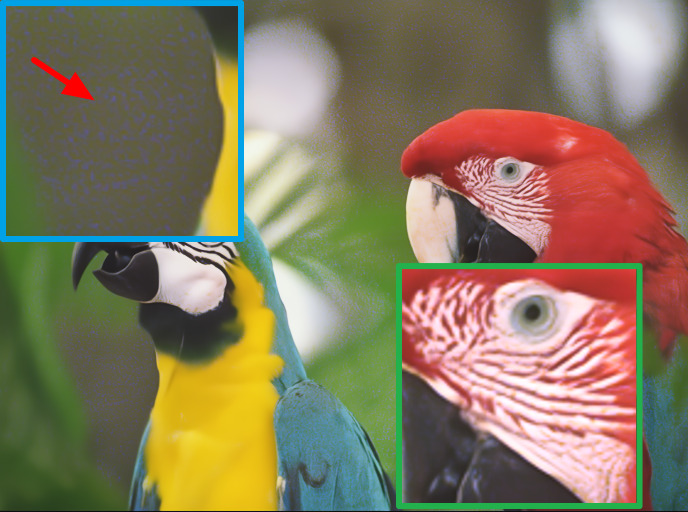}}
\hspace{-0.398em}%
\subfigure[CBM3D]{
\label{Fig4}
\includegraphics[width=0.83018in]{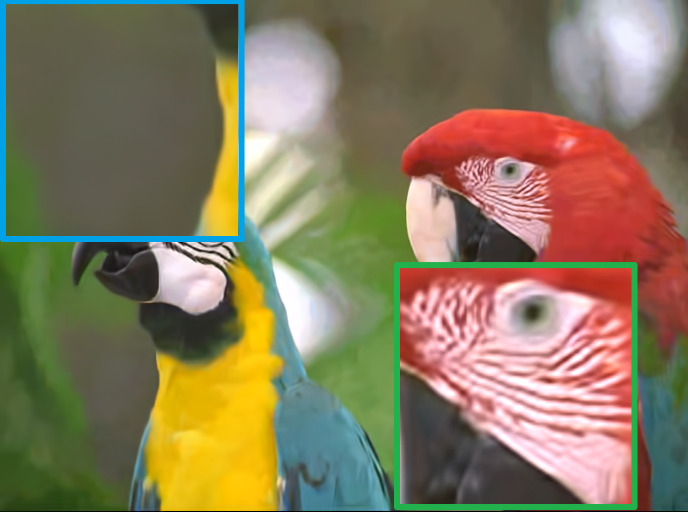}}
\hspace{-0.398em}%
\subfigure[A-Haar-tSVD]{
\label{Fig4}
\includegraphics[width=0.83018in]{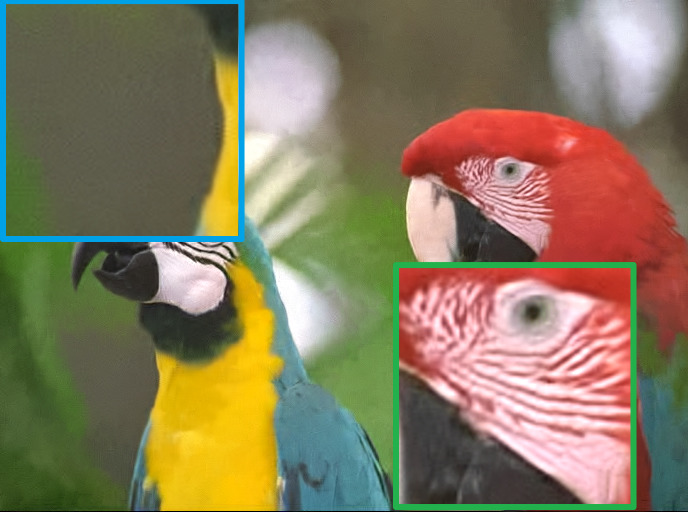}}
\hspace{-0.398em}%
\subfigure[RA-Haar-tSVD]{
\label{Fig4}
\includegraphics[width=0.83018in]{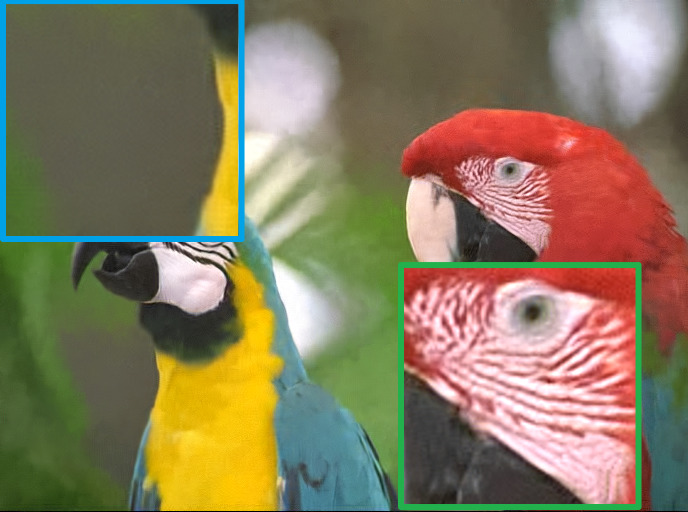}}
\vspace{-9.68pt}
\caption{Denoising comparison on the Kodak (case2) dataset.}
\vspace{-1.8pt}
\label{Fig_Kodak_high_noise}
\end{figure}

\begin{figure}[htbp]
\vspace{-9.18pt}
\graphicspath{{Figs/Selected_color_images/DND/Case3/combined/}}
\centering
\subfigure[Noisy]{
\label{Fig4}
\includegraphics[width=0.83016in]{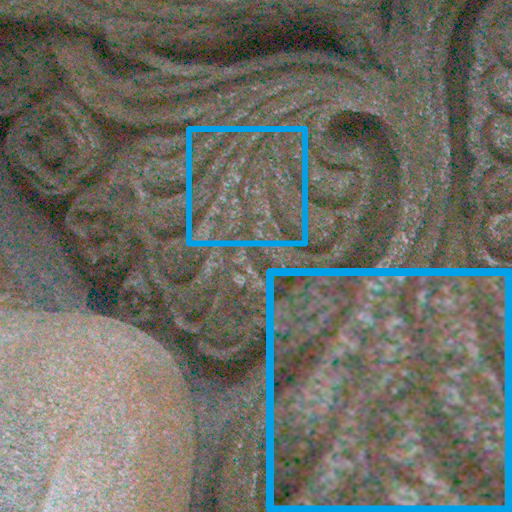}}
\hspace{-0.398em}%
\subfigure[Bitonic]{
\label{Fig4}
\includegraphics[width=0.83016in]{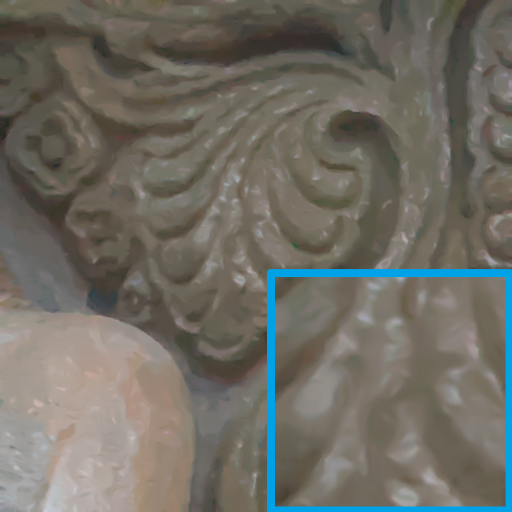}}
\hspace{-0.398em}%
\subfigure[NLHCC]{
\label{Fig4}
\includegraphics[width=0.83016in]{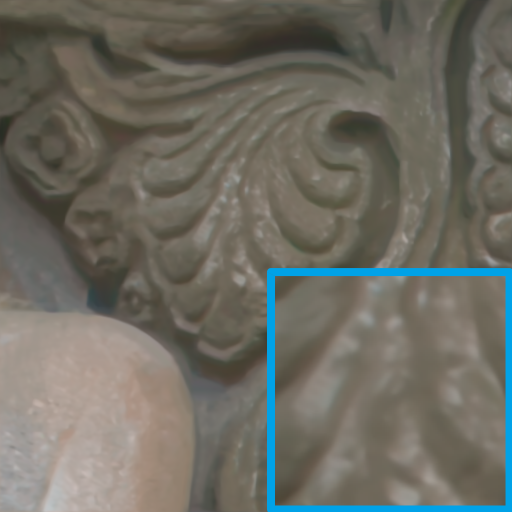}}
\hspace{-0.398em}%
\subfigure[SASL]{
\label{Fig4}
\includegraphics[width=0.83016in]{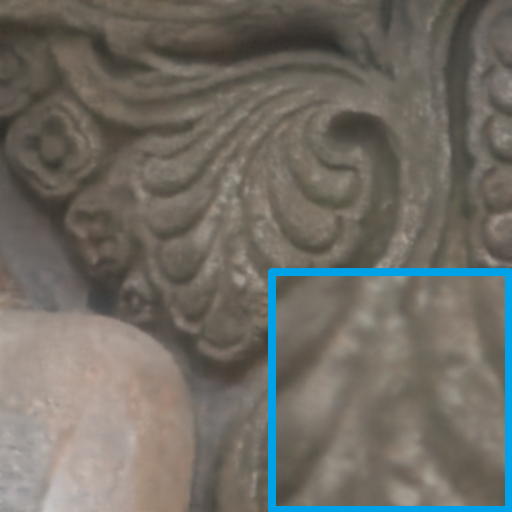}}\\
\vspace{-5.18pt}
\subfigure[DeepSN]{
\label{Fig4}
\includegraphics[width=0.83016in]{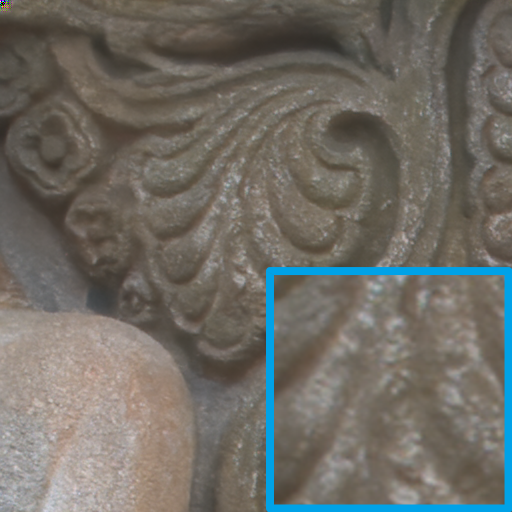}}
\hspace{-0.398em}%
\subfigure[Restormer]{
\label{Fig4}
\includegraphics[width=0.83016in]{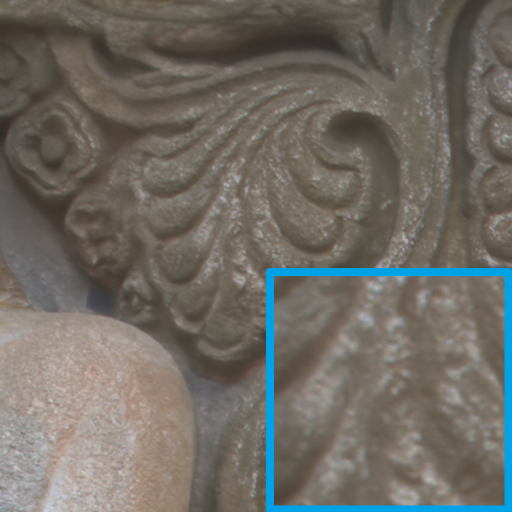}}
\hspace{-0.398em}%
\subfigure[NAFNet]{
\label{Fig4}
\includegraphics[width=0.83016in]{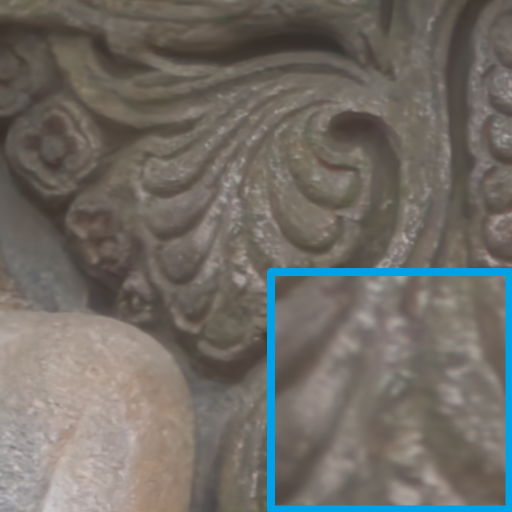}}
\hspace{-0.398em}%
\subfigure[RA-Haar-tSVD]{
\label{Fig4}
\includegraphics[width=0.83016in]{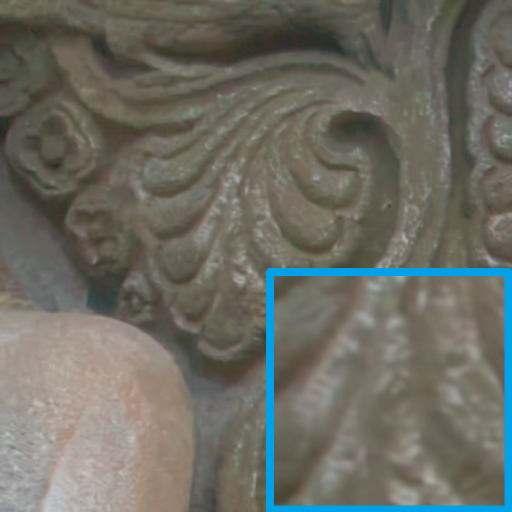}}
\vspace{-8.18pt}
\caption{Denoising comparison on the DND dataset.}
\vspace{-0.08pt}
\label{Fig_compare_with_DND}
\end{figure}

\begin{figure}[htbp]
\vspace{-1.68pt}
\graphicspath{{Figs/Selected_color_images/IOCI/Case5_new/Combined/}}
\centering
\subfigure[Mean]{
\label{Fig4}
\includegraphics[width=0.80016in]{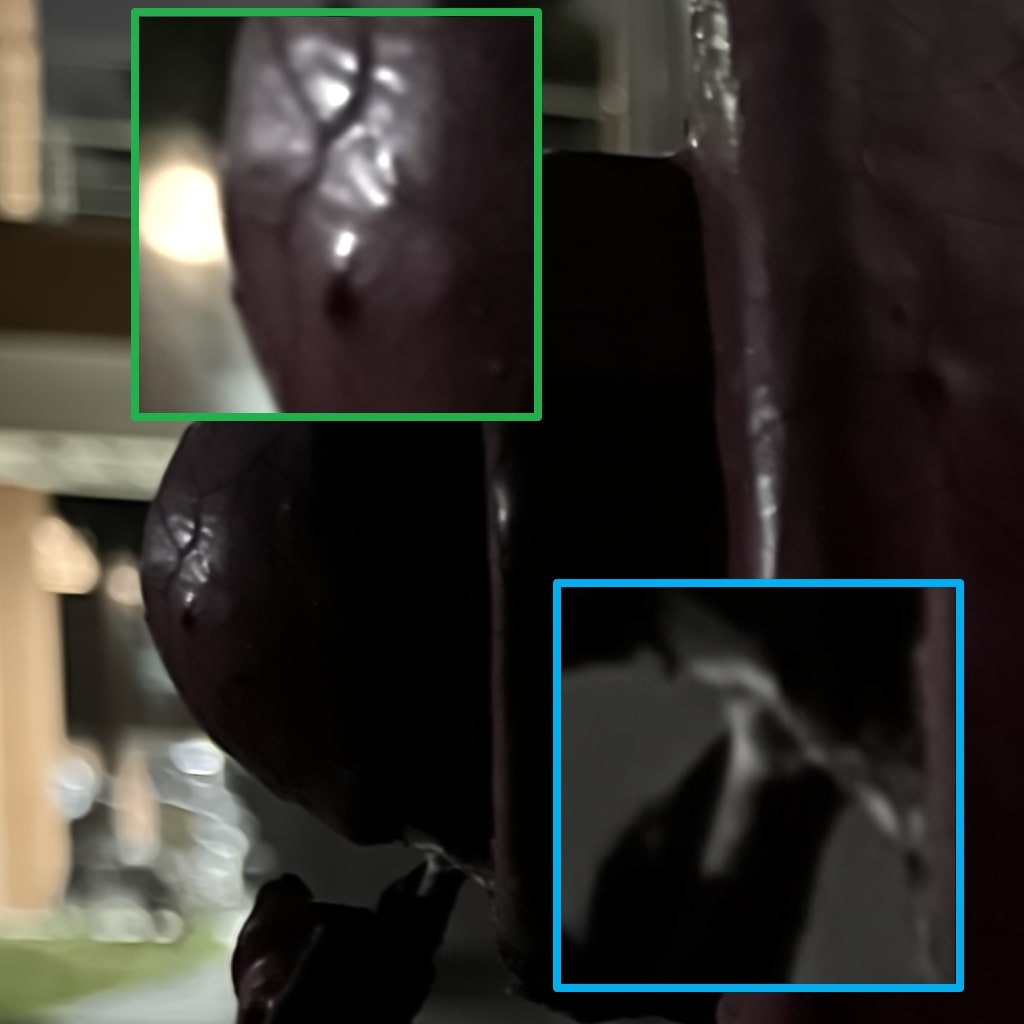}}
\hspace{-0.168em}%
\subfigure[Noisy]{
\label{Fig4}
\includegraphics[width=0.80016in]{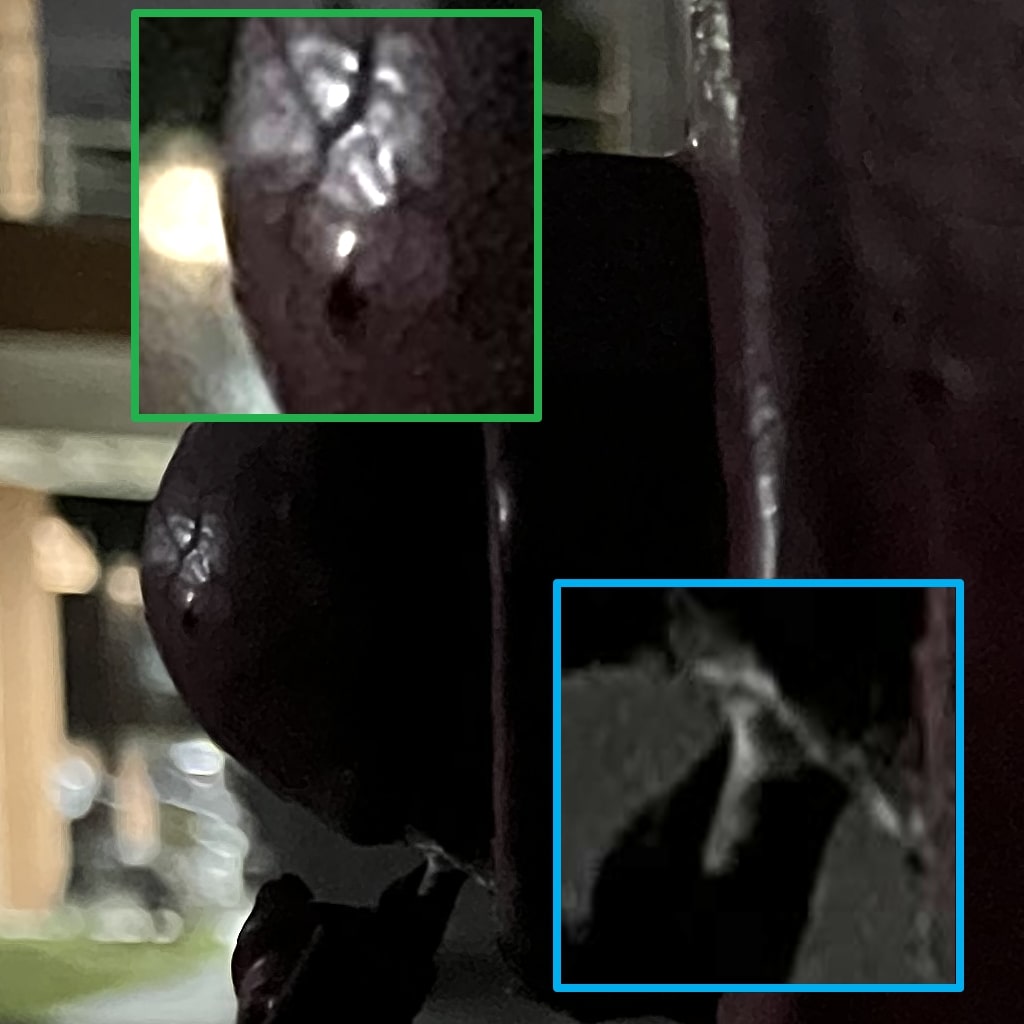}}
\hspace{-0.168em}%
\subfigure[Condformer]{
\label{Fig4}
\includegraphics[width=0.80016in]{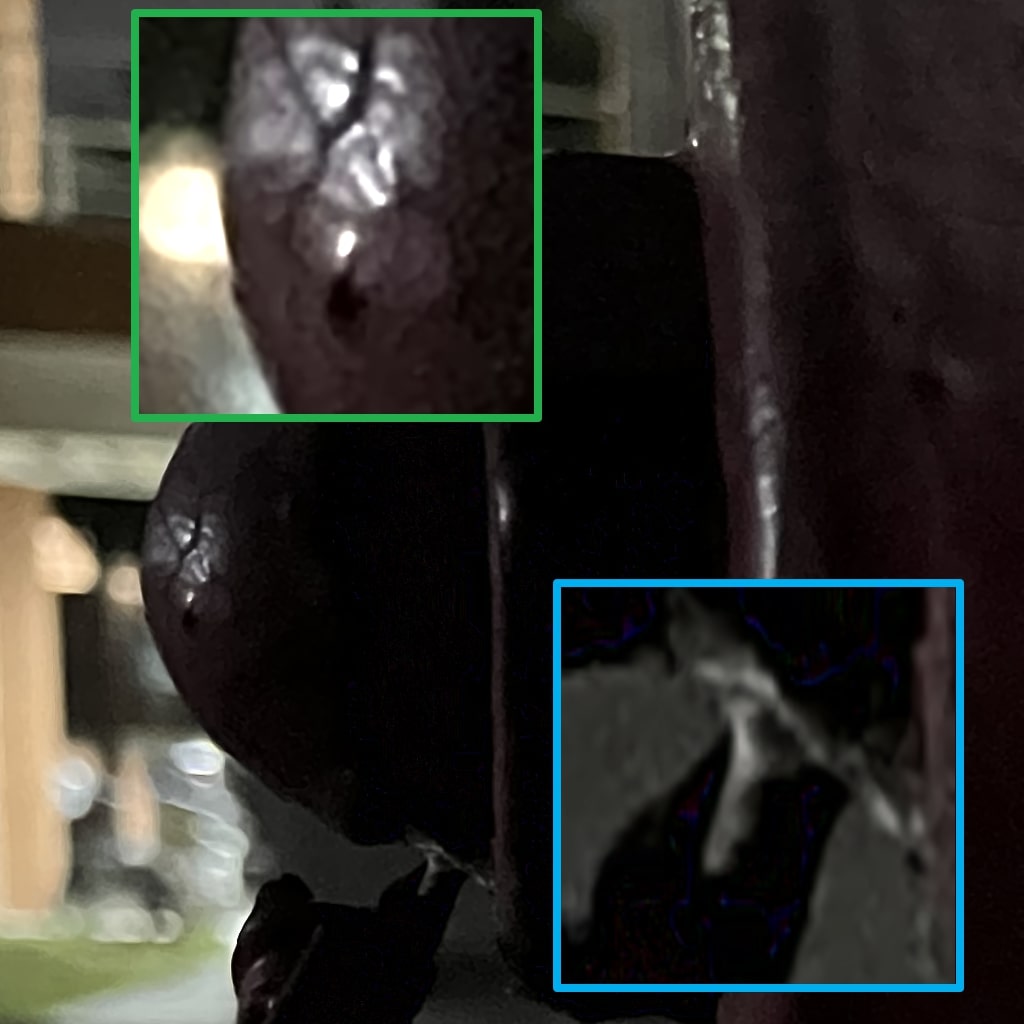}}
\hspace{-0.168em}%
\subfigure[DeepSN]{
\label{Fig4}
\includegraphics[width=0.80016in]{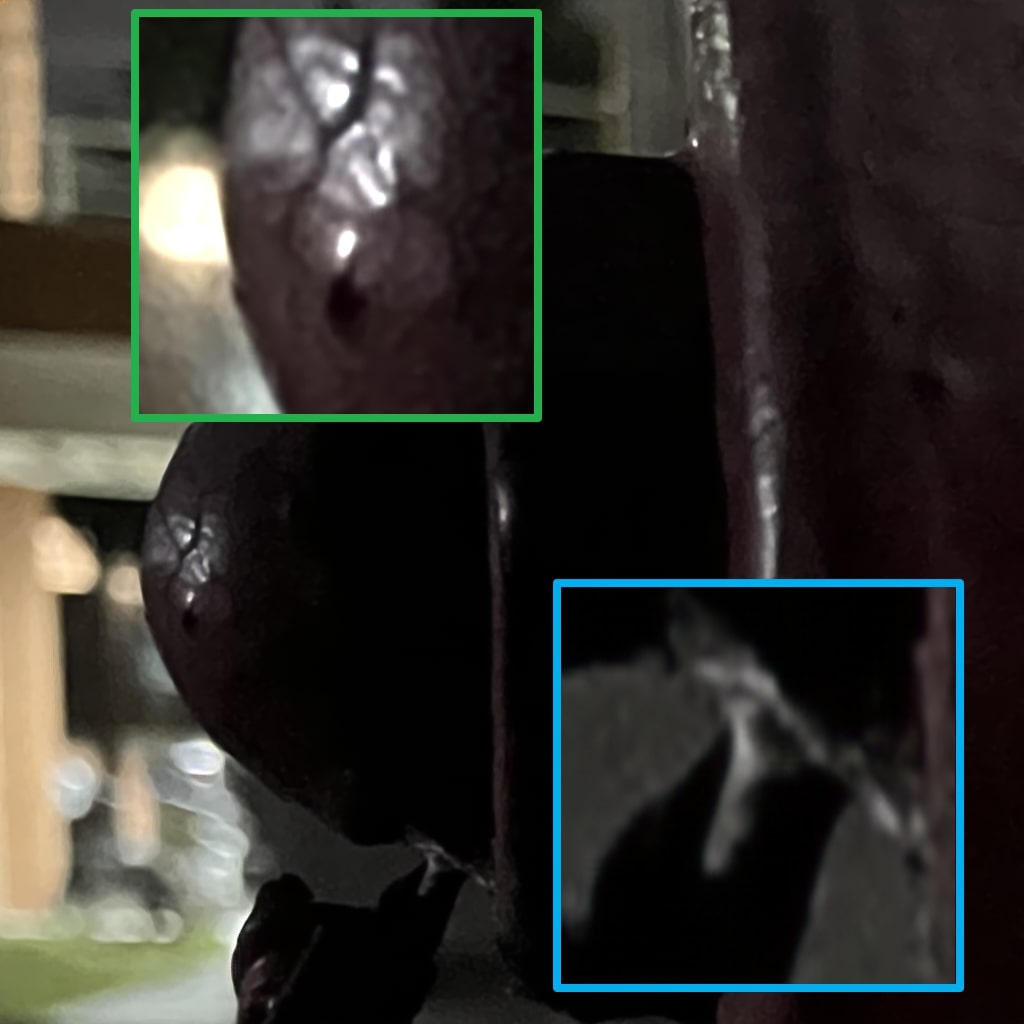}}
\hspace{-0.168em}%
\vspace{-3.18pt}
\subfigure[MaIR]{
\label{Fig4}
\includegraphics[width=0.80016in]{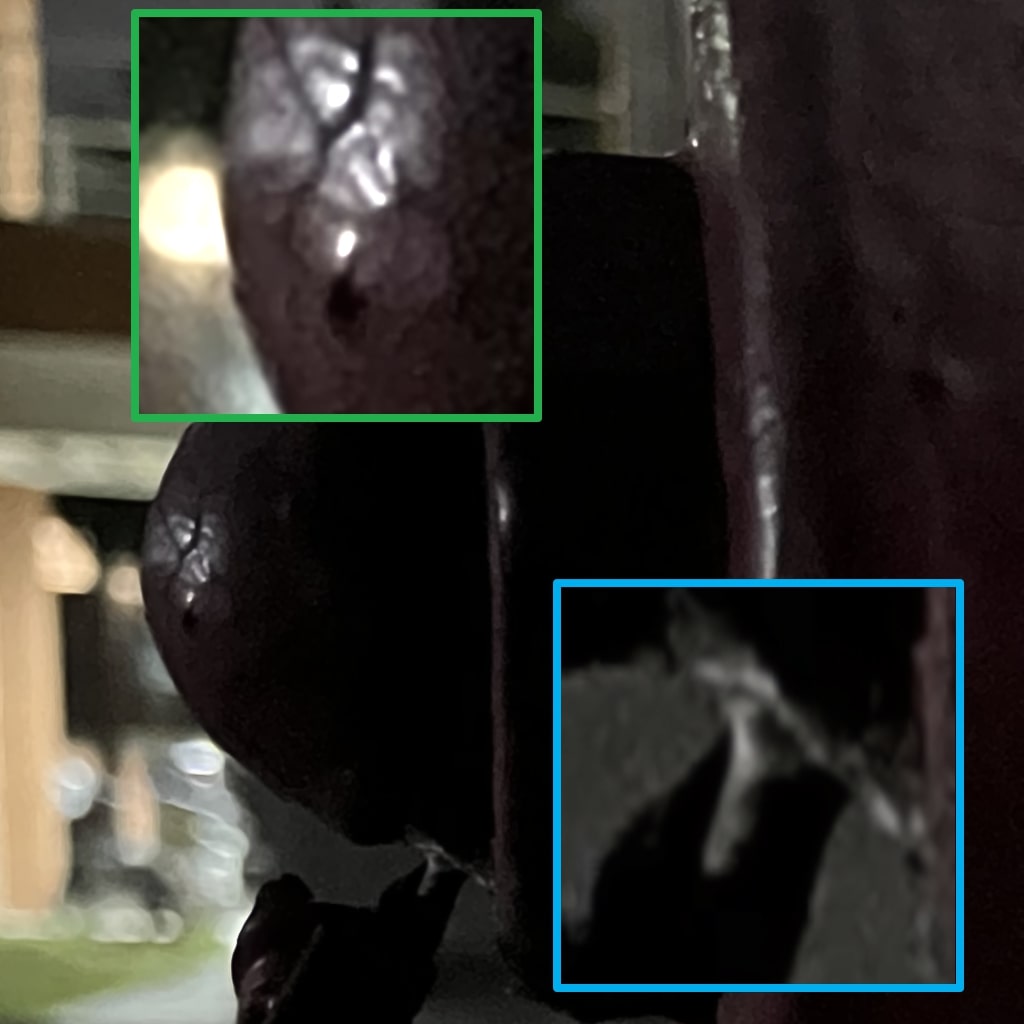}}
\hspace{-0.168em}%
\subfigure[Noise2VST]{
\label{Fig4}
\includegraphics[width=0.80016in]{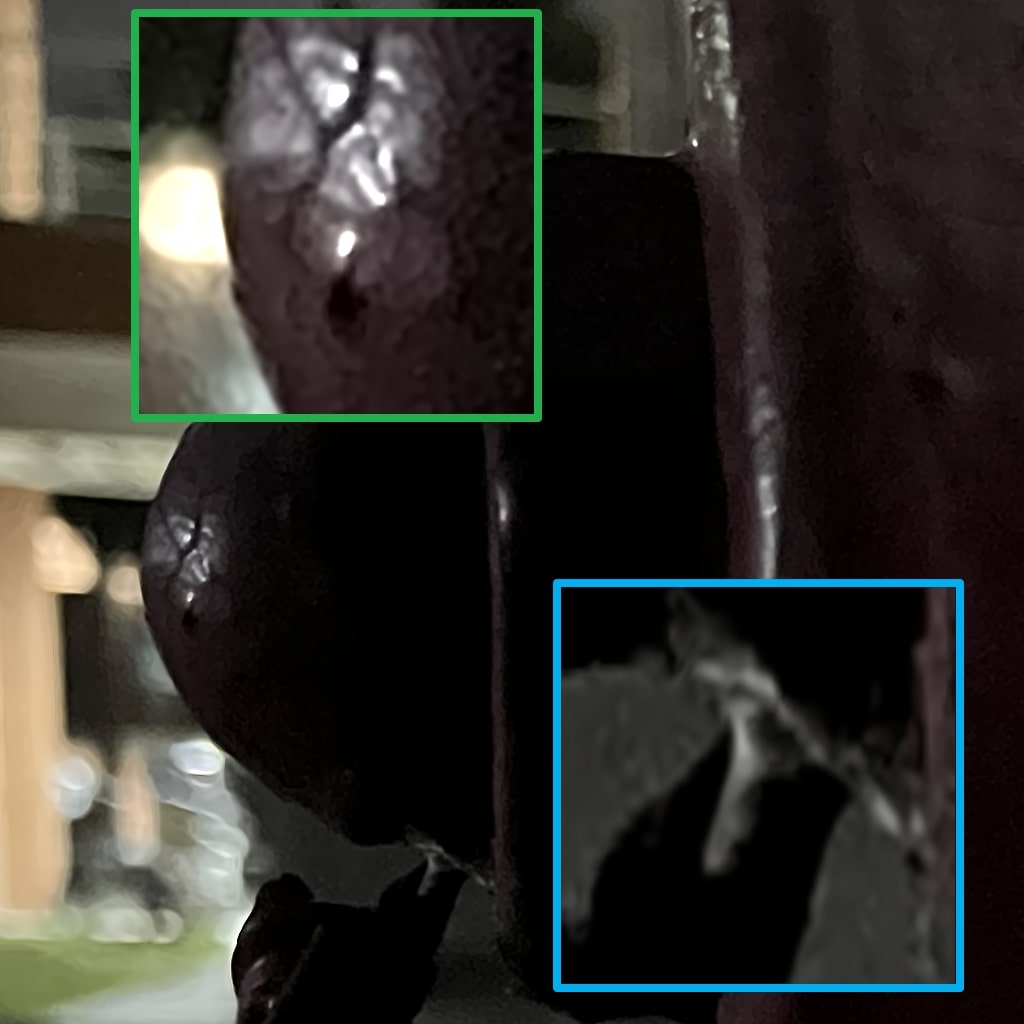}}
\hspace{-0.168em}%
\subfigure[Restormer]{
\label{Fig4}
\includegraphics[width=0.80016in]{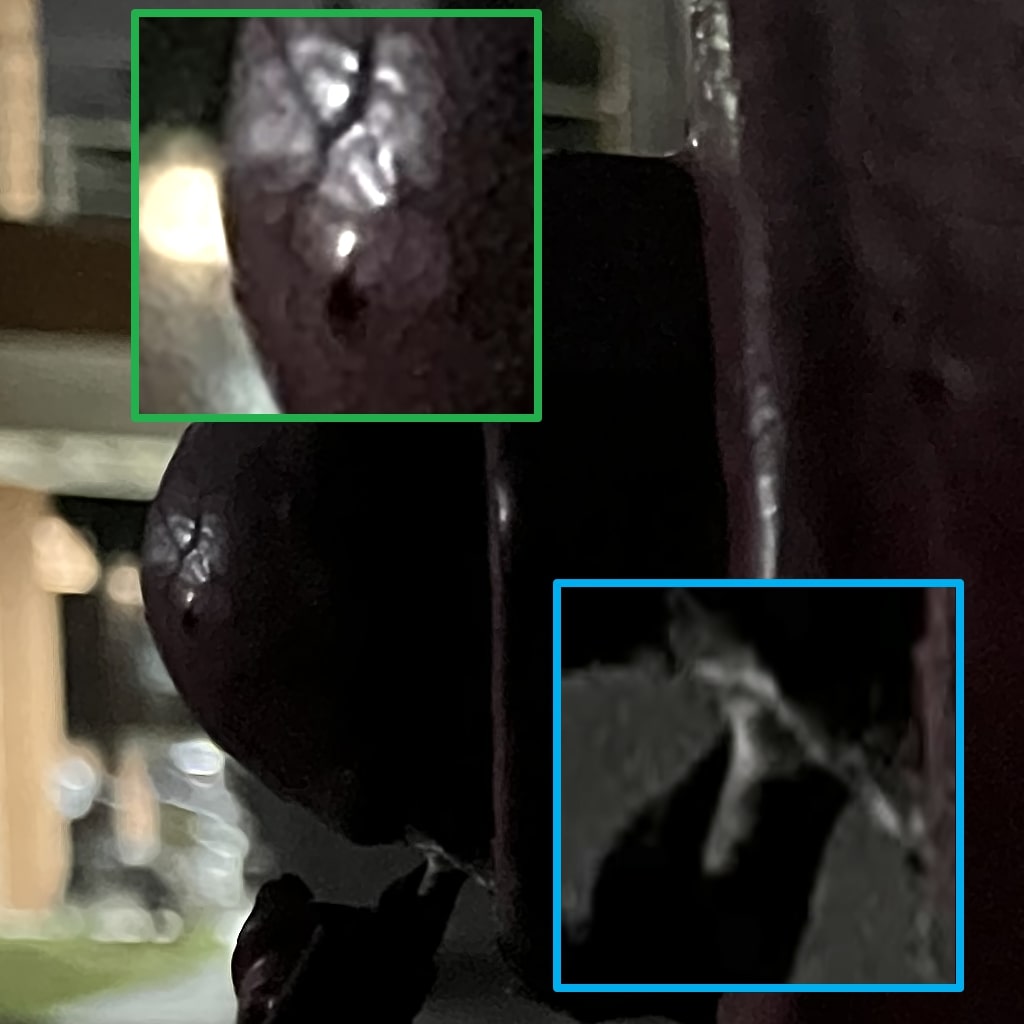}}
\hspace{-0.168em}%
\subfigure[A-Haar-tSVD]{
\label{Fig4}
\includegraphics[width=0.80016in]{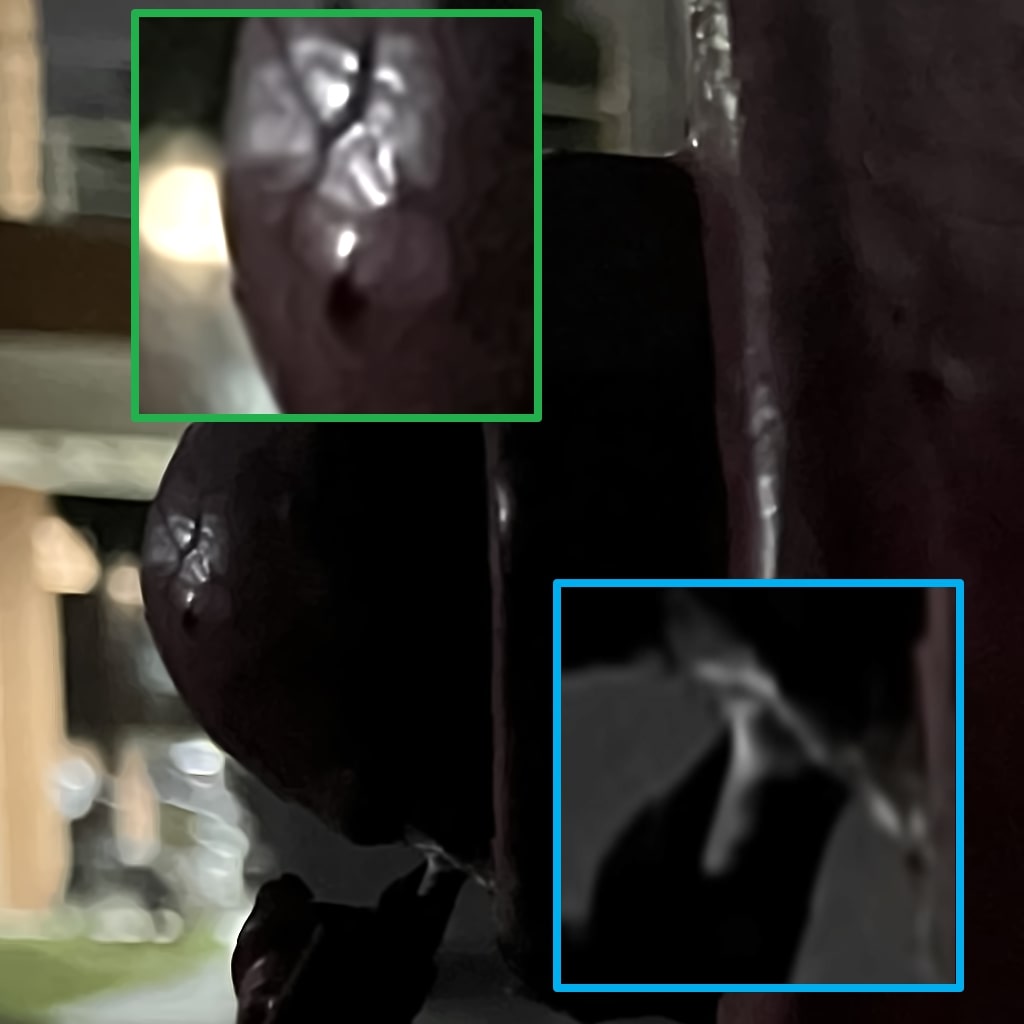}}
\vspace{-6.98pt}
\caption{Denoising comparison on the IOCI datset.}
\label{Fig_compare_with_IOCI_IPHONE13_new}
\vspace{-0.18pt}
\end{figure}

Visual evaluations are presented in Fig. \ref{Fig_Kodak_high_noise} to Fig. \ref{Fig_compare_with_IOCI_IPHONE13_new}. Specifically, Fig. \ref{Fig_Kodak_high_noise} shows the effectiveness of integrating NNs with circulant representation for noise removal and detail preservation, while compared methods slightly over-smooth textures in certain regions. Fig. \ref{Fig_compare_with_DND} highlights the limitations of patch-based denoisers when handling heavily corrupted images, where noise disrupts grouping and local transforms. Nevertheless, the proposed method produces less distortions and color artifacts compared to SASL and Bitonic, thanks to the redundancy encoded in the global-local circulant structures and the enhancement strategy. Fig. \ref{Fig_compare_with_IOCI_IPHONE13_new} shows that when confronted with unseen noise patterns, the well-trained DNN models may leave unwanted artifacts and suffer from over-smooth effects. By comparison, the adaptive scheme shows its strengths by exploiting both the CNN estimator and nonlocal information, which render certain robustness and adaptability, therefore achieving a balance between noise suppression and detail recovery. 
%


\subsubsection{Denoising efficiency} As reported in Table \ref{Table_Color_Image_time}, supervised DNN models benefit from modern GPUs and achieve fast inference. For example, Restormer and Condformer can process an image of size $512\times 512 \times 3$ within 1 second. The proposed Haar-tSVD is among the few methods parallel with the state-of-the-art BM3D in terms of both efficiency and effectiveness, as its grouping is performed only on the green/opponent channel and it avoids recursive learning of local transforms. Moreover, compared with other complex network architectures, RA-Haar-tSVD is lightweight and enjoys considerably low complexity. In particular, the simple CNN noise estimator and the FCN used in the adaptive and enhancement strategies require less than $\frac{1}{50}$ of the training time of an advanced model like Restormer. This combination of channel-wise prior, lightweight design, and avoidance of repeated local learning makes the proposed method highly practical for real-world denoising applications.
\begin{table}[htbp]
\vspace{-8.18pt}
\scriptsize
\centering
\setlength{\tabcolsep}{3.2pt} 
\caption{Computational complexity of different denoising methods when processing $512 \times 512 \times 3$ sRGB images.}
\scalebox{0.938}{
\begin{tabular}{@{}ccccccc@{}}
\toprule
Method & CBM3D & NLHCC & RA-Haar-tSVD & DIDN & Condformer & Restormer \\
\midrule
Test time (s)  & 3.6 & 41.2 & 4.5 & 7.3 & 0.9 & \textbf{0.8} \\
\midrule
Train time (m) & --  & --   & \textbf{23.2} & --  & --  & $>1000$ \\
\midrule
\# Params (M)  & --  & --   & \textbf{7.5} & 22.1 & 26.4 & 25.3 \\
\bottomrule
\end{tabular}}
\vspace{-5.18pt}
\label{Table_Color_Image_time}
\vspace{-9.18pt}
\end{table}

\subsection{Real-world Color Video Denoising}

\begin{table*}[htbp]
\vspace{-6.8pt}
  \centering
  \vspace{-3.8pt}
  \caption{Quantitative comparison on real-world color video denoising datasets. Best results are highlighted in bold.}
  \scriptsize
  \setlength{\tabcolsep}{3.518pt} 
  \scalebox{0.968}{
  \begin{tabular}{c cccc ccc ccccccc}
    \toprule
    \multirow{2}{*}{Dataset} 
      & \multicolumn{4}{c}{Traditional denoisers} 
      & \multicolumn{3}{c}{Traditional + DNN} 
      & \multicolumn{7}{c}{DNN models} \\
    \cmidrule(lr){2-5}\cmidrule(lr){6-8}\cmidrule(lr){9-15}
    & MSt-SVD & VBM4D & VIDOSAT & Haar-tSVD
      & A-Haar-tSVD & RA-Haar-tSVD & VNLNet
      & DVDNet & FastDVDNet & FloRNN & STBN & UDVD & ViDeNN & VRT \\
    & \cite{kong2019color} & \cite{maggioni2012video} & \cite{wen2018vidosat} & (Ours)
      & (Ours) & (Ours) & \cite{davy2019non}
      & \cite{tassano2019dvdnet} & \cite{Tassano_2020_CVPR} & \cite{li2022unidirectional}
      & \cite{chen2025spatiotemporal} & \cite{sheth2021unsupervised}
      & \cite{claus2019videnn} & \cite{liang2024vrt} \\
    \midrule
    \multirow{2}{*}{CRVD \cite{yue2020supervised}}
      & 36.66 & 34.14 & 34.16 & 36.80
      & 37.00 & \textbf{37.13} & 36.11
      & 34.50 & 35.84 & 36.66 & 36.51 & - & 32.31 & 36.94 \\
    & 0.946 & 0.908 & 0.938 & 0.953
      & 0.961 & \textbf{0.963} & 0.945
      & 0.949 & 0.931 & 0.960 & 0.960 & - & 0.845 & 0.956 \\
    \midrule
    \multirow{2}{*}{IOCV \cite{kong2023comparison}}
      & 38.22 & 38.76 & - & 38.83
      & 38.92 & \textbf{39.04} & 38.76
      & 38.53 & 37.57 & 38.64 & 38.76 & 35.02 & 36.13 & 38.50 \\
    & 0.974 & 0.976 & - & 0.976
      & 0.977 & \textbf{0.978} & 0.977
      & 0.975 & 0.970 & 0.974 & 0.977 & 0.966 & 0.951 & 0.967 \\
    \bottomrule
  \end{tabular}}
  \label{Table_video_denoising_results}
  \vspace{-8.8pt}
\end{table*}

\begin{figure}[htbp]
\vspace{-3.88pt}
\graphicspath{{Figs/Selected_color_images/IOCV/Case2/Combined_new/}}
\centering
\subfigure[Mean]{
\label{Fig4}
\includegraphics[width=0.80016in]{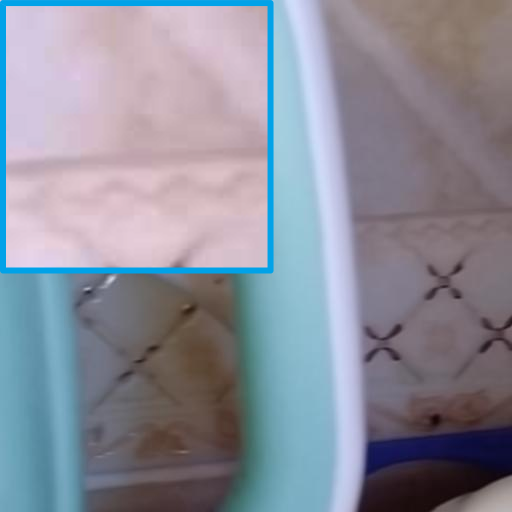}} 
\hspace{-0.168em}%
\subfigure[Noisy]{
\label{Fig4}
\includegraphics[width=0.80016in]{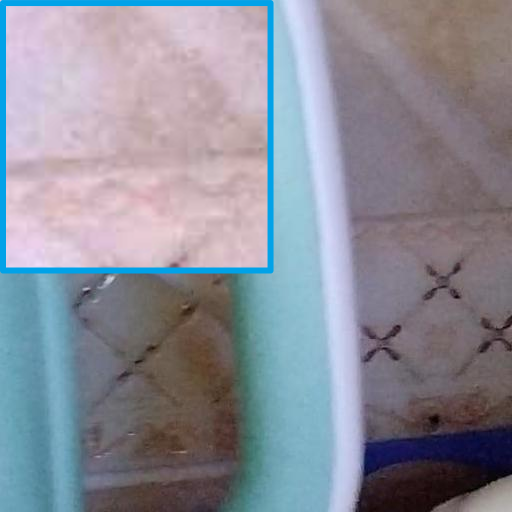}}
\hspace{-0.168em}%
\subfigure[VBM4D]{
\label{Fig4}
\includegraphics[width=0.80016in]{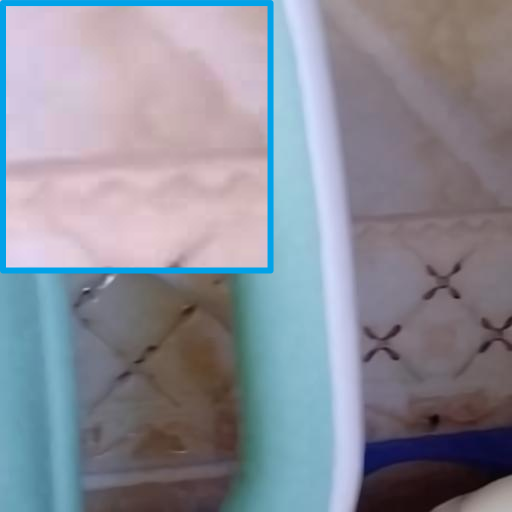}}
\hspace{-0.168em}%
\subfigure[FastDVDNet]{
\label{Fig4}
\includegraphics[width=0.80016in]{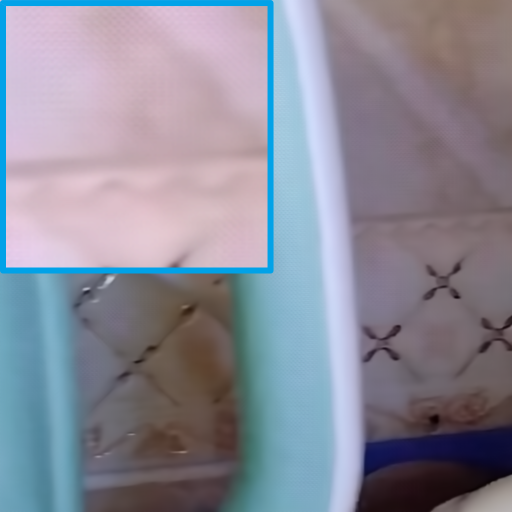}}\\
\vspace{-5.18pt}
\subfigure[FloRNN]{
\label{Fig4}
\includegraphics[width=0.80016in]{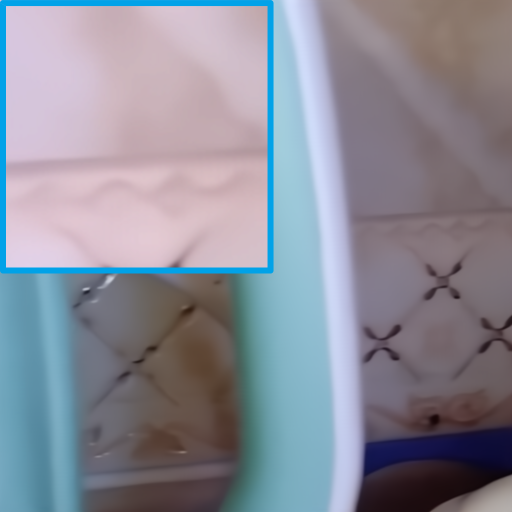}}
\hspace{-0.168em}%
\subfigure[VRT]{
\label{Fig4}
\includegraphics[width=0.80016in]{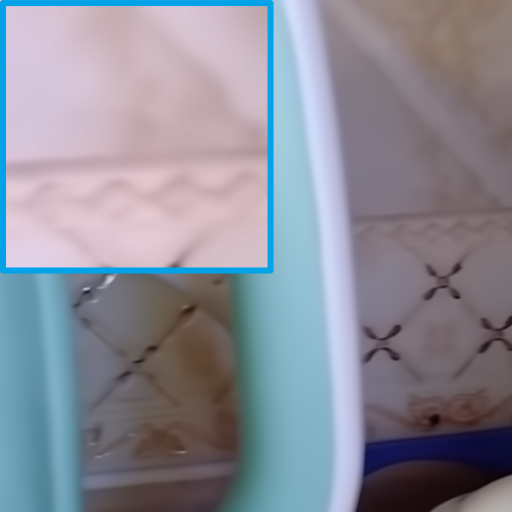}}
\hspace{-0.168em}%
\subfigure[VNLNet]{
\label{Fig4}
\includegraphics[width=0.80016in]{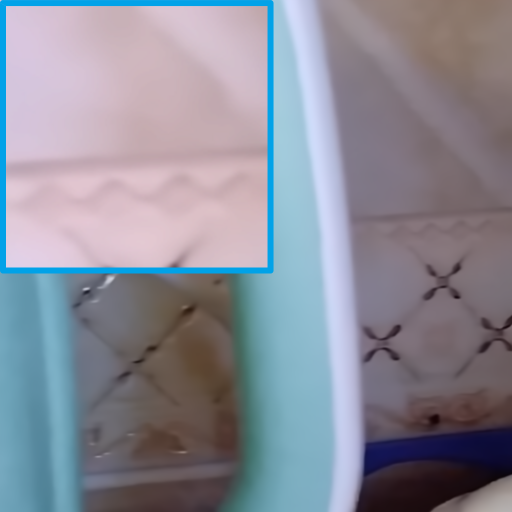}}
\hspace{-0.168em}%
\subfigure[A-Haar-tSVD]{
\label{Fig4}
\includegraphics[width=0.80016in]{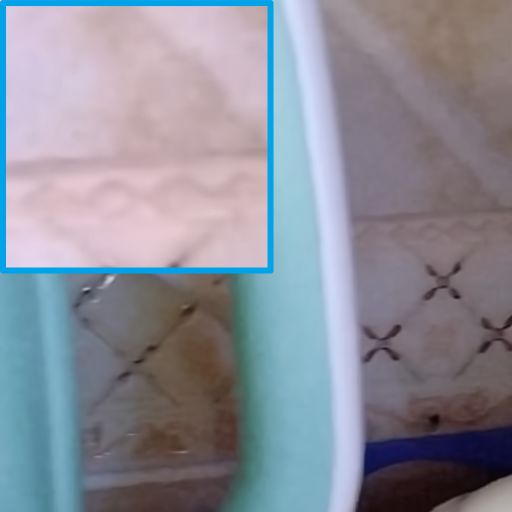}}
\vspace{-5.98pt}
\caption{Denoising comparison on the IOCV dataset.}
\vspace{-1.98pt}
\label{Fig_compare_with_IOCV_case2}
\end{figure}

\begin{figure}[htbp]
\vspace{-2.18pt}
\graphicspath{{Figs/Selected_color_images/CRVD/Case9/Combined_new/}}
\centering
\subfigure[Mean]{
\label{Fig4}
\includegraphics[width=0.80016in]{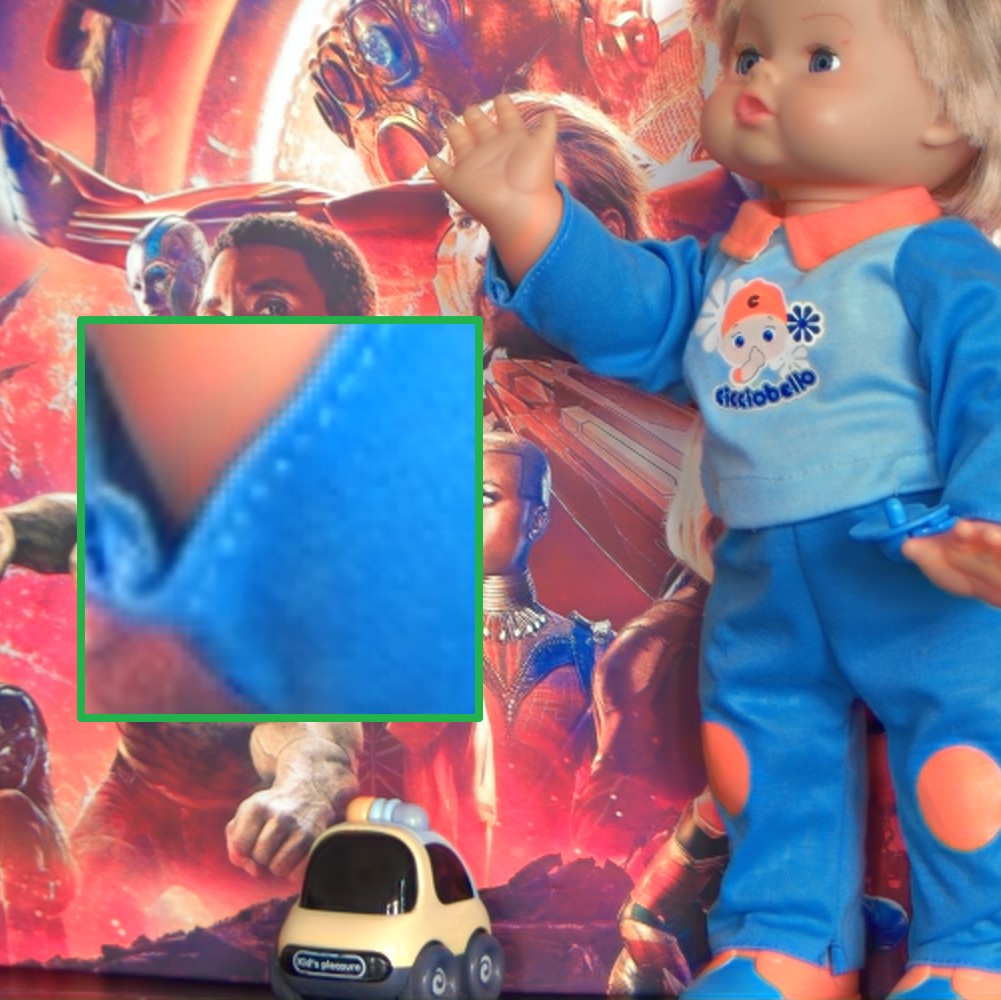}} 
\hspace{-0.168em}%
\subfigure[Noisy]{
\label{Fig4}
\includegraphics[width=0.80016in]{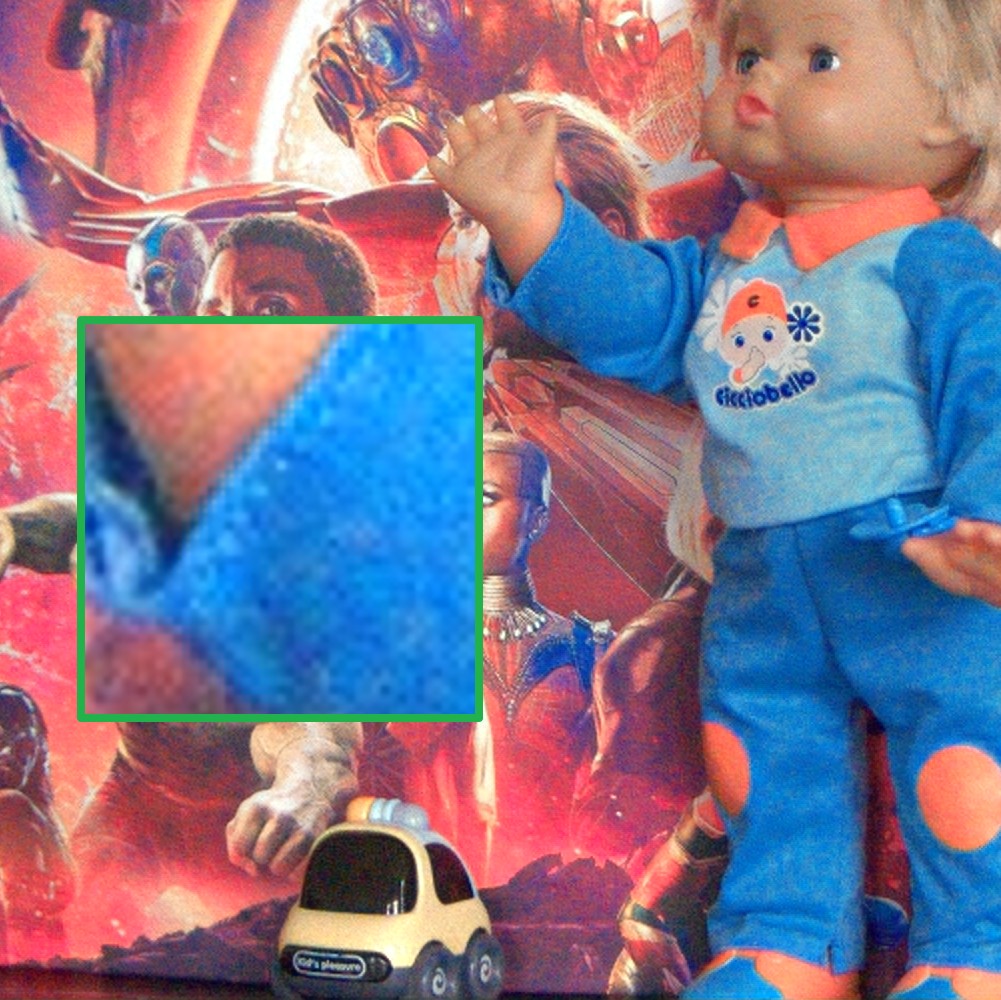}}
\hspace{-0.168em}%
\subfigure[VBM4D]{
\label{Fig4}
\includegraphics[width=0.80016in]{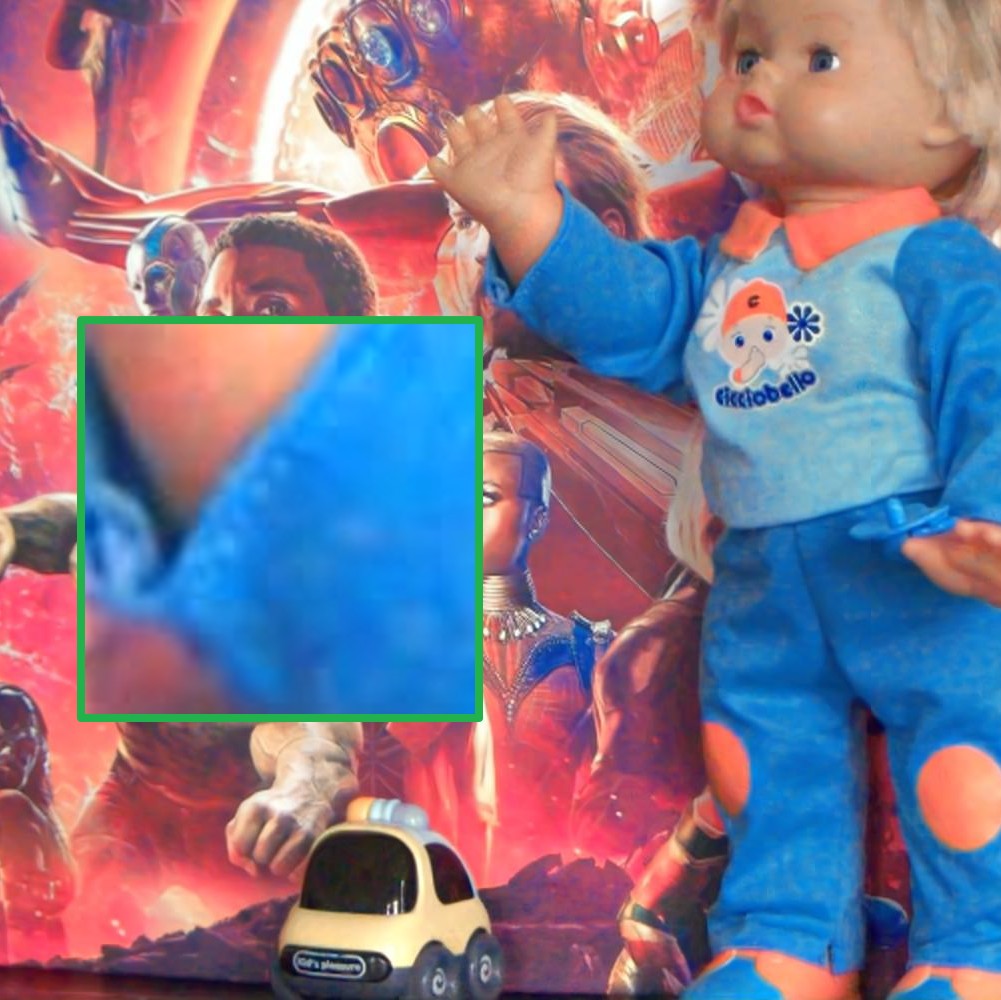}}
\hspace{-0.168em}%
\subfigure[FastDVDNet]{
\label{Fig4}
\includegraphics[width=0.80016in]{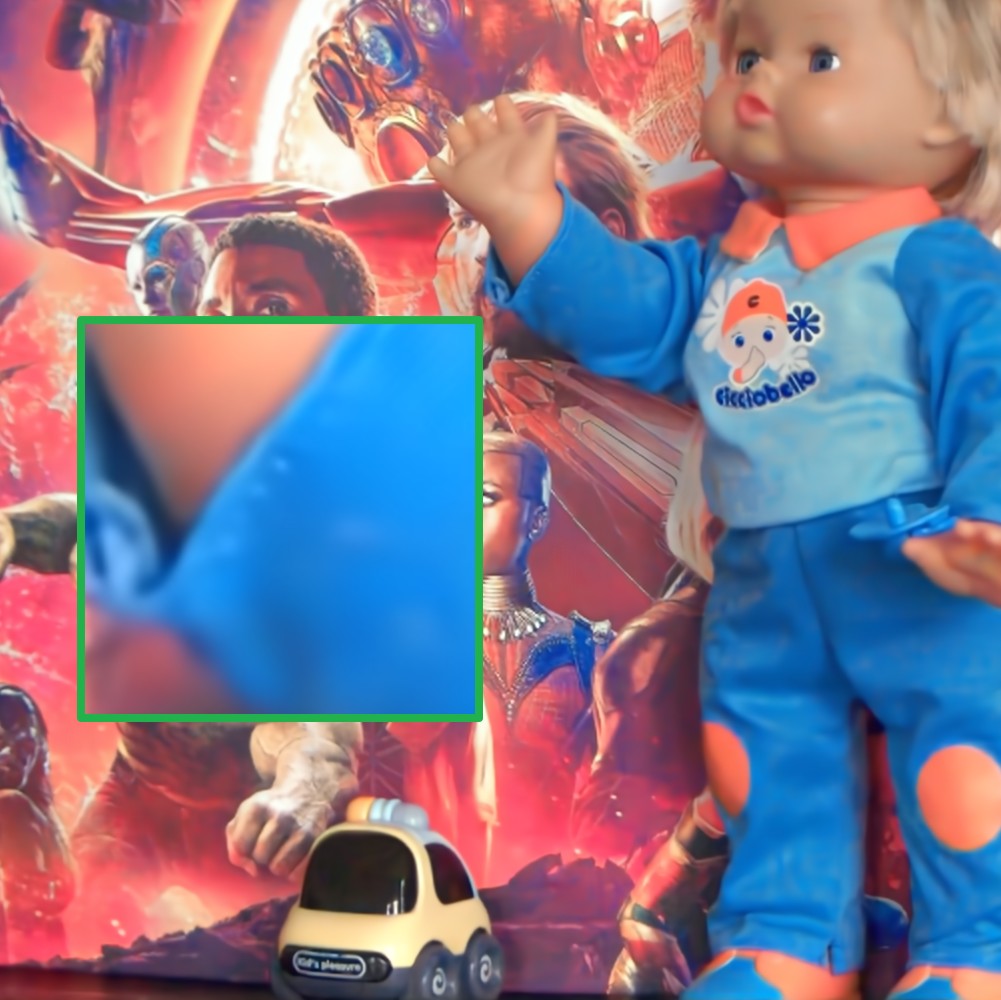}}\\
\vspace{-8.18pt}
\subfigure[FloRNN]{
\label{Fig4}
\includegraphics[width=0.80016in]{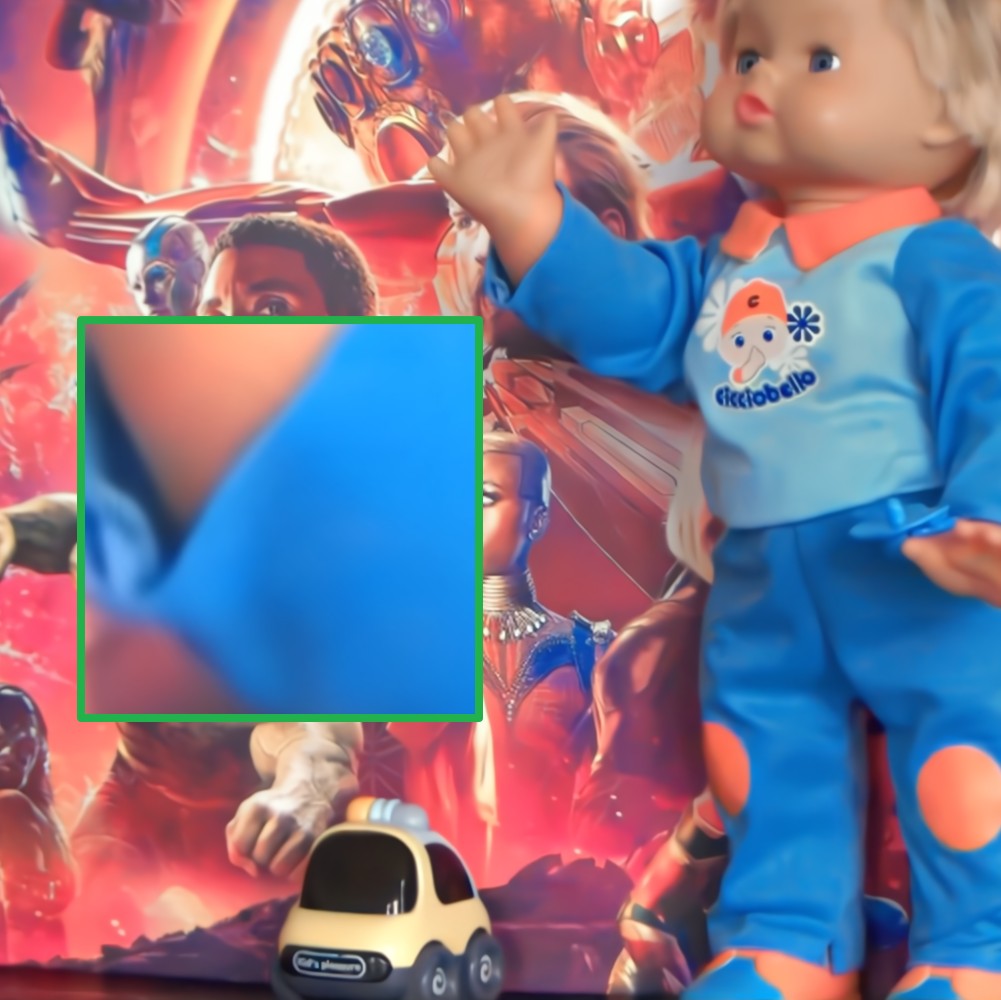}}
\hspace{-0.168em}%
\subfigure[VRT]{
\label{Fig4}
\includegraphics[width=0.80016in]{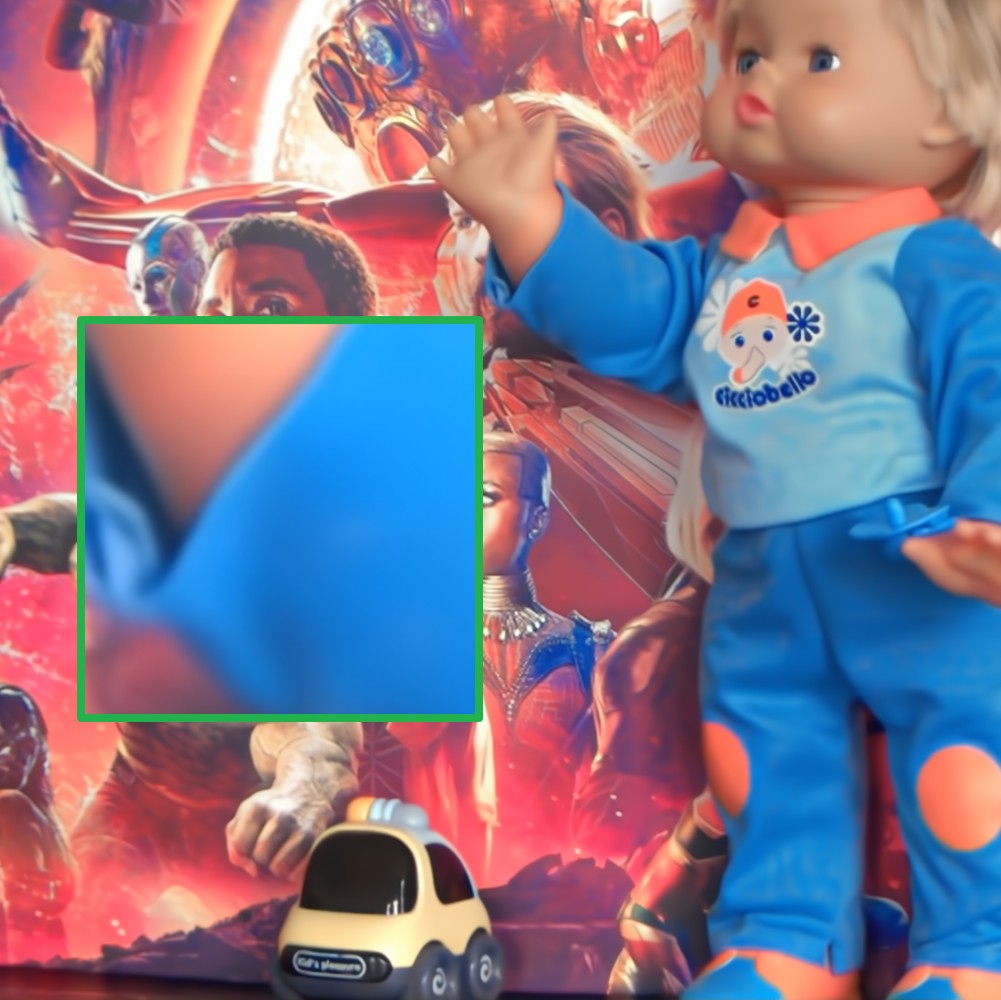}}
\hspace{-0.168em}%
\subfigure[VNLNet]{
\label{Fig4}
\includegraphics[width=0.80016in]{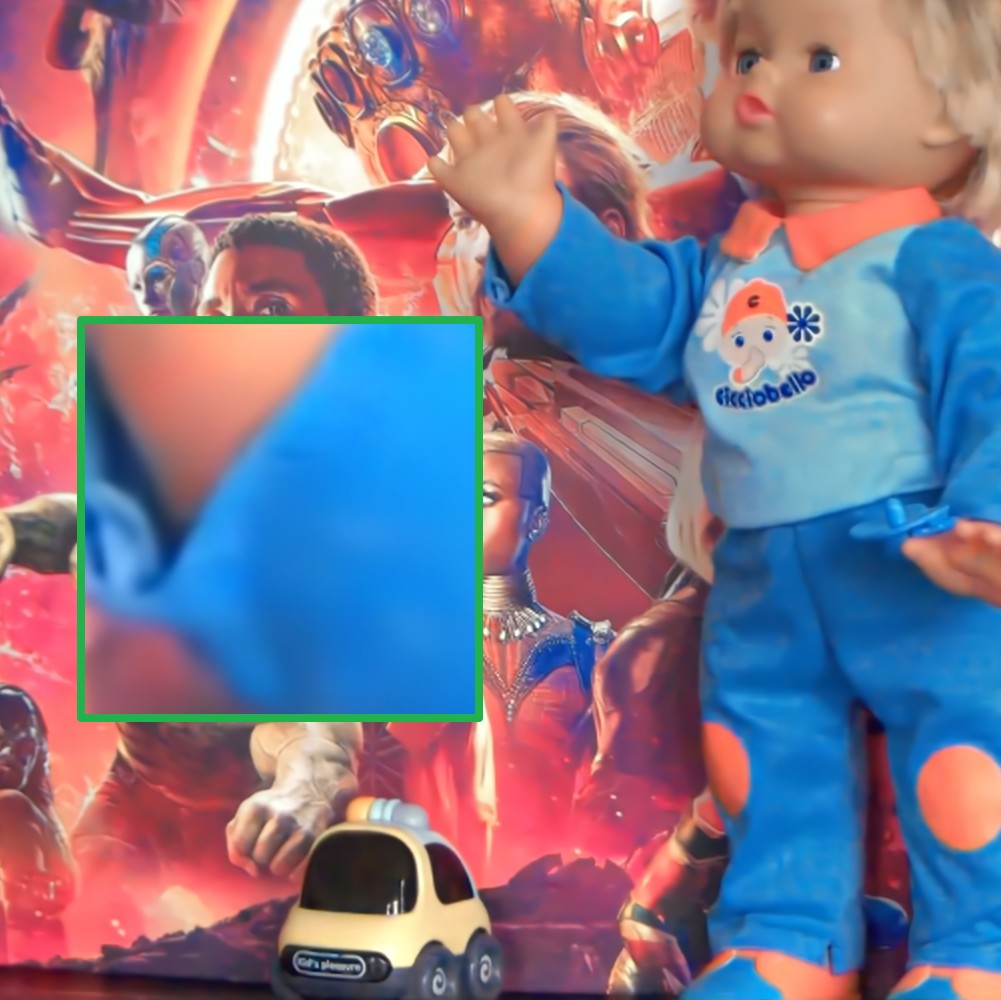}}
\hspace{-0.168em}%
\subfigure[A-Haar-tSVD]{
\label{Fig4}
\includegraphics[width=0.80016in]{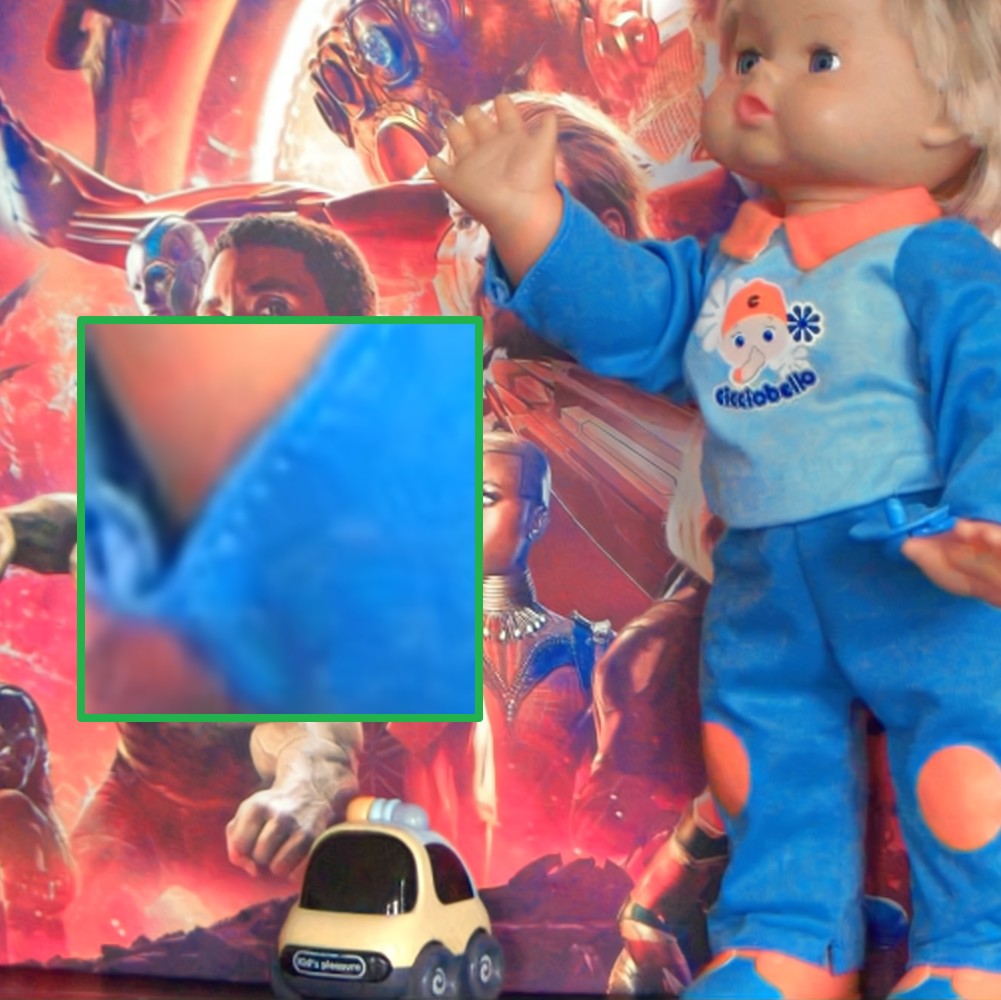}} 
\vspace{-6.98pt}
\caption{Denoising comparison on the CRVD dataset (ISO = 6400).}
\label{Fig_compare_with_CRVD_case9}
\vspace{-0.18pt}
\end{figure}

\begin{figure}[htbp]
\vspace{-13.68pt}
\graphicspath{{Figs/Selected_color_images/CRVD/Sample5/combined_new/}}
\centering
\subfigure[Mean]{
\label{Fig4}
\includegraphics[width=0.83016in]{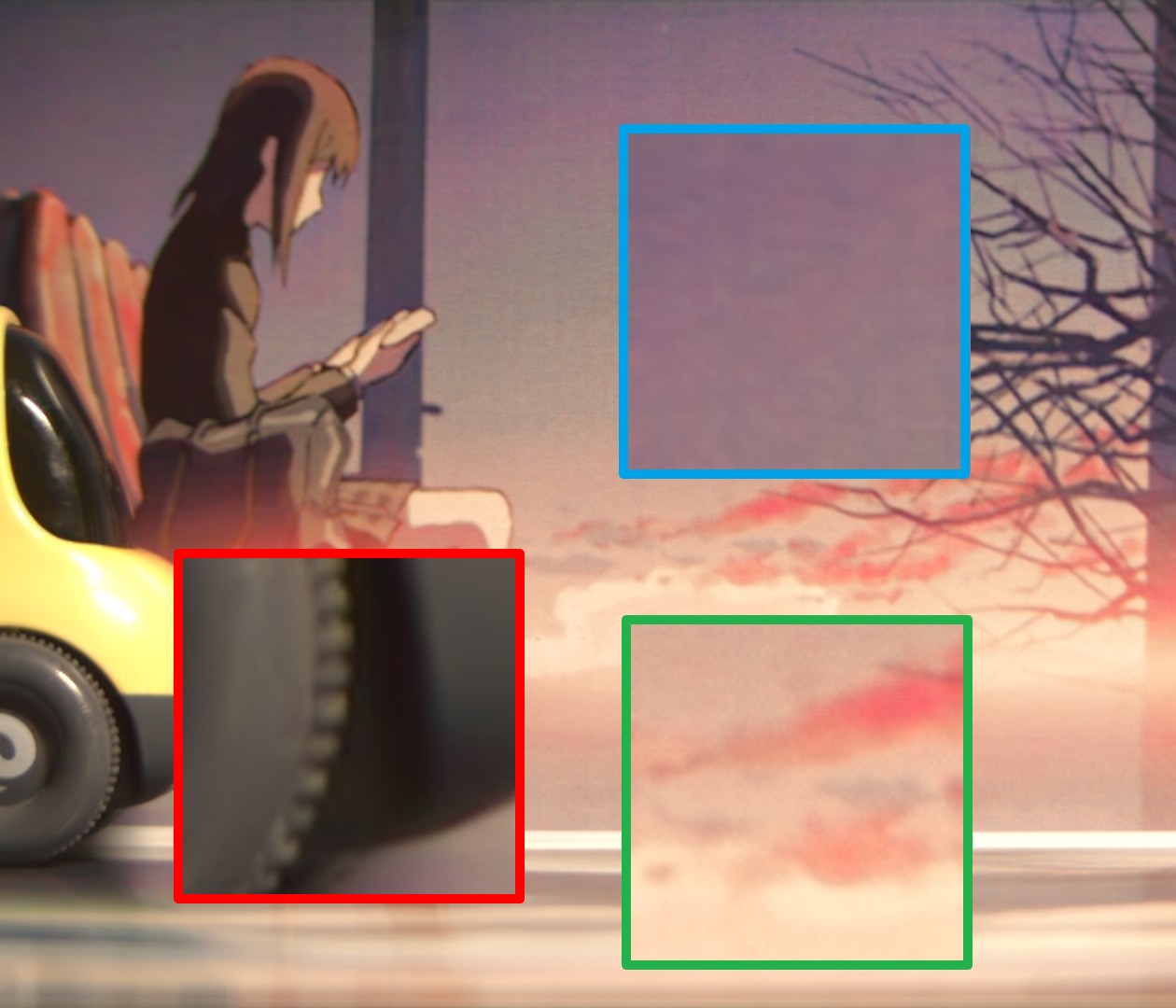}}
\hspace{-0.398em}%
\subfigure[Noisy]{
\label{Fig4}
\includegraphics[width=0.83016in]{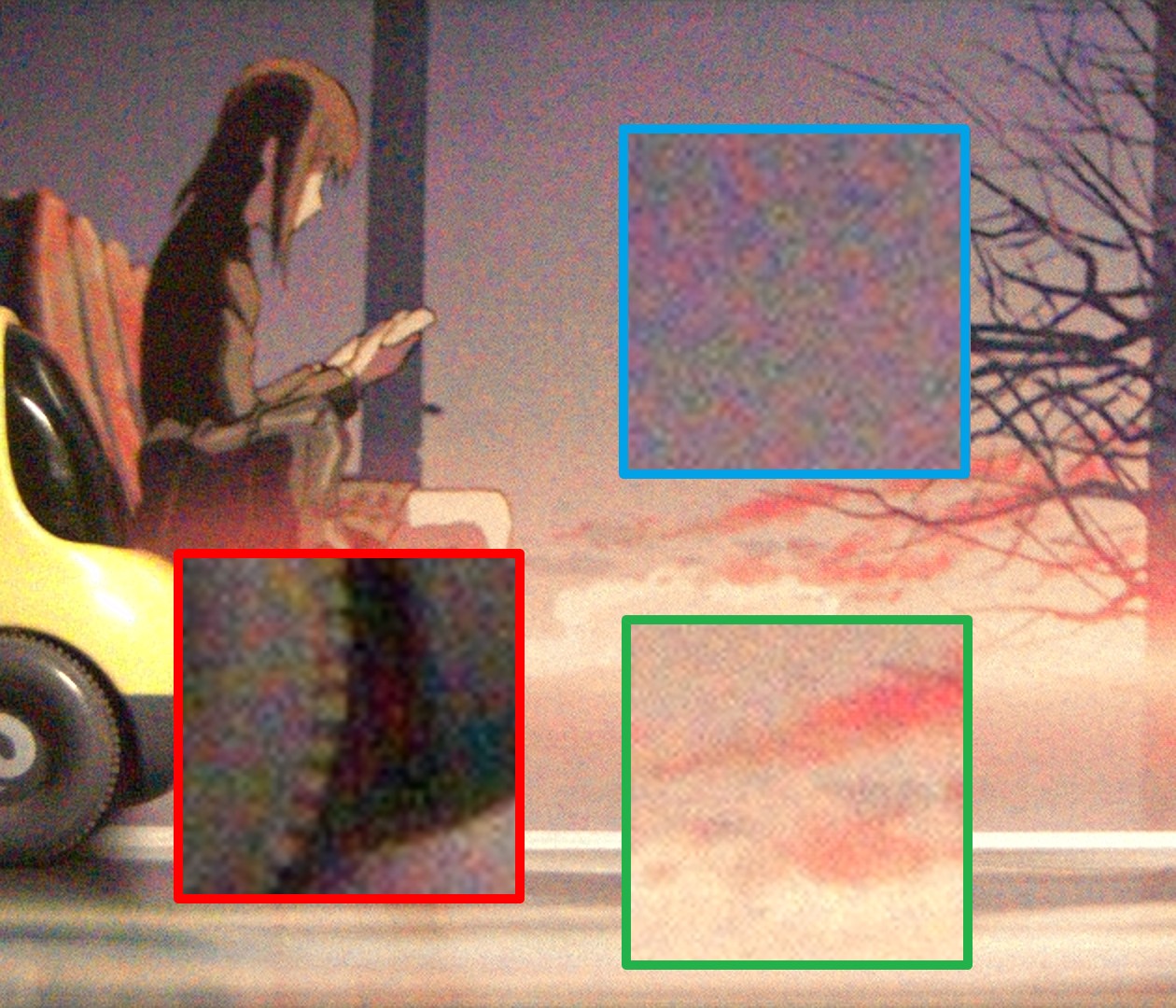}}
\hspace{-0.398em}%
\subfigure[FastDVDNet]{
\label{Fig4}
\includegraphics[width=0.83016in]{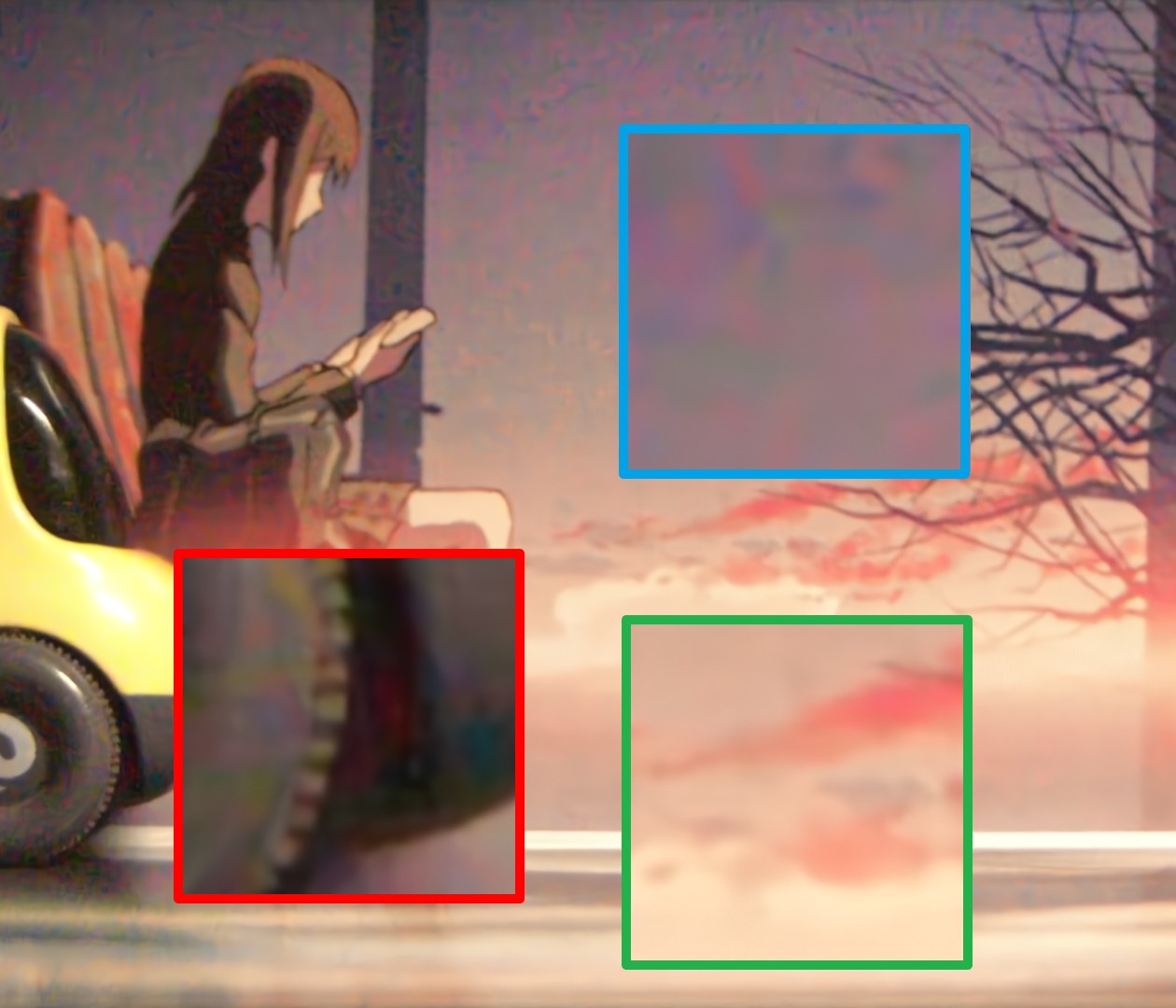}}
\hspace{-0.398em}%
\subfigure[STBN]{
\label{Fig4}
\includegraphics[width=0.83016in]{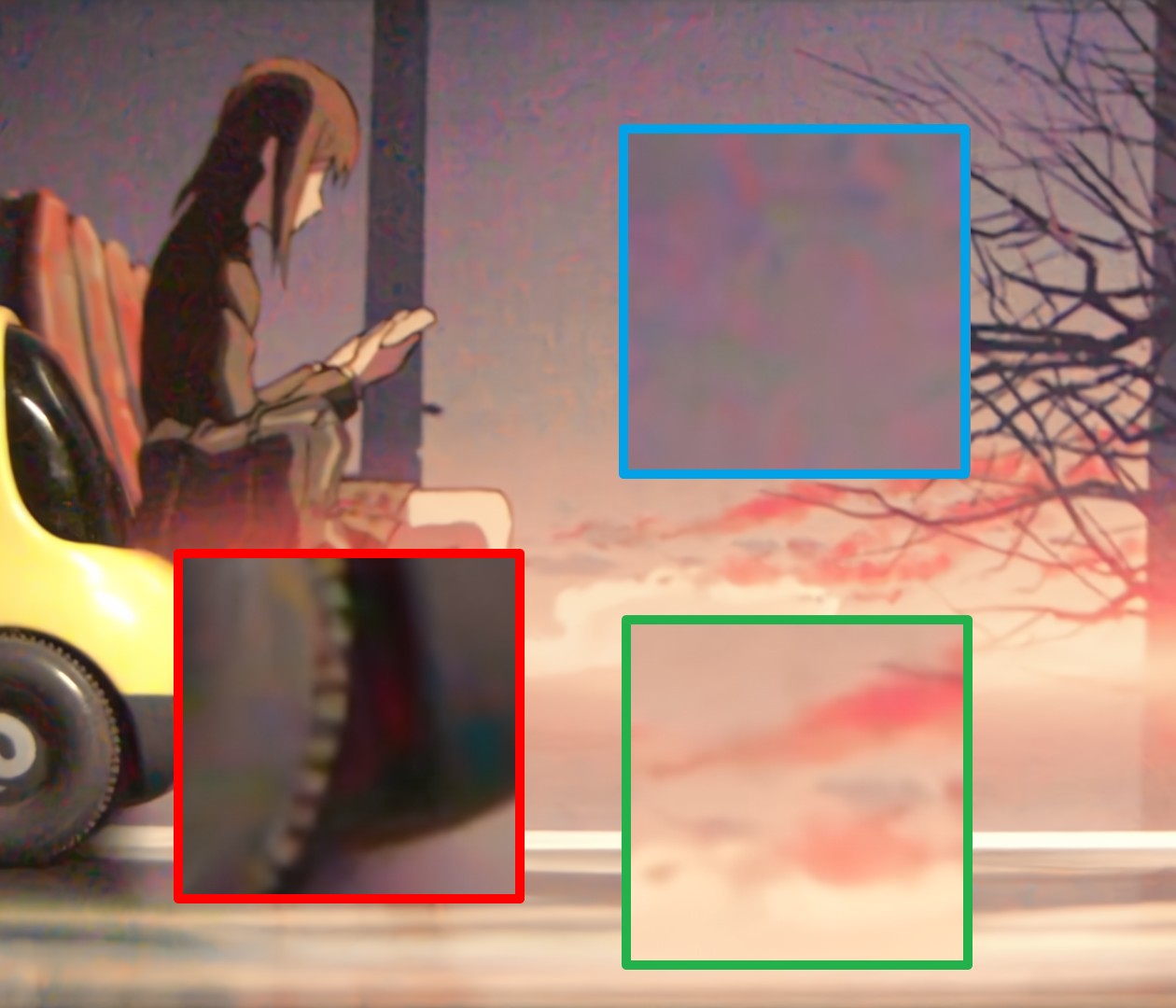}}\\
\vspace{-5.18pt}
\subfigure[VBM4D]{
\label{Fig4}
\includegraphics[width=0.83016in]{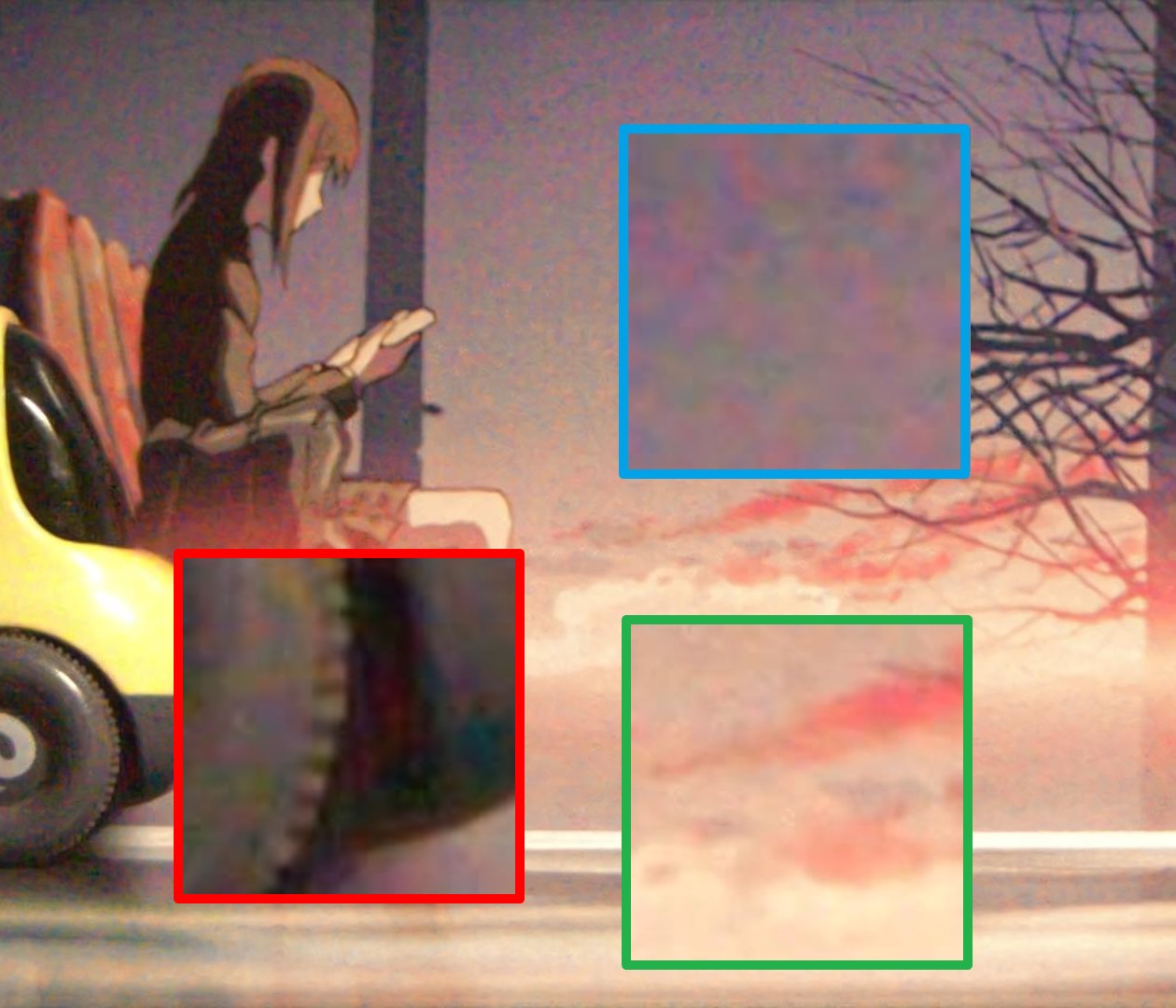}}
\hspace{-0.398em}%
\subfigure[VRT]{
\label{Fig4}
\includegraphics[width=0.83016in]{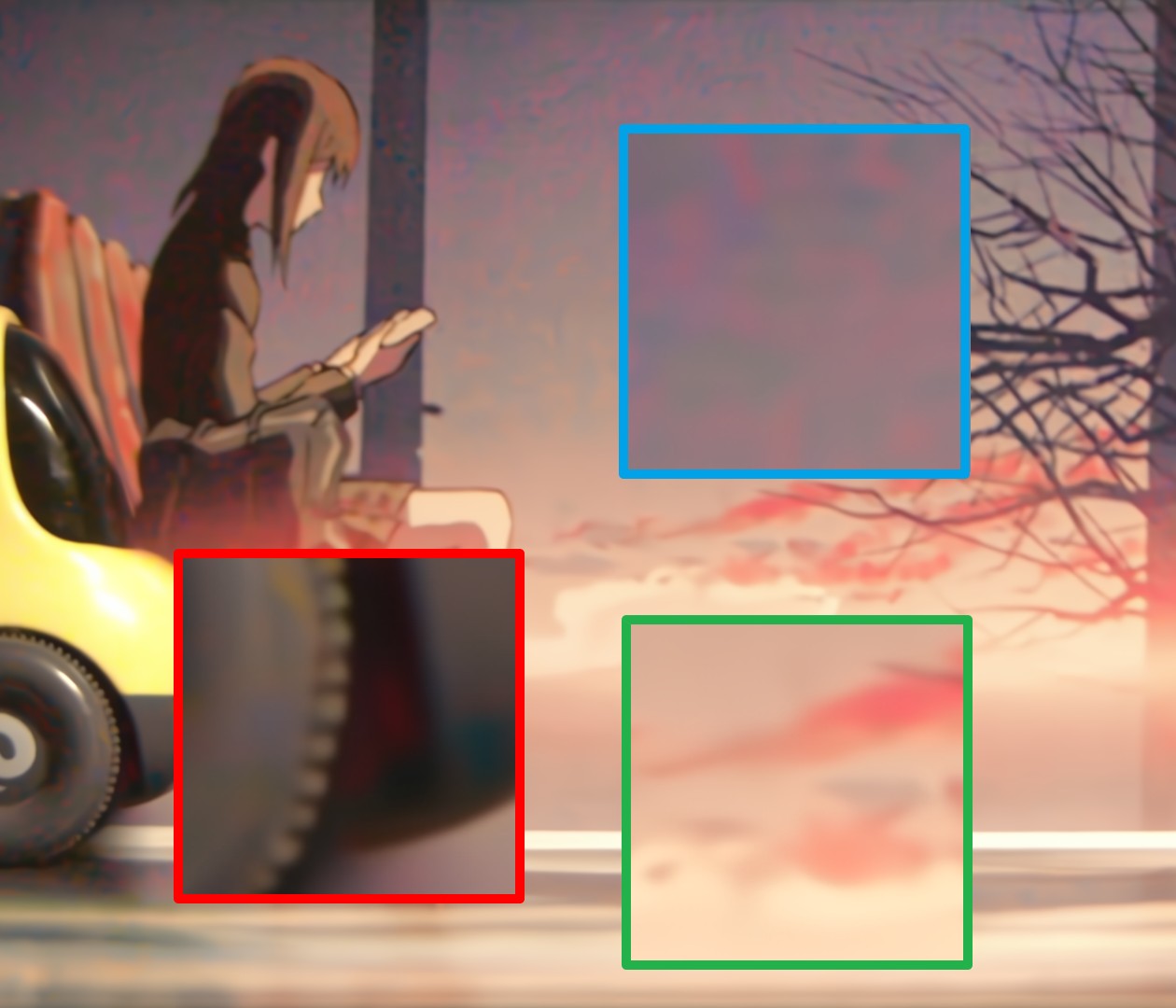}}
\hspace{-0.398em}%
\subfigure[VNLNet]{
\label{Fig4}
\includegraphics[width=0.83016in]{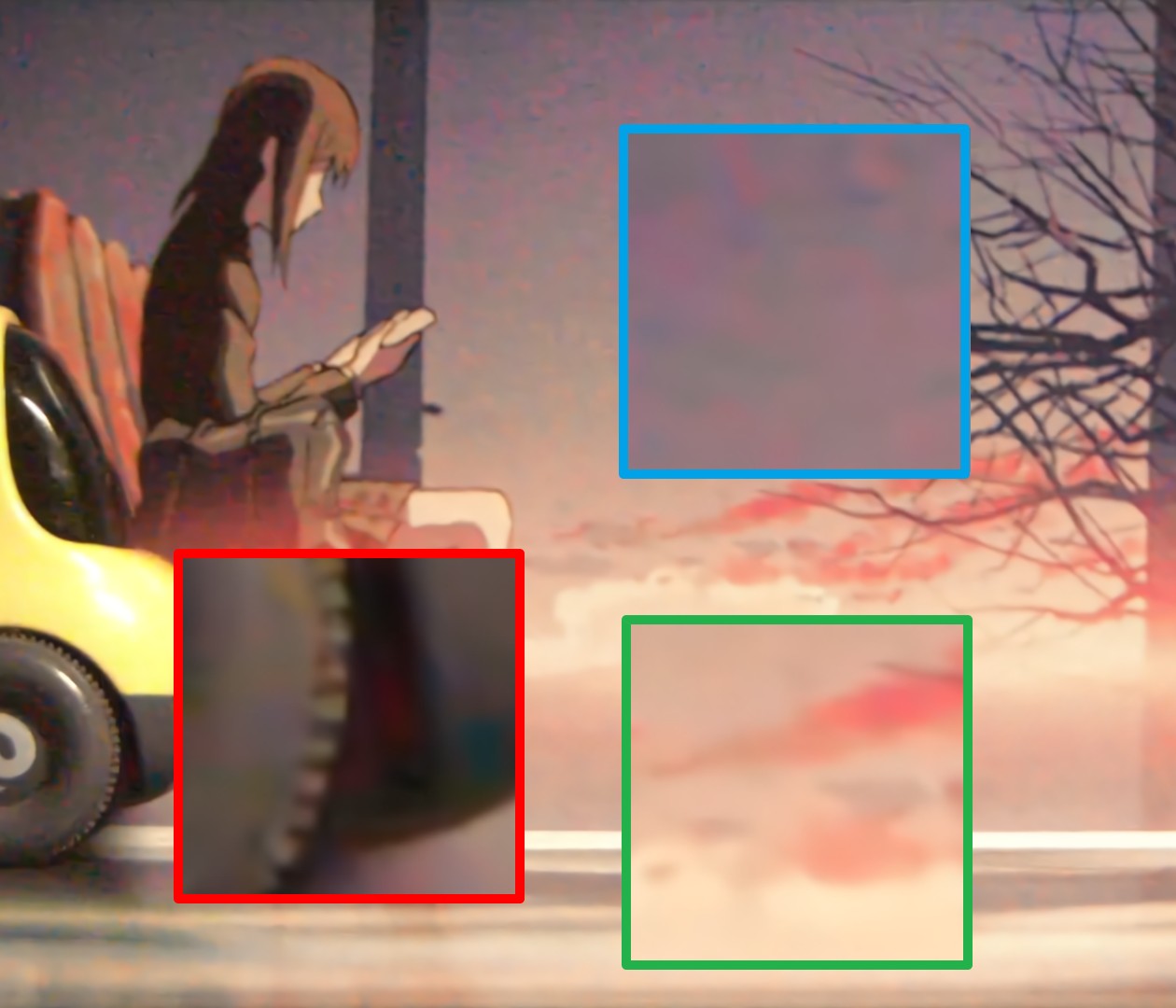}}
\hspace{-0.398em}%
\subfigure[RA-Haar-tSVD]{
\label{Fig4}
\includegraphics[width=0.83016in]{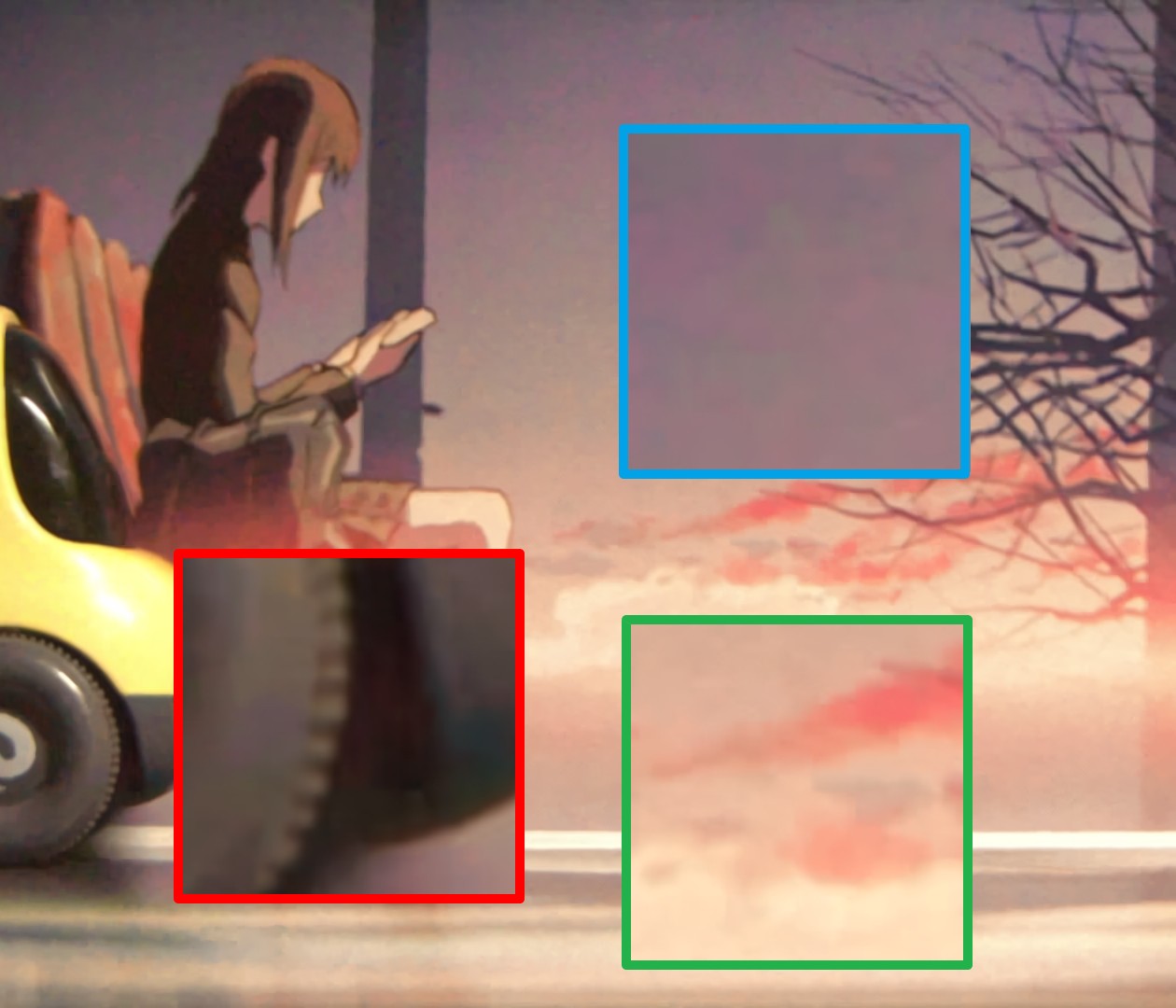}}
\vspace{-6.18pt}
\caption{Denoising comparison on the CRVD dataset (ISO = 12800).}
\vspace{-12.98pt}
\label{Fig_compare_with_CRVD_Sample5}
\end{figure}

Videos are more informative than images with dynamic objects and temporal continuity. The proposed method can be readily extended to handle video sequences. Specifically, for Haar-tSVD, the patch search is applied to both spatial and temporal dimension. The trained CNN-based estimator and the adaptive noise adjustment of A-Haar-tSVD can be adopted in a frame-by-frame fashion, while the enhancement strategy is imposed on groups obtained from spatio-temporal patch search. In our experiments, all video sequences of CRVD and IOCV are used for evaluations. The objective metrics are computed as the average across all frames \cite{Tassano_2020_CVPR}. Table \ref{Table_video_denoising_results} lists the quantitative results of compared methods. The proposed method achieves very competitive results on both datasets, and the enhancement strategy RA-Haar-tSVD outperforms baselines by a margin of at least 0.28dB in PSNR. These results demonstrate the effectiveness of combining the global t-SVD with the Haar transform in capturing the nonlocal characteristics of video data across frames.\\
\indent Visual comparisons are provided in Fig. \ref{Fig_compare_with_IOCV_case2} to Fig. \ref{Fig_compare_with_CRVD_Sample5}. As shown in Fig. \ref{Fig_compare_with_IOCV_case2} and Fig. \ref{Fig_compare_with_CRVD_case9}, even under moderate noise levels, the pretrained DNN models tend to exhibit more obvious over-smooth effects. In particular, Fig. \ref{Fig_compare_with_CRVD_case9} depicts a scene with static background, where the toy girl in blue is dynamic and moves in more than one directions. Consequently, some details and textures appear only in certain frames. Benefiting from the adaptive noise estimation strategy and the nonlocal characteristics, the proposed A-Haar-tSVD method effectively captures spatiotemporal similarity and preserves more structural information. Besides, Fig. \ref{Fig_compare_with_CRVD_Sample5} demonstrates the robustness of the proposed enhancement strategy to severe noise corruptions. By leveraging the patch- and group-level redundancy across different frames, the proposed method is able to suppress noise and mitigate color artifacts.

\subsection{Real-world HSI Denoising}
\begin{table*}[htbp]
\scriptsize
  \centering
  \caption{Denoising results of compared methods on the Real-HSI dataset.}
  \scalebox{0.800168}{
    \begin{tabular}{cccccccccccccccc}
    \toprule
    \multirow{3}[3]{*}{Datasets} & \multirow{3}[3]{*}{Metrics} & \multicolumn{8}{c}{Traditonal denoisers}                      & \multicolumn{6}{c}{DNN methods} \\
\cmidrule{3-16}          &       & BM4D  & LLRT  & LTDL  & MSt-SVD & NGMeet & SSTPTV & Haar-tSVD & A-Haar-tSVD & FlexDID & HSI-DeNet & Mac-Net & QRNN3D & RAS2S & sDeCNN \\
          &       & \cite{maggioni2012nonlocal} & \cite{chang2017hyper} & \cite{gong2020low} & \cite{kong2019color} & \cite{he2019non} & \cite{zhao2025spatial} & (Ours) & (Ours) & \cite{chen2024flex}& \cite{chang2018hsi} & \cite{xiong2021mac} & \cite{wei2020physics} & \cite{xiao2024region} & \cite{maffei2019single} \\
    \midrule
    \multirow{4}[8]{*}{Real-HSI} & PSNR $\uparrow$  & 25.88  & \textbf{25.90} & 25.80  & 25.86  & 25.87  & 25.21  & 25.82  & 25.81  & 25.31  & 25.63  & 25.88  & 25.82  & 25.87  & 25.70  \\
\cmidrule{2-16}          & SSIM $\uparrow$  & 0.865  & 0.861  & 0.841  & 0.866  & 0.866  & 0.815  & 0.865  & 0.866  & 0.820  & 0.853  & 0.863  & \textbf{0.869} & 0.868  & 0.860  \\
\cmidrule{2-16}          & SAM \cite{yuhas1990determination} $\downarrow$   & 0.066  & 0.060  & 0.074  & 0.063  & \textbf{0.051} & 0.072  & 0.063  & 0.064  & 0.075  & 0.092  & 0.056  & 0.064  & 0.054  & 0.093  \\
\cmidrule{2-16}          & EGRAS \cite{wald2002data} $\downarrow$  & 222.68  & 224.05  & 223.32  & \textbf{222.64} & 222.69  & 233.51  & 222.96  & 223.12  & 232.12  & 232.73  & 222.84  & 225.32  & 222.83  & 227.88  \\
    \midrule
    Platform & -     & CPU   & CPU   & CPU   & CPU   & CPU   & CPU   & CPU   & CPU/GPU   & GPU (T4) & GPU (T4) & GPU (T4) & GPU (T4) & GPU (T4) & GPU (T4) \\
    \midrule
    Time  & minutes & 4.1   & 16.5  & 35.0  & 2.8   & 4.1   & 4.9   & \textbf{0.1} & 0.2 & 6.1   & 0.8   & 0.2   & \textbf{0.1} & \textbf{0.1} & 0.7  \\
    \bottomrule
    \end{tabular}}%
  \label{Table_Real_HSI}%
  \vspace{-12.8pt}
\end{table*}%

\begin{figure}[htbp]
\vspace{-6.98pt}
\graphicspath{{Figs/Fig_HSI/Fig_curve_graphs/}}
  \centering
  \includegraphics[width=3.5019in]{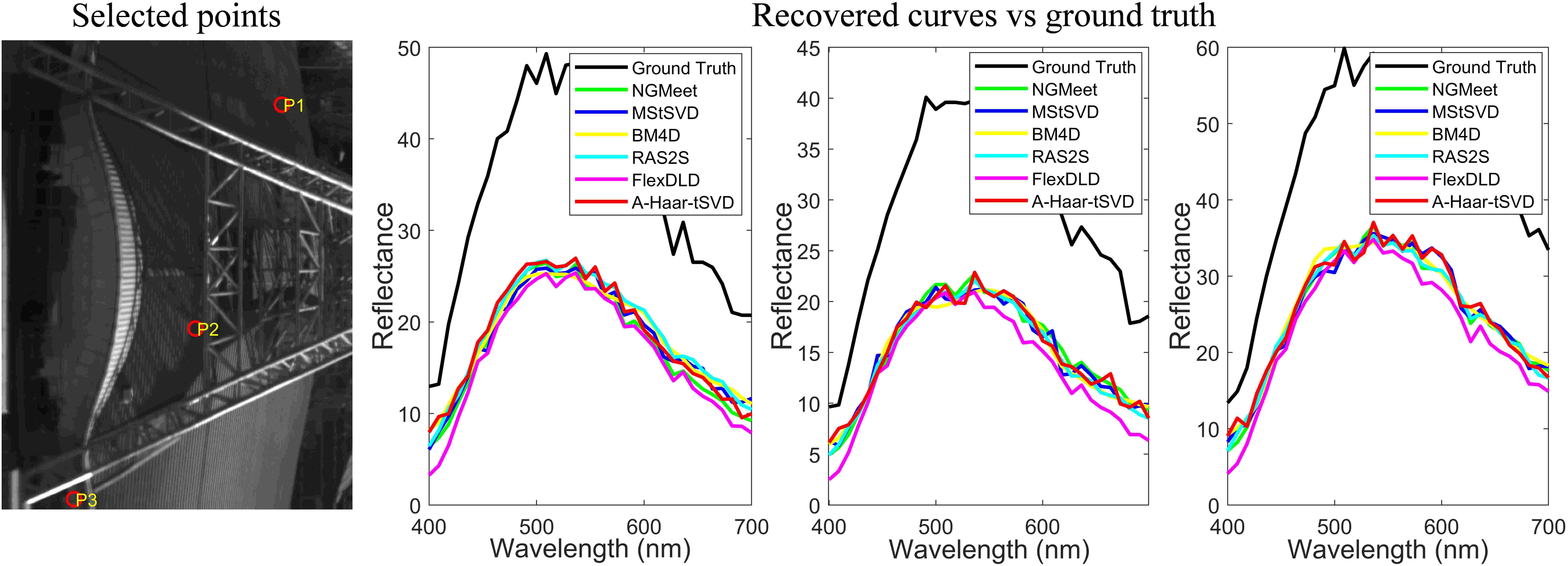}
  \vspace{-10.98pt}
  \caption{Spectral curves of randomly selected points across different regions.}
  \label{Fig_Real_HSI_Curve_graph}
  \vspace{-6.18pt}
\end{figure}

\begin{figure}[htbp]
\vspace{-6.98pt}
\graphicspath{{Figs/Fig_HSI/Fig_selected_est_nosie_band/}}
\centering
\subfigure[PaviaU]{
\label{Fig4}
\includegraphics[width=1.68016in]{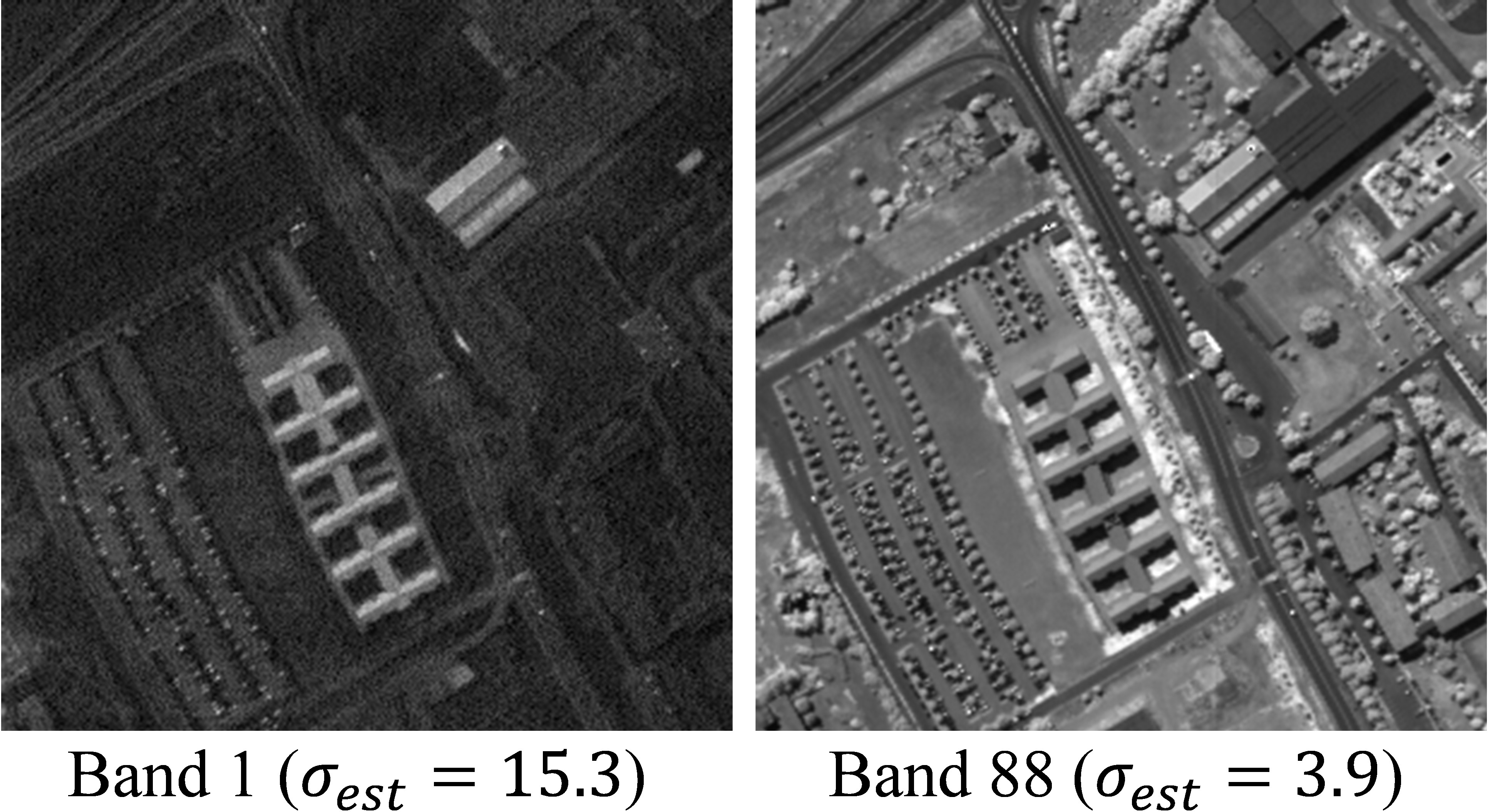}}
\hspace{-0.168em}%
\subfigure[Urban]{
\label{Fig4}
\includegraphics[width=1.68016in]{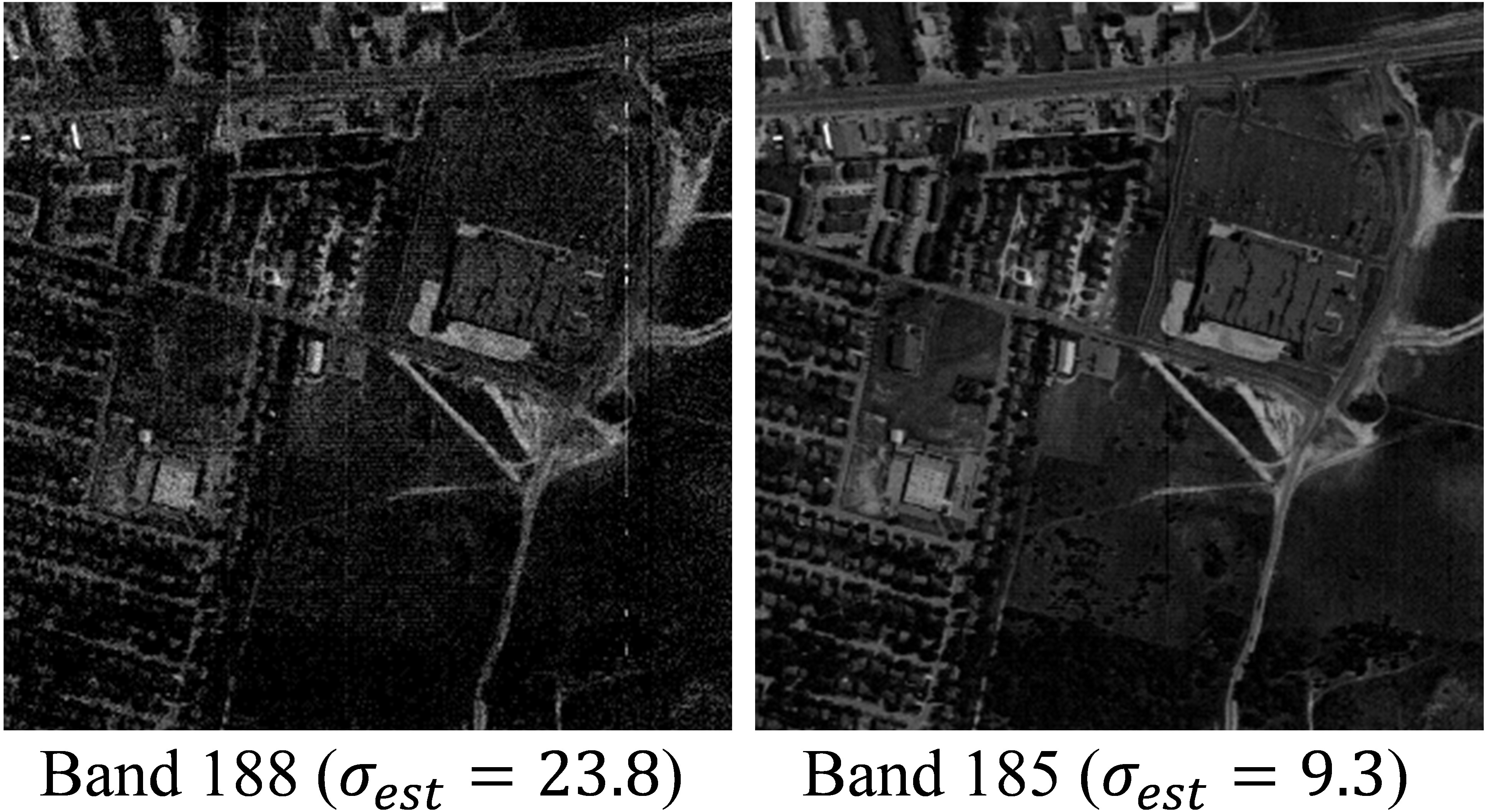}}
\vspace{-5.98pt}
\caption{Estimated noise $\sigma_{est}$ for different bands with the proposed strategy.}
\label{Fig_HSI_est_noise_by_band}
\vspace{-5.18pt}
\end{figure}

\begin{figure}[htbp]
\vspace{-8.18pt}
\graphicspath{{Figs/Real_HSI_witout_ref_selected/PaviaU/Combined_new/}}
\centering
\subfigure[Noisy]{
\label{Fig4}
\includegraphics[width=0.8116in]{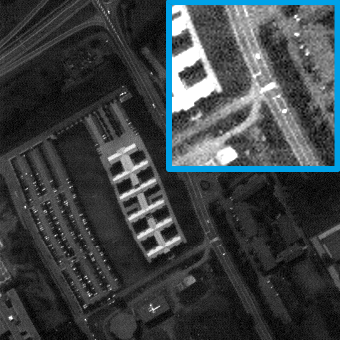}}\hfill
\subfigure[BM4D]{
\label{Fig4}
\includegraphics[width=0.8116in]{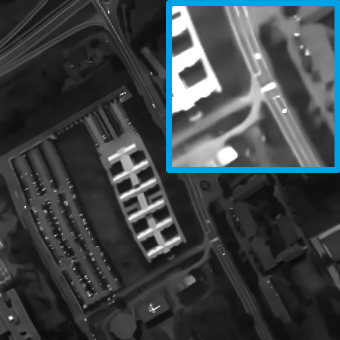}}\hfill
\subfigure[RAS2S]{
\label{Fig4}
\includegraphics[width=0.8116in]{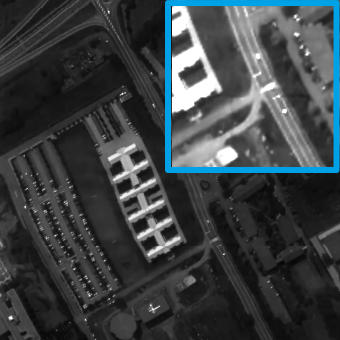}}\hfill
\subfigure[A-Haar-tSVD]{
\label{Fig4}
\includegraphics[width=0.8116in]{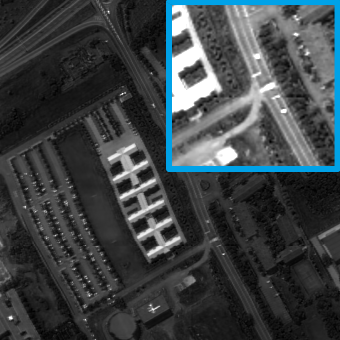}}
\vspace{-2.8pt}
\caption{Denoising comparison on the PaviaU data \cite{CupriteHSI}.}
\label{Fig_HSI_PaviaU}
\vspace{-12.8pt}
\end{figure}

\begin{figure}[htbp]
\vspace{-10.18pt}
\graphicspath{{Figs/Real_HSI_witout_ref_selected/Urban/Combined_new/}}
\centering
\subfigure[Noisy]{
\label{Fig4}
\includegraphics[width=0.8196in]{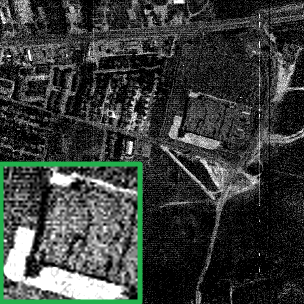}}\hfill
\subfigure[BM4D]{
\label{Fig4}
\includegraphics[width=0.8196in]{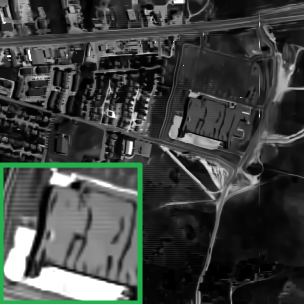}}\hfill
\subfigure[RAS2S]{
\label{Fig4}
\includegraphics[width=0.8196in]{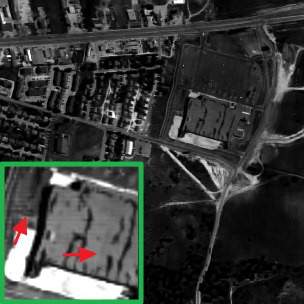}}\hfill
\subfigure[A-Haar-tSVD]{
\label{Fig4}
\includegraphics[width=0.8196in]{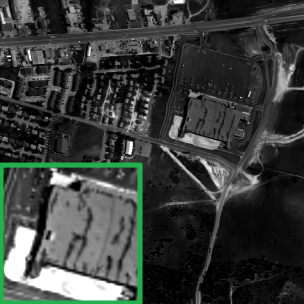}}
\vspace{-9.18pt}
\caption{Denoising comparison on the Urban data \cite{kalman1997classification}.}
\label{Fig_HSI_Urban}
\vspace{-6.98pt}
\end{figure}

\begin{figure}[htbp]
\vspace{-12.18pt}
\graphicspath{{Figs/Real_HSI_witout_ref_selected/EO1/Combined_new/}}
\centering
\subfigure[Noisy]{
\label{Fig4}
\includegraphics[width=0.8116in]{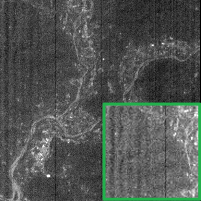}}\hfill
\subfigure[BM4D]{
\label{Fig4}
\includegraphics[width=0.8116in]{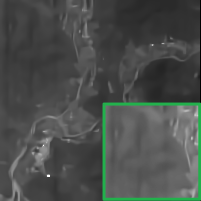}}\hfill
\subfigure[RAS2S]{
\label{Fig4}
\includegraphics[width=0.8116in]{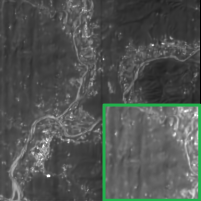}}\hfill
\subfigure[A-Haar-tSVD]{
\label{Fig4}
\includegraphics[width=0.8116in]{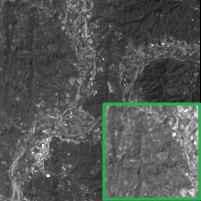}}
\vspace{-6.8pt}
\caption{Denoising comparison on the EO1 data \cite{middleton2013earth}.}
\vspace{-3.98pt}
\label{Fig_HSI_EO1}
\end{figure}

\indent HSI plays a vital role in a variety of remote sensing applications\cite{song2020unsupervised}. To handle an HSI data $\mathcal{Y} \in \mathbb{R}^{H \times W \times N_{\text{bands}}}$, we treat each local patch as a long tube and set $ps = 8 \times 8 \times N_{\text{bands}}$, while other parameters are kept the same as those used in the image denoising task. To apply the CNN noise estimator without retraining, we estimate the noise of $\mathcal{Y}$ based on the mean value across all spectral bands. Objective results of compared methods are given in Table \ref{Table_Real_HSI}. By effectively capturing the rich spatial–spectral correlations in HSI data with circulant structures, the proposed Haar-tSVD and its adaptive variant achieve competitive performance compared with state-of-the-art approaches. Meanwhile, by eliminating the need for complex iterative filtering and local transform learning, our method attains inference time comparable to advanced DNN models. These properties make it a practical and efficient solution for large-scale HSI denoising. 
\begin{figure}[htbp]
\vspace{-3.68pt}
\graphicspath{{Figs/Selected_color_images/Real-HSI/Case4/Combined/}}
\centering
\subfigure[Mean]{
\label{Fig4}
\includegraphics[width=0.816in]{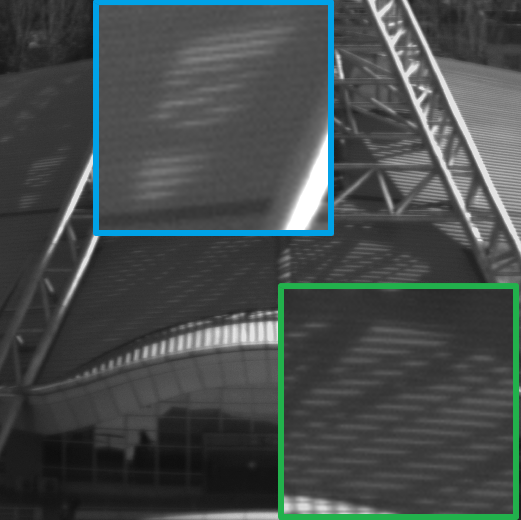}}\hfill
\subfigure[Noisy]{
\label{Fig4}
\includegraphics[width=0.816in]{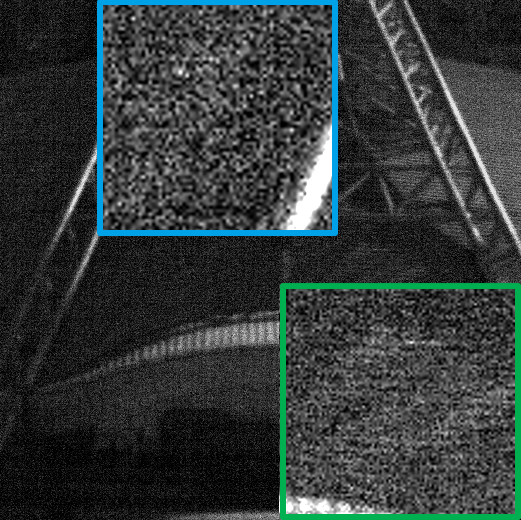}}\hfill
\subfigure[BM4D]{
\label{Fig4}
\includegraphics[width=0.816in]{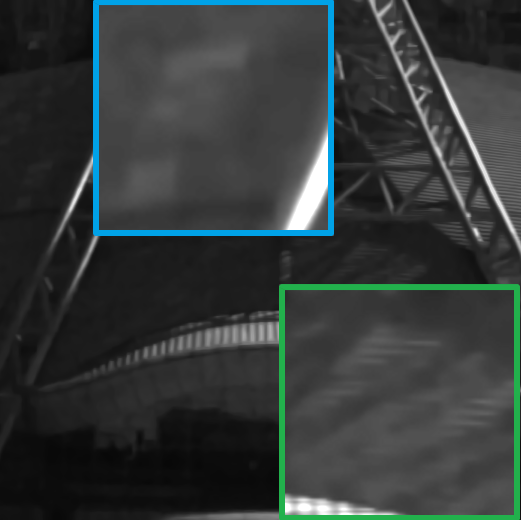}}\hfill
\subfigure[OLRT]{
\label{Fig4}
\includegraphics[width=0.816in]{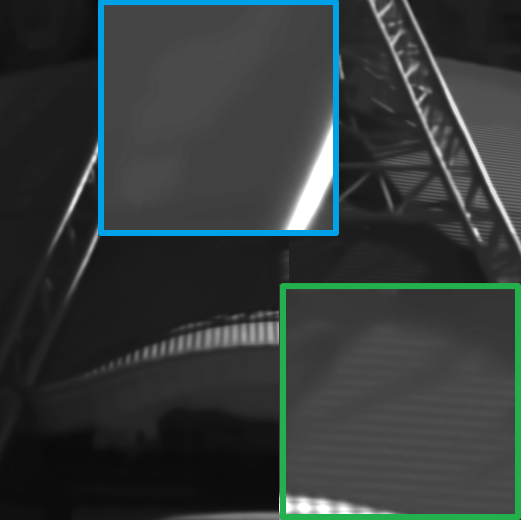}}\hfill \\
\vspace{-6.98pt}
\subfigure[FlexDLD]{
\label{Fig4}
\includegraphics[width=0.816in]{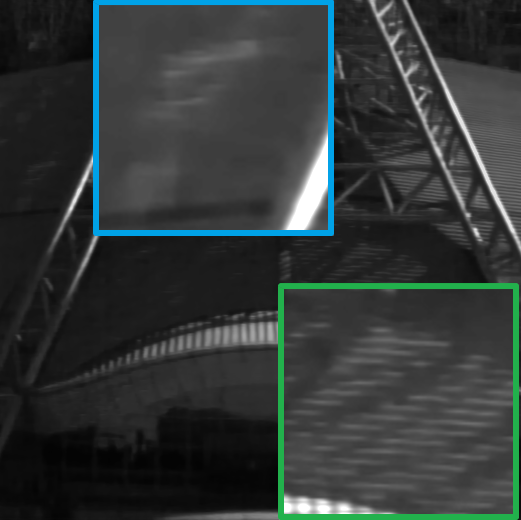}}\hfill
\subfigure[QRNN3D]{
\label{Fig4}
\includegraphics[width=0.816in]{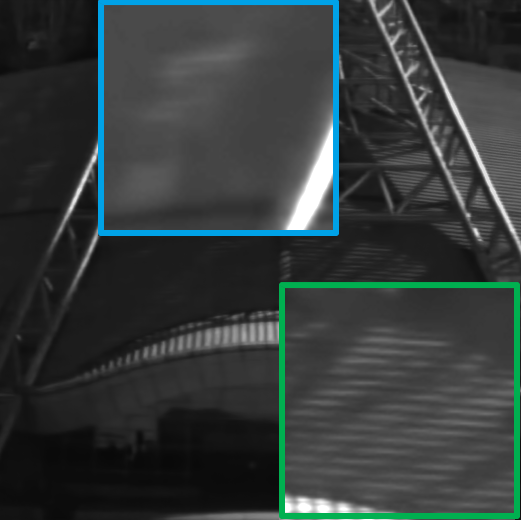}}\hfill
\subfigure[RAS2S]{
\label{Fig4}
\includegraphics[width=0.816in]{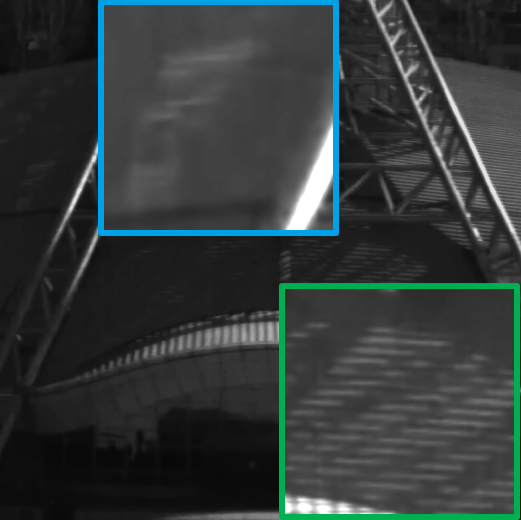}}\hfill
\subfigure[A-Haar-tSVD]{
\label{Fig4}
\includegraphics[width=0.816in]{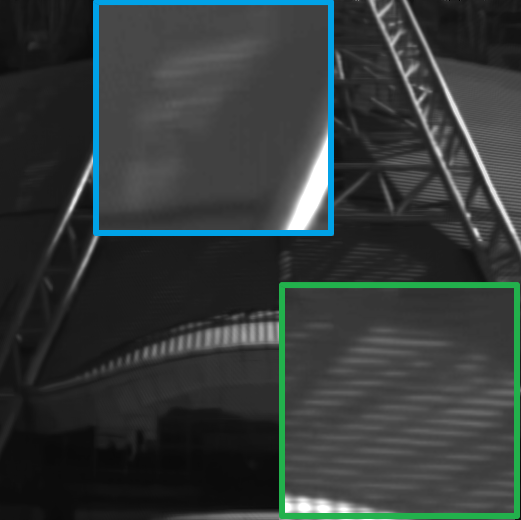}}
\vspace{-6.8pt}
\caption{Denoising comparison on the Real-HSI dataset.}
\label{Fig_compare_with_Real_HSI_case4}
\vspace{-0.18pt}
\end{figure}
\\
\indent Visual comparisons of competitive methods on the Real-HSI dataset are shown in Fig. \ref{Fig_compare_with_Real_HSI_case4}. We can see that The proposed method achieves a good balance between smoothness and detail preservation. In contrast, BM4D tends to produce obvious oversmooth effects, since its predefined patch-level transforms may not fully exploit the correlation across all the spectral bands. Additionally, the state-of-the-art low-rank tensor method OLRT struggles to preserve high-frequency components such as edges and textures. These observations suggest that increasing the number of iterations and similar patches may not help preserve fine details and structure of HSI data. Moreover, although the DNN-based methods FlexDLD and RAS2S demonstrate impressive noise suppression, they introduce mild artifacts. Fig.~\ref{Fig_Real_HSI_Curve_graph} illustrates the spectral reflectance curves of the ground-truth and the reconstructed HSIs, further confirming the competitive performance of our method in this challenging scenario.
\\
\indent In many cases, the noise distribution can be highly non-uniform across spectral bands, with a few bands suffering from severe contamination. To address this issue, we perform band-wise noise estimation, enabling the identification of heavily corrupted bands based on the predicted noise level $\sigma_{est}$, as illustrated in Fig. \ref{Fig_HSI_est_noise_by_band}. For severely contaminated bands (defined as $\sigma_{est} > 2{\sigma_{mean}}$, where $\sigma_{mean}$ denotes the mean noise level across all bands), we adopt the highest noise estimation output to ensure noise suppression. The idea is to exploit the rich redundancy information inherent in HSI tubes to achieve noise removal. However, this may cause oversmooth effects to the cleaner bands. Therefore, for the less corrupted spectral bands, we can refine the denosing result by adopting their mean noise estimation value $\sigma_{mean}$ to preserve fine details.
 \\ 
\indent Visual evaluations for several real-world HSI datasets are provided in Fig. \ref{Fig_HSI_PaviaU} to Fig. \ref{Fig_HSI_EO1}, illustrating the proposed method's performance on challenging cases. The qualitative evaluations suggest that benefiting from the adaptive noise estimation scheme, the proposed method can achieve promising results. In particular, Fig. \ref{Fig_HSI_Urban} and Fig. \ref{Fig_HSI_EO1} demonstrate its ability of handling complex noise patterns, showing a reasonable ability to suppress both random noise and structured striping artifacts. Moreover, fine-grained textural details are well preserved without introducing noticeable artifacts.

\vspace{-3.8pt}
\subsection{Parameter Analysis}
\begin{figure}[htbp]
\vspace{-6.88pt}
\graphicspath{{Figs/Parameter_tuning/}}
\centering
\subfigure[Patch size $ps$]{
\label{Fig4}
\includegraphics[width=1.1129in]{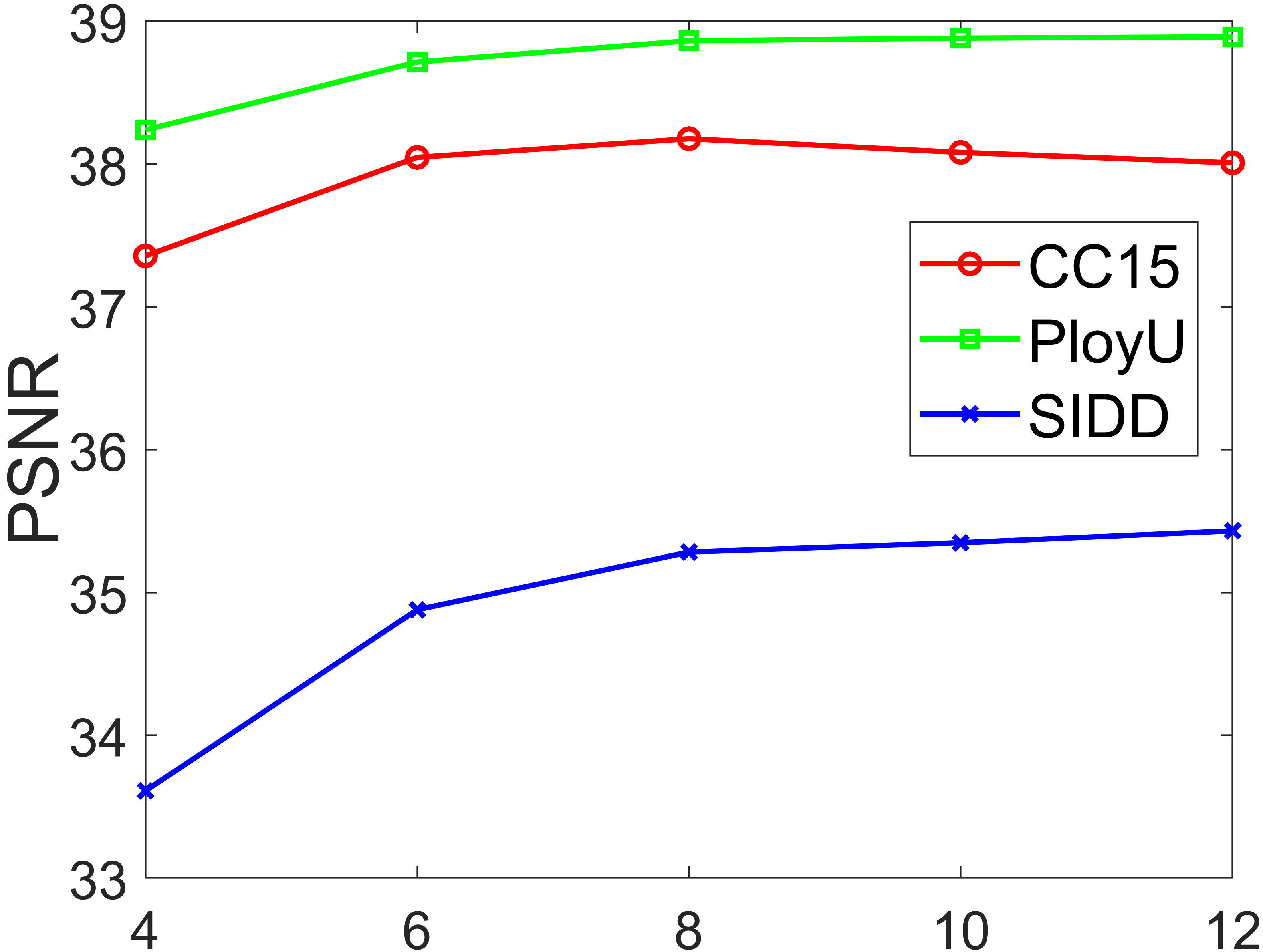}}\hfill
\subfigure[Search range $W$]{
\label{Fig4}
\includegraphics[width=1.1129in]{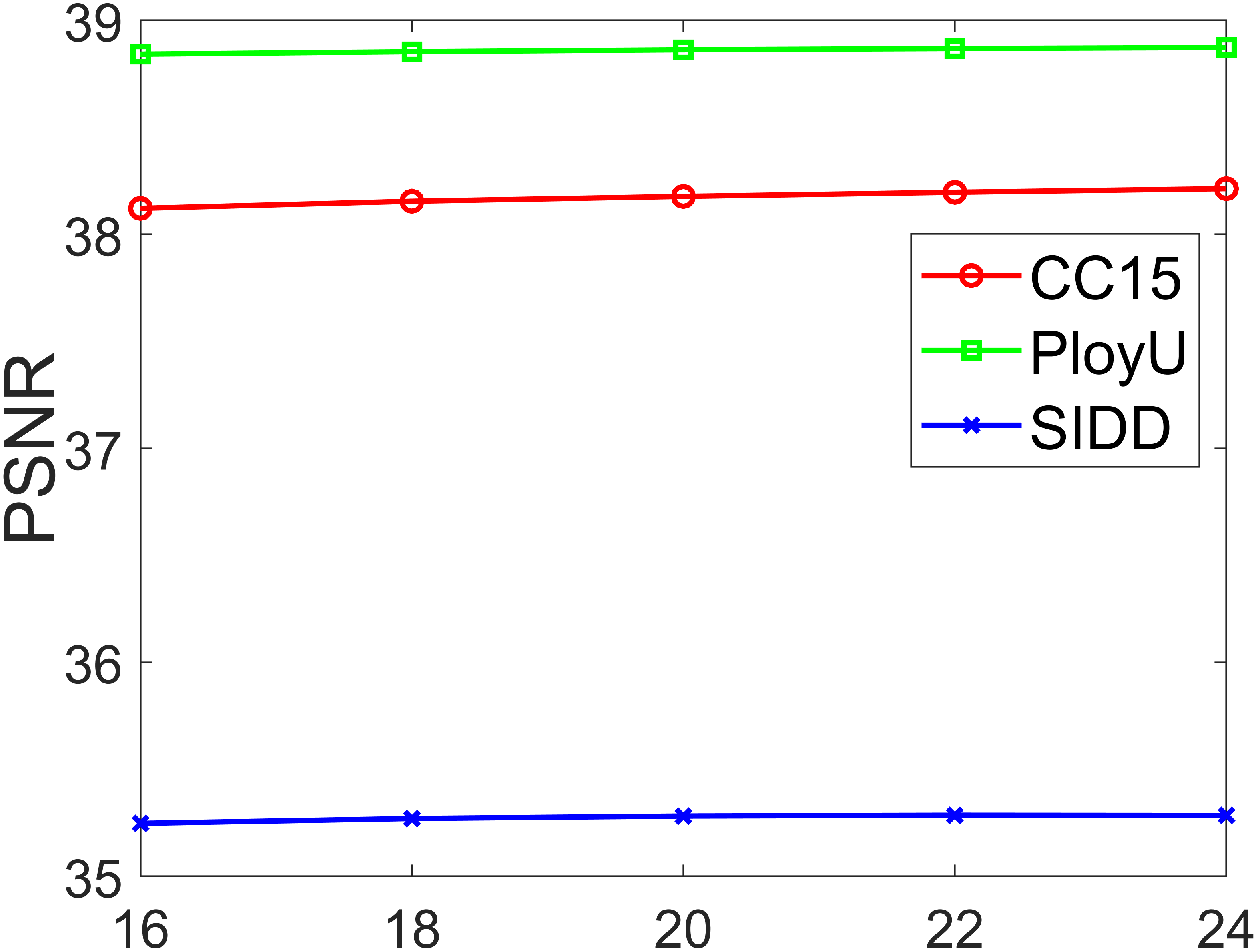}}\hfill
\subfigure[Similar patches $K$]{
\label{Fig4}
\includegraphics[width=1.1129in]{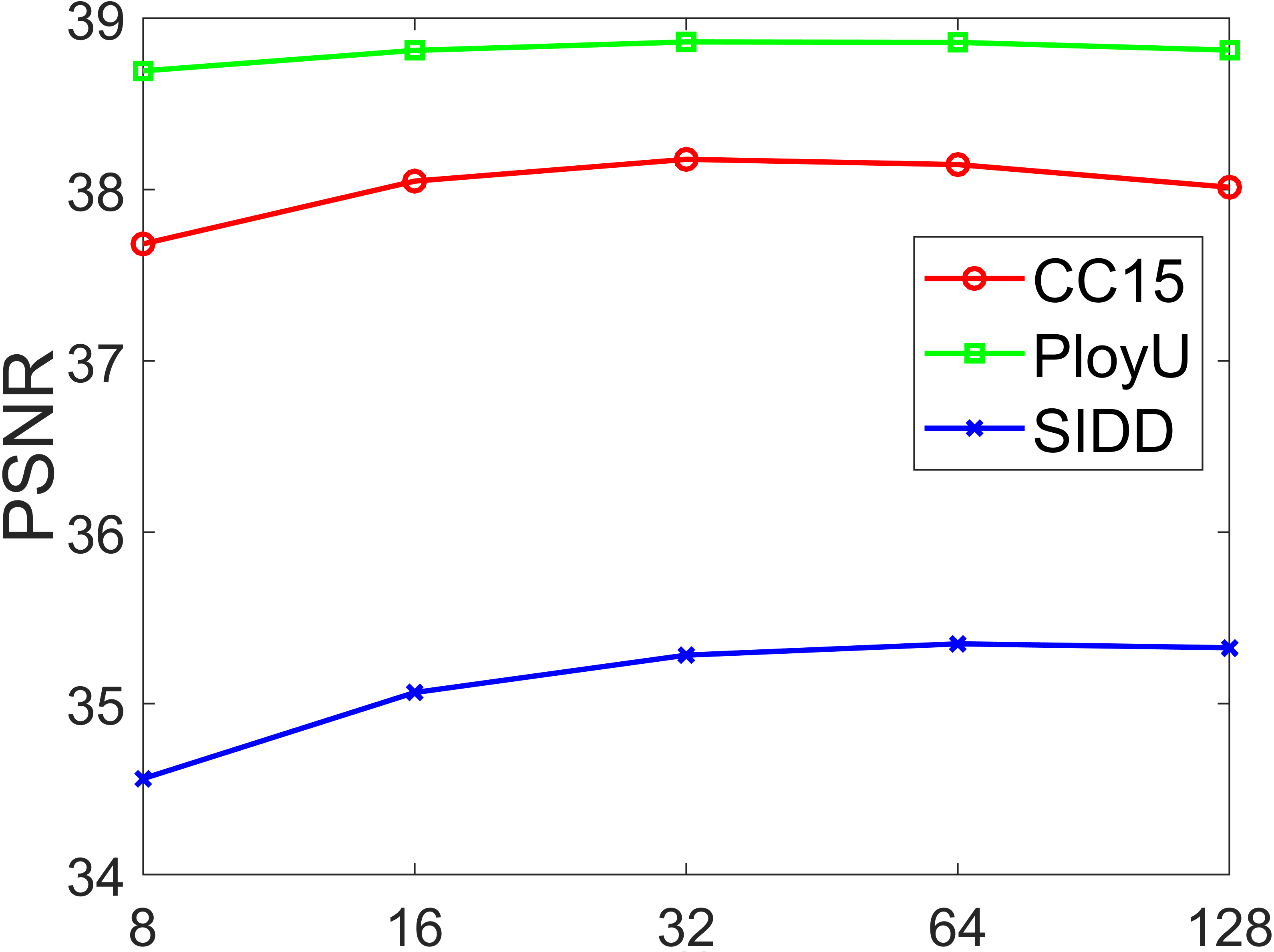}}\hfill
\vspace{-2.8pt}
\caption{Influence of different parameters on the Haar-tSVD transform.}
\label{Fig_parameter_tuning_ps_SR_maxK}
\vspace{-3.98pt}
\end{figure}

Following the classic patch-based denoising paradigm, the proposed Haar-tSVD transform relies on several key parameters such as the patch size $ps$, search window range $W$ and number of nonlocal similar patches $K$. Fig. \ref{Fig_parameter_tuning_ps_SR_maxK} evaluates the impact of these parameters on the denoising performance of Haar-tSVD across different datasets. For the patch size $ps$, a comparison between the CC15 and SIDD datasets shows that larger patch size can improve robustness against heavy noise contamination. However, increasing $ps$ may drastically raise computational burden. Similarly, expanding the search range $W$ and choosing more grouped patches $K$ do not guarantee performance enhancement due to the rare patch effect. Based on these observations, we mainly set $ps = 8$, $W = 18$ and $K = 32$ in our experiments to achieve a tradeoff between denoising performance and speed. 

\vspace{-8.98pt}
\subsection{Discussion}
\subsubsection{Ablation study}
\begin{figure}[htbp]
\vspace{-5.18pt}
\graphicspath{{Figs/Ablation_study/}}
\centering
\subfigure[$\beta$]{
\label{Fig4}
\includegraphics[width=1.5989in]{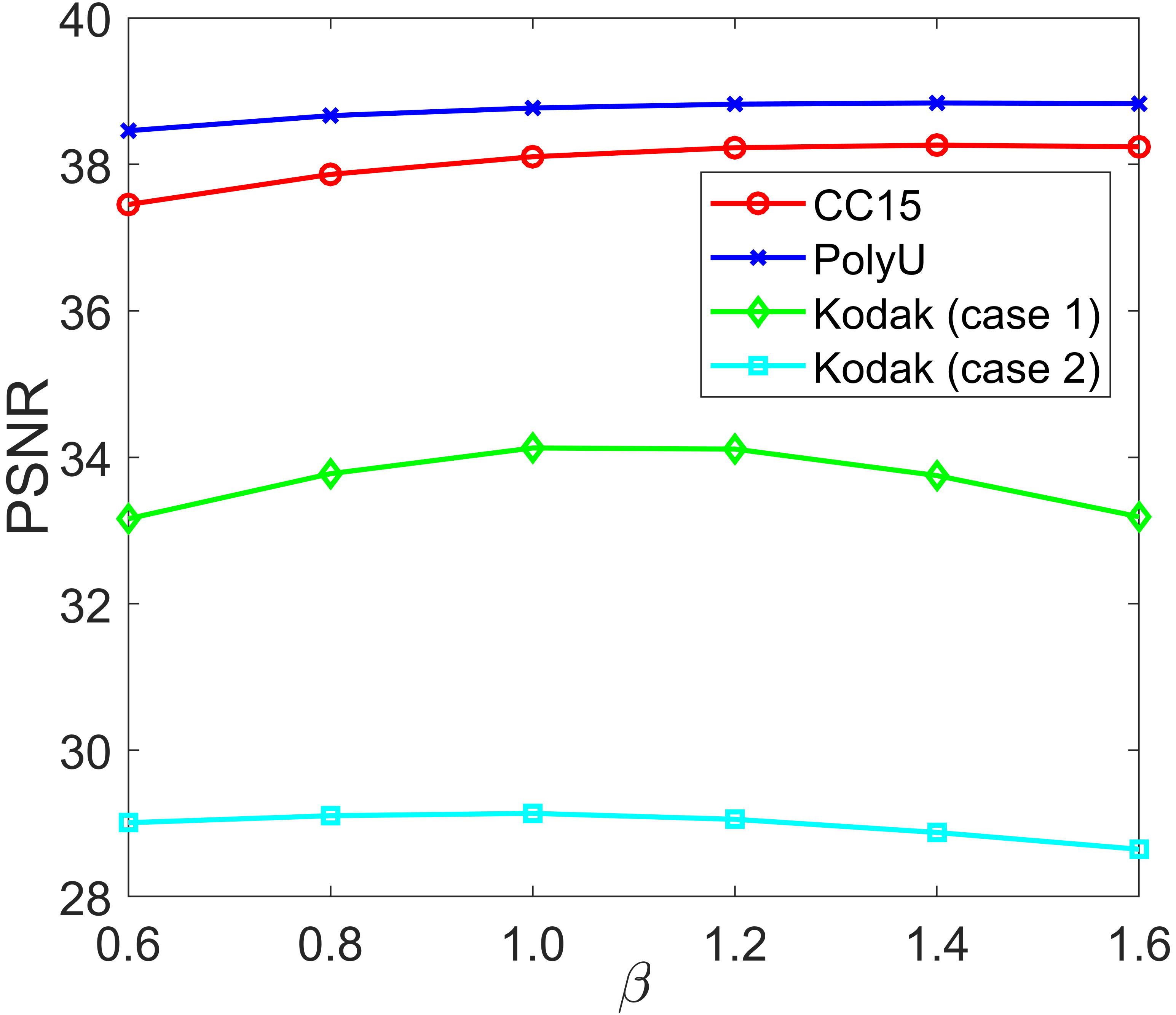}}
\subfigure[$\gamma$]{
\label{Fig4}
\includegraphics[width=1.5989in]{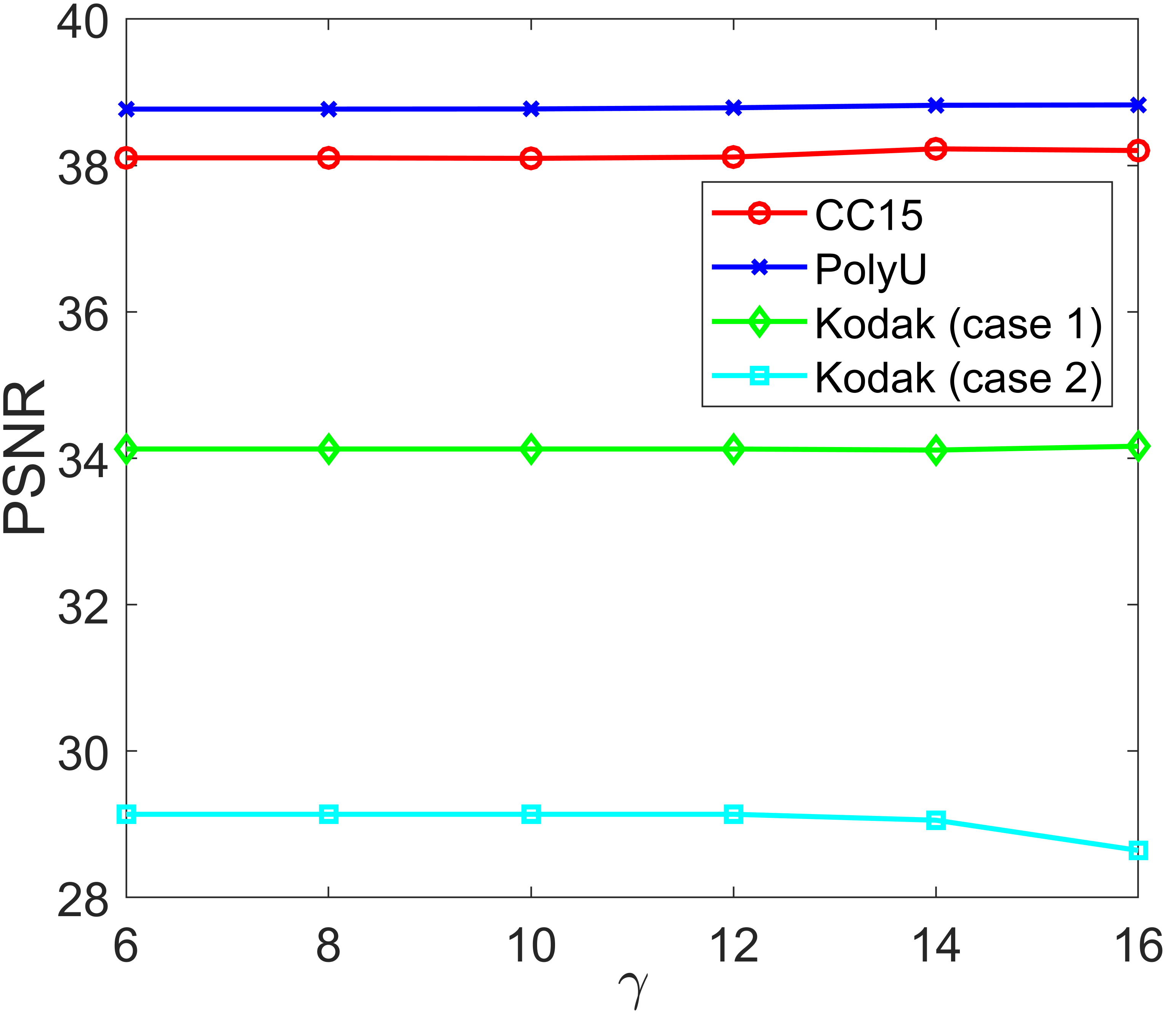}}
\vspace{-2.68pt}
\caption{Impact of weighting parameters $\beta$ and $\gamma$ on A-Haar-tSVD.}
\label{Fig_ablation_study_beta_gamma}
\vspace{-0.08pt}
\end{figure}

The effectiveness and robustness of the effective adaptive variant A-Haar-tSVD largely stems from the local adjustment of noise levels through eigenvalue analysis in Equ. (\ref{Equ_other_eigen_value}). To further assess its impact, we conduct an ablation study. As shown in Table \ref{Table_ablation_study_W_WO_PCA_est}, A-Haar-tSVD consistently benefits from the adaptive noise level adjustment scheme in Equ. (\ref{Equ_adjust_noise_level}) across different scenarios, while incurring minimal computational cost, which justifies the efficiency and applicability of the proposed adaptive mechanism.
\begin{table}[htbp]
\vspace{-5.98pt}
  \centering
  \caption{Comparison of the proposed A-Haar-tSVD with and without local noise level adjustment.}
  \renewcommand{\arraystretch}{0.59998}
  \scalebox{0.939}{
    \begin{tabular}{ccc}
    \toprule
    Dataset & Without local adjustment & With local adjustment \\
    \midrule
    CC15  & 38.10/0.961 & \textbf{38.24/0.963} \\
    \midrule
    PolyU & 38.77/0.969 & \textbf{38.89/0.971} \\
    \midrule
    HighISO & 40.53/0.974 & \textbf{40.63/0.974} \\
    \midrule
    IOCI  & 41.39/0.977 & \textbf{41.52/0.978} \\
    \midrule
    SIDD-val & 35.20/0.893 & \textbf{35.28/0.894} \\
    \midrule
    Time (s) & \textbf{3.65} & 3.93 \\
    \bottomrule
    \end{tabular}}%
  \label{Table_ablation_study_W_WO_PCA_est}%
  \vspace{-2.8pt}
\end{table}%
\\
\indent The adaptive variant A-Haar-tSVD introduces two weighting parameters, $\beta$ and $\gamma$, which enhance robustness and adaptiveness based on the eigenvalue analysis in (\ref{Equ_other_eigen_value}). We conduct a sensitivity study to evaluate the impact of $\beta$ and $\gamma$, with the results presented in Fig. \ref{Fig_ablation_study_beta_gamma}. We notice that a small $\beta(<1)$ tends to overestimate the noise level, lead to oversmoothing and lower PSNR. On the other hand, choosing an excessively large $\gamma$ may fail to capture the inner-group similarity under noisy conditions. Hence, a reasonable range for $\gamma$ is between 12 and 16. Based on these findings, $\beta$ and $\gamma$ are empirically set to 1.2 and 13 in our experiments, respectively. 

\subsubsection{Limitations and potential enhancement}
From Table \ref{Table_Color_Image_results}, classic patch-based transforms struggle with severely-corrupted images, where noise energy may spread across all frequencies and all principal directions, thus the coefficients of noise overlap in magnitude with those of the true image. Hence, the thresholding operation fails to separate noise from signal. Beyond integrating DNNs with Haar-tSVD, it is interesting to devise an alternative approach within the patch-based paradigm to handle severe noise. Recently, Zontak et al. \cite{zontak2013separating} observed that down-sampled noisy images contain patches with reduced noise and structures similar to clean ones. Therefore, instead of directly filtering the large-size noisy observation, an alternative is to first handle downsized image, and then restore the original resolution using image super-resolution techniques \cite{dong2015image, wang2020deep}.
\section{Conclusion}
In this paper, we present Haar-tSVD, an efficient and effective one-step method for image denoising. We leverage global and local circulant representation to capture similarity and correlation among image patches at both patch and group levels. Under the circulant formulation, we establish a theoretical connection between the Haar transform and PCA, and demonstrate that Haar-tSVD can be modeled by a unified t-SVD and Haar bases, resulting in a plug-and-play and parallelizable filtering approach. The proposed method reduces computational cost by avoiding the training of local transform bases. To achieve further acceleration, we also design and implement fast, parallelizable strategies. Moreover, we investigate the integration and combination of neural networks with the proposed Haar-tSVD. This leads to a flexible noise estimation scheme and an enhancement strategy based on the eigenvalue characteristics of circulant structures. The promising results motivate exploration of the proposed method and its adaptive scheme for broader applications in other imaging domains \cite{fu2024weconvene, luo2024revisiting}. 

\bibliographystyle{IEEEtran}
\bibliography{IEEEabrv, Reference}

\end{document}